\documentclass{article}

\PassOptionsToPackage{numbers, square}{natbib}
\usepackage[final]{neurips_2020}

\usepackage[utf8]{inputenc} 
\usepackage[T1]{fontenc}    
\usepackage{hyperref}       
\usepackage{url}            
\usepackage{booktabs}       
\usepackage{amsfonts}       
\usepackage{nicefrac}       
\usepackage{microtype}      

\usepackage{microtype}
\usepackage{graphicx}
\usepackage{subcaption}
\usepackage{bbm}
\usepackage{amsmath}
\usepackage{bm}
\usepackage{xcolor}

\usepackage{amsthm,amssymb}
\DeclareMathOperator*{\argmax}{arg\,max}
\usepackage{algorithm}
\usepackage{algorithmic}

\title{BAIL: Best-Action Imitation Learning for\\
Batch Deep Reinforcement Learning}

\author{
Xinyue Chen$^1$ \And Zijian Zhou$^1$ \And Zheng Wang$^1$
\AND
Che Wang$^{1, 2}$\And Yanqiu Wu$^{1, 2}$ 
\And Keith Ross$^{1, 2}$\thanks{Correspondence to: Keith Ross <keithwross@nyu.edu>.}\\
\AND
$^1$ \text{\normalfont 
New York University Shanghai}\\ 
$^2$
New York University\\
}
\begin{document}
\maketitle

\begin{abstract}
There has recently been a surge in research in batch Deep Reinforcement Learning (DRL), which aims for learning a high-performing policy from a given dataset without additional interactions with the environment.
We propose a new algorithm, Best-Action Imitation Learning (BAIL), which strives for both simplicity and performance. 
BAIL learns a V function, uses the V function to select actions it
believes to be high-performing, and then uses those actions to train a policy network using imitation learning. For the MuJoCo benchmark, we provide a comprehensive experimental study of BAIL, comparing its performance to four other batch Q-learning and imitation-learning schemes for a large variety of batch datasets.
Our experiments show that BAIL's performance is much higher than the other schemes, and is also computationally much faster than the batch Q-learning schemes. 
\end{abstract}

\section{Introduction}
The field of Deep Reinforcement Learning (DRL) has recently seen a surge in research in batch reinforcement learning, which is the problem of sample-efficient learning from a given dataset without additional interactions with the environment. 
Batch RL enables reusing the data collected by a policy to possibly improve the policy without further interactions with the environment, it has the potential to leverage existing large datasets to obtain much better sample efficiency. A batch RL algorithm can also be deployed as part of a growing-batch algorithm, where the batch algorithm seeks a high-performing exploitation policy using the data in an experience replay buffer \cite{lin1992self}, combines this policy  with exploration to add fresh data to the buffer, and then repeats the whole process \citep{lange2012batch, ernst2005tree}.
Batch RL may also be necessary for learning a policy in safety-critical systems where a partially trained policy cannot be deployed online to collect data. 

\citet{fujimoto2019bcq} made the critical observation that when conventional Q-function based algorithms, such as Deep Deterministic Policy Gradient (DDPG), are directly applied to batch reinforcement learning, they learn very poorly, or even entirely diverge due to {\em extrapolation error}. Therefore, in order to obtain high-performing policies from batch data, new algorithms are required. Recent batch DRL algorithms roughly fall into two categories: Q-function-based algorithms such as BCQ \citep{fujimoto2019bcq} and BEAR \citep{kumar2019bear}; and Imitation Learning (IL)-based algorithms such as MARWIL \citep{wang2018marwil} and AWR \citep{peng2019awr}.

We propose a new algorithm, Best-Action Imitation Learning (BAIL), which strives for both simplicity and performance. BAIL is an advanced IL method and its value estimates are updated only with data in the batch, giving stable estimates. BAIL not only provides state-of-the-art performance, it is also computationally fast. Moreover, it is conceptually and algorithmically simple, thereby satisfying the principle of Occam's razor. 

BAIL has three steps. In the first step, BAIL learns a V function by training a neural network to obtain the ``upper envelope of the data''. In the second step, it selects from the dataset the state action-pairs whose Monte Carlo returns are close to the upper envelope. In the last step, it simply trains a policy network with vanilla imitation learning using the selected actions. The method thus combines a novel approach for V-learning with IL. 

Because the BCQ and BEAR codes are publicly available, we are able to make a careful and comprehensive comparison of the performance of BAIL, BCQ, BEAR, MARWIL and vanilla Behavior Cloning (BC) using the Mujoco benchmark. 
For our experiments, we create training batches in a manner identical to what was done in the BCQ paper (using DDPG \citep{lillicrap2015ddpg} to create the batches), and add additional training batches for the environments Ant and Humanoid using SAC \citep{haarnoja2018sacapps}, giving a total of 22 training batches with non-expert data. Our experimental results show that BAIL wins for 20 of the 22 batches, with overall performance 42\% or more higher than the other algorithms. 
Moreover, BAIL is computationally 30-50 times faster than the Q-learning algorithms. 
BAIL therefore achieves state-of-the-art performance while being significantly simpler and faster than BCQ and BEAR. 

In summary, the contributions of this paper are as follows: $(i)$ BAIL, a new high-performing batch DRL algorithm, along with the novel concept of ``the upper envelope of data''; $(ii)$ extensive, carefully-designed experiments comparing five batch DRL algorithms over diverse datasets. The computational results give significant insight into how different types of batch DRL algorithms perform for different types of data sets. 
We provide public open source code for reproducibility \footnote{\url{https://github.com/lanyavik/BAIL}}. We will also make our datasets publicly available for future benchmarking.

\section{Related work}
Batch reinforcement learning in both the tabular and function approximator settings has long been studied \citep{lange2012batch, strehl2010learning} and continues to be a highly active area of research \citep{swaminathan2015batch, jiang2015doubly, thomas2016data, farajtabar2018more, irpan2019off,jaques2019way}. 

The difficulty of training deep neural networks effectively in the batch setting has been studied and discussed in a series of recent works,  \cite{levine2020offline}. In the case of Q-function based methods, this difficulty is a combined result of extrapolation error and overestimation in Q updates \cite{fujimoto2019bcq, fujimoto2018td3, van2016ddqn}.

Batch-Constrained deep Q-learning (BCQ) avoids the extrapolation error problem by constraining the set of actions over which the approximate Q-function is maximized \citep{fujimoto2019bcq}. More specifically, BCQ first trains a state-dependent Variational Auto Encoder (VAE) using the state action pairs in the batch data. When optimizing the approximate Q-function over actions, instead of optimizing over all actions, it optimizes over a subset of actions generated by the VAE. The BCQ algorithm is further complicated by introducing a perturbation model, which employs an additional neural network that outputs an adjustment to an action. BCQ additionally employs a modified version of clipped-Double Q-Learning to obtain satisfactory performance.  \citet{kumar2019bear} recently proposed BEAR for batch DRL. Like BCQ, BEAR also constrains the actions over which it maximizes the approximate Q function. BEAR is relatively complex, 
employing Maximum Mean Discrepancy \citep{gretton2012kernel}, kernel selection, a parametric model that fits a tanh-Gaussian distribution to the dataset, and a test policy that is different from the learned actor policy. 

Monotonic Advantage Re-Weighted Imitation Learning (MARWIL) \citep{wang2018marwil} uses exponentially weighted imitation learning, with the weights being determined by estimates of the advantage function. Advantage Weighted Regression (AWR) \citep{peng2019awr}, another IL-based scheme, which is conceptually very similar to MARWIL, was primarily designed for online learning, but can also be employed in batch RL. BAIL, being an IL-based algorithm, shares some similarities with MARWIL and AWR; however, instead of weighting with advantage estimates, it uses the novel concept of the upper envelope to learn a V-function and select best actions, providing major performance improvements. In the online case, Self-Imitation Learning learns only from data that have a cumulative discounted return higher than the current value estimates to enhance exploration \citep{oh2018self}, this idea has also been explored in the multi-agent scenario \cite{vinyals2019alphastar}. 

\citet{Agarwal2019offpolicydqn} proposed Random Ensemble Mixture (REM), an ensembling scheme which enforces optimal Bellman consistency on random convex combinations of the Q-heads of a multi-headed Q-network. For the Atari 2600 games, batch REM can out-perform the policies used to collect the data. REM and BAIL are orthogonal, and it may be possible to combine them in the future to achieve even higher performance. Batch algorithms that apply to discrete actions space have been benchmarked on the Atari environments, showing that a discrete variant of BCQ has robust performance and outperform a number of other methods \cite{fujimoto2019benchmarking}. In the continuous action space case, MuJoCo has been the major benchmark. Very recently, other benchmarks environments have also been proposed for more systematic comparison and analysis \cite{fu2020d4rl}. 

\section{Batch Deep Reinforcement Learning}

We represent the environment with a Markov Decision Process (MDP) defined by a tuple $(\mathcal{S} , \mathcal{A} , g , r,  \rho,\gamma)$, where $\mathcal{S}$ is the state space, $\mathcal{A}$ is the action space, $\rho$ is the initial state distribution, and $\gamma$ is the discount factor.  The functions $g(s,a)$ and $r(s,a)$ represent the dynamics and reward function, respectively. In this paper we assume that the dynamics of the environment are deterministic, that is, there are real-valued functions $g(s,a)$ and $r(s,a)$ such that when in state $s$ and action $a$ is chosen, then the next state is $s' = g(s,a)$ and the reward received is $r(s,a)$. We note that all the simulated robotic locomotion environments in the MuJoCo benchmark are deterministic.
Furthermore, many of the Atari game environments are deterministic \citep{bellemare2013arcade}. Thus, from an applications perspective, the class of deterministic environments is a large and important class. Although we assume that the environment is deterministic, as is typically the case with reinforcement learning, we do not assume the functions $g(s,a)$ and $r(s,a)$ are known.

In batch reinforcement learning, we are provided a batch of $m$ data points $\mathcal{B} = \{ (s_i,a_i,r_i,s'_i), \, i = 1,...,m \}$. Using this batch, the goal is train a high-performing policy without any and further interaction with the environment. 
Typically the batch $\mathcal{B}$ is training data obtained while training a policy in some episodic fashion, or is execution data obtained with a fixed deterministic policy over multiple episodes.
In the batch reinforcement learning problem, we do not have knowledge of the algorithm, policy, or seeds that were used to generate the episodes in the batch $\mathcal{B}$. 

\section{Best-Action Imitation Learning (BAIL)}

In this paper we present BAIL, an algorithm that not only provides state-of-the-art performance on simulated robotic locomotion tasks, but is also fast and algorithmically simple.
The motivation behind BAIL is as follows. For a given deterministic MDP, let $V^*(s)$ be the optimal value function. 
For a particular state-action pair $(s,a)$, 
let $G(s,a)$ denote a return using some policy when beginning state $s$ and choosing action $a$.  
Any action $a^*$ that satisfies $G(s,a^*) = V^*(s)$ is an optimal action for state $s$.
Thus, ideally we would like to construct an algorithm which finds actions that satisfy $G(s,a^*) = V^*(s)$ for each state $s$. 


In  batch reinforcement learning, since we are only given limited data, we can only hope to obtain an approximation of $V^*(s)$. In BAIL, we first try to make the best possible estimate of $V^*(s)$ using only the limited information in the batch dataset. Call this estimate $V(s)$. We then select state-action pairs from the dataset whose associated returns $G(s,a)$ are close to $V(s)$. 
Finally, we train a policy with IL using the selected state-action pairs. Thus, BAIL combines both V-learning and IL. To obtain the estimate $V(s)$ of the value function, we introduce the ``upper envelope of the data''. 
 
 \subsection{Upper envelope of the data}
 
We first define a $\lambda$-regularized upper envelope, and then provide an algorithm for finding it. To the best of our knowledge, the notion of the upper envelope of a dataset is novel.
 
Recall that we have a batch of data $\mathcal{B} = \{(s_i,a_i,r_i,s'_i), \, i = 1,...,m\}$. Although we do not assume we know what algorithm was used to generate the batch, we make the natural assumption that the data in the batch was generated in an episodic fashion, and that the data in the batch is ordered accordingly. 
For each data point $i \in \{1,\ldots,m\}$, we calculate the Monte Carlo return $G_i$ as the sum of the discounted rewards from state $s_i$ to the end of the episode as $G_i = \sum_{t=i}^{T} \gamma^{t-i} r_t$
where $T$ denotes the time at which the episode ends for the episode that contains the $i$th data point.

Having defined the return for each data point in the batch, we now seek an upper-envelope of the data $\cal{G}$ $:= \{(s_i,G_i), \, i = 1,...,m\}$. Let $V_{\phi}(s)$ denote a neural networks parameterized by $\phi=(w,b)$ that takes as input a state $s$ and outputs a real number, where $w$ and $b$ denote the weights and bias,  respectively. 
For a fixed $\lambda \geq 0$,
we say that $V_{\phi^\lambda}(s)$ is a $\lambda$-regularized upper envelope for $\cal{G}$ if $\phi^\lambda$ is an optimal solution for the following constrained optimization problem:
\begin{equation}
\min_\phi \sum_{i=1}^m[V_\phi(s_i)-G_i]^2 + \lambda\|w\|^2 \qquad s.t. \qquad V_\phi(s_i) \geq G_i, \qquad i = 1,2, \dots, m 
\label{UE-objective}
\end{equation}
Note that a $\lambda$-regularized upper envelope always lies above all the returns. The optimization problem strives to bring the envelope as close to the data as possible while maintaining regularization to prevent overfitting. 
The solution $\phi^\lambda$ to the constrained optimization problem may not be unique. Nevertheless, we have the following theorem to characterize the limiting behavior of $\lambda$-regularized upper envelopes.

\newtheorem{theorem}{Theorem}[section]
\begin{theorem}
\label{theorem1}
Suppose $V_\phi(s)$ is a multi-layer fully connected neural network with ReLu activation units, and there is a bias term at the output layer.
For each $\lambda \geq 0$, let $V_{\phi^\lambda}(s)$ be a $\lambda$-regularized upper envelope for $\mathcal{G}$, that is, $\phi^\lambda=(w^\lambda, b^\lambda)$ is an optimal solution of the above constrained optimization problem. Then, we have
\begin{enumerate}
    \item[(1)] 
    $\lim \limits_{\lambda \rightarrow \infty} V_{\phi^\lambda}(s)  = \max \limits_{1\le i \le m}\{G_i\}$ for all $s \in \mathcal{S}$. 
    \item[(2)] When $\lambda = 0$, if there are sufficient number of activation units and layers, then $V_{\phi^0}(s) $ will interpolate the data in $\mathcal{G}$, i.e., $V_{\phi^0}(s_i)=G_i$ for all $i = 1,\dots,m$.
\end{enumerate}
\end{theorem}
From the above theorem, we see that when $\lambda$ is very small, the upper envelope aims to interpolate the data, and when $\lambda$ is large, the upper envelope approaches a constant going through the data point with the highest return.
Just as in classical regression, there is a sweet-spot for $\lambda$, the one that provides the best generalization. 

We solve the constrained optimization problem (\ref{UE-objective}) 
with a penalty-function approach. 
Specifically, to obtain an approximate upper envelope  of the data $\mathcal{G}$, we solve an unconstrained optimization problem with a penalty loss function (with $\lambda$ fixed):
\begin{equation} 
\begin{split}
L^K(\phi) &= \sum_{i=1}^m ( V_{\phi}(s_i) - G_i )^2 \{ \mathbbm{1}_{(V_{\phi}(s_i) \ge  G_i)} + K \cdot \mathbbm{1}_{(V_{\phi}(s_i) <  G_i)} \} + \lambda \|w\|^2
\end{split}
\label{penalty_loss}
\end{equation}           
where $K >> 1$ is the penalty coefficient and $\mathbbm{1}_{(\cdot)}$ is the indicator function.
For a finite $K$ value, the penalty loss function will produce an approximate upper envelope $V(s_i)$, since $V(s_i)$ may be slightly less than $G_i$ for some data points.
In practice, we find $K=1000$ works well for all environments tested. For $K \rightarrow \infty$, we have the following theoretical justification of the approximation:
\begin{theorem}
\label{theorem2}
Let $\phi^K$ be a solution that minimizes $L^K(\phi)$ with penalty constant $K$. Let $\phi^*$ be a limit point of $\{\phi^K\}$. Then 
$V_{\phi^*}(s)$  is an exact $\lambda$-regularized upper envelope, i.e., $\phi^*$ is an optimal solution for the constrained optimization problem (\ref{UE-objective}).
\end{theorem}

In practice, instead of $L_2$ regularization,  we employ 
a mechanism similar to early-stopping regularization.
We split the data into a training set and validation set. During training, after every epoch, we check whether the validation error for the penalty loss function 
(\ref{penalty_loss}) decreases, and stop updating the network parameters when the validation loss increases repeatedly. We describe the details of the early-stopping scheme in the supplementary materials. 

Figure \ref{fig:ue_visual_4env} provides some examples of upper envelopes obtained with training sets consisting of 1 million data points. 
Each figure shows the upper envelope and the returns for one environment. To aid visualization, the states are ordered in terms of their upper envelope $V(s_i)$ values.  

\begin{figure}[ht]
\label{UE_figures}
\vskip 0.1in
\centering
\begin{subfigure}{0.245\columnwidth}
 \centering
 \includegraphics[width=1\linewidth]{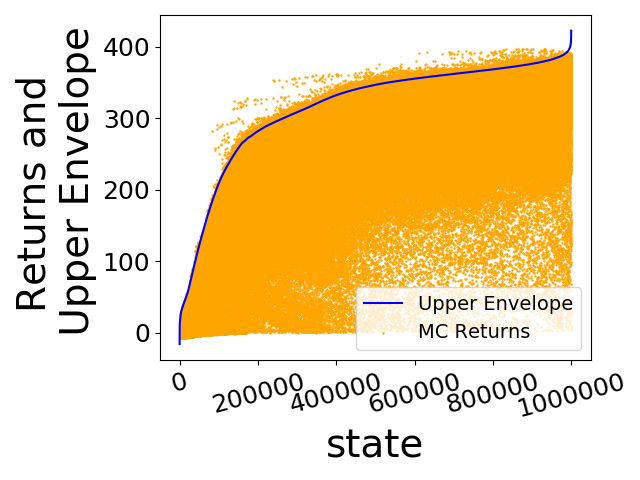}
 \caption{Hopper}
\end{subfigure}
\begin{subfigure}{0.245\columnwidth}
 \centering
 \includegraphics[width=1\linewidth]{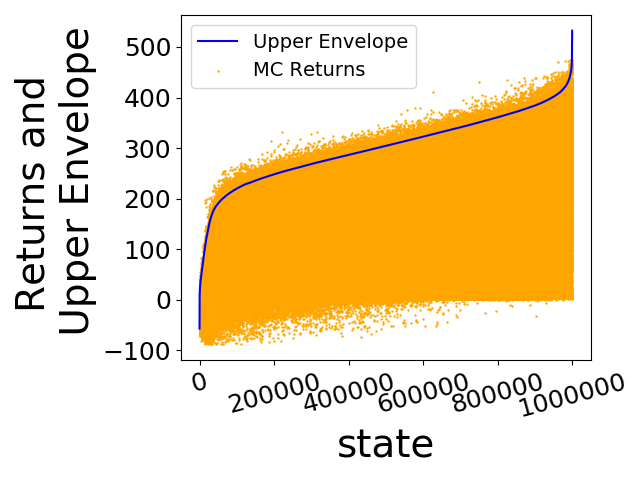}
 \caption{Walker2d}
\end{subfigure}
\begin{subfigure}{0.245\columnwidth}
 \centering
 \includegraphics[width=1\linewidth]{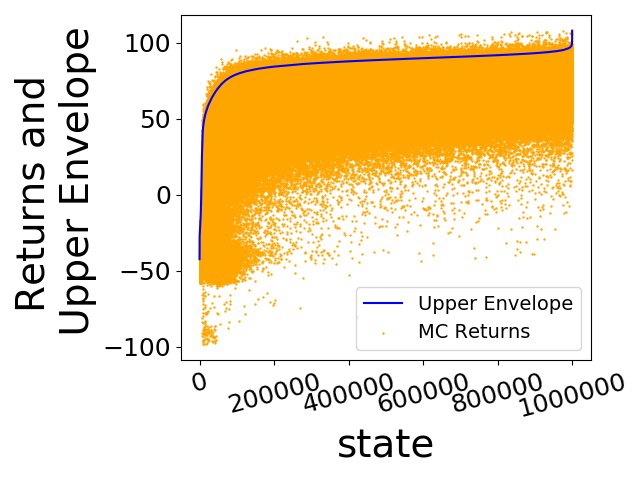}
 \caption{HalfCheetah}
\end{subfigure}
\begin{subfigure}{0.245\columnwidth}
 \centering
 \includegraphics[width=1\linewidth]{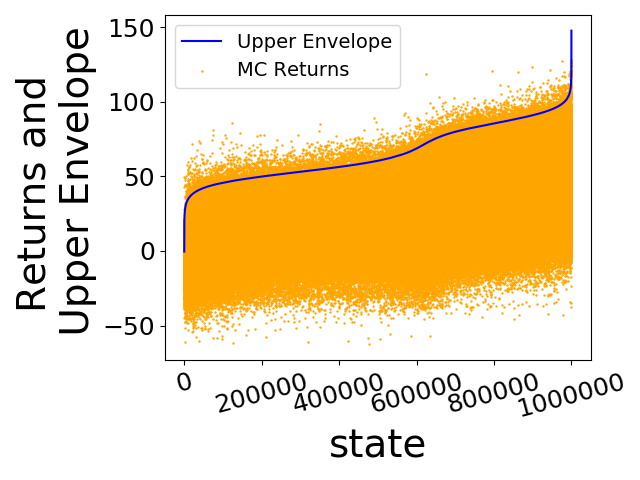}
 \caption{Ant}
\end{subfigure}
\caption{Upper Envelopes trained on batches from different MuJoCo environments. }
\label{fig:ue_visual_4env}
\vskip -0.1in
\end{figure}

\subsection{Selecting the best actions}
BAIL employs the upper envelope to select the best $(s,a)$ pairs from the batch data $\mathcal{B}$. 
Let $V(s)$ denote the upper envelope obtained from minimizing the penalty loss function (\ref{penalty_loss}) for a fixed value of $K$. We consider two approaches for selecting the best actions. In the first approach, which we call BAIL-ratio, for a fixed $x > 0$, we choose all $(s_i,a_i)$ pairs from the batch data set $\mathcal{B}$ such that
\begin{equation}
\label{eq:bail-ratio_selection}
G_i > x V(s_i)
\end{equation}
We set $x$ such that $p\%$ of the data points are selected, where $p$ is a hyper-parameter. In this paper we use $p = 25\%$ for all environments and batches. In the second approach, which we call BAIL-difference, for a fixed $x > 0$, we choose all $(s_i,a_i)$ pairs from the batch data set $\mathcal{B}$ such that
\begin{equation}
\label{eq:bail-difference_selection}
G_i  \geq V(s_i) - x
\end{equation}
In our experiments, BAIL-ratio and BAIL-difference have similar performance, with BAIL-ratio sometimes a little better. We henceforth only consider BAIL-ratio, and simply refer to it as BAIL.

In summary, BAIL employs two neural networks. The first network is used to approximate the optimal value function based on the data in the batch $\mathcal{B}$. The second network is the policy, which is trained with imitation learning. 
We refer to the algorithm just described as BAIL. We also consider a variation, which we call Progressive BAIL, in which we train the upper envelope parameters $\phi$ and the policy network parameters $\theta$ in parallel rather than sequentially. Progressive BAIL doesn't change how we obtain the upper envelope, since the upper envelope does not depend on the policy parameters in either BAIL or Progressive BAIL. It does, however, affect the training of the policy parameters. We provide detailed pseudo-code for both BAIL and Progressive BAIL  in the supplementary materials. 
Our experimental results show that BAIL and Progressive BAIL both perform well with about the same performance over all batches. But BAIL might be a better choice since it's much faster.  

\subsection{Augmented and oracle returns for the MuJoCo benchmark}

Both BCQ and BEAR papers use the MuJoCo robotic locomotive benchmarks to gauge the performance of their algorithms \citep{fujimoto2019bcq} \citep{kumar2019bear}. We will compare the performance of BAIL with BCQ, BEAR, MARWIL and BC using the same MuJoCo environments. 

The MuJoCo environments are naturally infinite-horizon non-episodic continuing-task environments \citep{sutton2018reinforcement}. During training, however, researchers typically create artificial episodes of maximum length 1000 time steps; after 1000 time steps, a random initial state is chosen and a new episode begins. This means that to apply BAIL, we need to approximate infinite-horizon discounted returns using the finite-length episodes in the data set. For data points appearing near the beginning of the episode, the finite-horizon return will closely approximate the (idealized) infinite-horizon return due to discounting; but for a data point near the end of an episode, the finite horizon return can be inaccurate and should be augmented. To calculate the augmentation for the $i$th data point, we use the following heuristic. Let $\mathcal{E} \subset \mathcal{B}$ denote the episode of data that contains the $i$th data point, and let $s'$ be the last state in episode $\mathcal{E}$. We then set $s_j$ to be the state in
the first $\max\{1000 - i, 200\}$ data points of the episode $\mathcal{E}$ that is closest (in Euclidean norm) to the ``terminal state'' $s'$. We then set
\begin{equation}
     G_i = \sum_{t=i}^{T} \gamma^{t-i} r_t + \gamma^{T-i+1} \sum_{t=j}^{T} \gamma^{t-j} r_t
     \label{augmentation}
\end{equation}
Note that $G_i$ in (\ref{augmentation}) will have at least 800 terms, so there is no need for additional terms due to the discounting. Importantly, the rewards in the two sums in (\ref{augmentation}) are generated by the same policy. The first sum uses the actual rewards accrued until to the end of the episode; the second sum approximates what the actual rewards would have been if the episode was allowed to continue past 1000 time steps. 

To validate this heuristic, we compute oracle returns by letting episodes run up to 2000 time steps. In this manner, every return is calculated with at least 1000 actual rewards, and is therefore essentially exact due to discounting. With oracle returns, we can analyze the effect of the augmentation heuristic. Specifically, we compared the performance of BAIL with the augmentation heuristic and with oracle for Hopper-v2 for seven diverse batches. 
Learning curves are shown in the supplementary materials. For all five batches, the augmentation heuristic has similar performance compared to the oracle. We conclude that our augmentation heuristic is a satisfactory method for addressing continual environments such as MuJoCo, which is also confirmed with good performance in Table \ref{all-data-table} .

\section{Experimental results}

Along with the BAIL algorithm, the experimental results are the main contribution of this paper. Using 62 diverse batches (many of which are similar to those used in the BCQ and BEAR papers), we provide a comprehensive comparison of 
five algorithms: BAIL, BCQ, BEAR, MARWIL and BC. We use authors' code and recommended hyper-parameters when available, and we strive to make the comparisons as fair as possible. 

\begin{table*}[t]
\centering
	\caption{Performance of five Batch DRL algorithms for 22 different training datasets.} 
	\label{all-data-table}
	\vskip 0.15in
	\begin{center}
		\begin{small}
			\begin{sc}
				\begin{tabular}{lccccr}
					\toprule
					Environment & BAIL & BCQ & BEAR & BC & MARWIL\\
                    \midrule
$\sigma=0.1$ Hopper B1  & $\bm{2173\pm291}$ & $1219\pm114$ & $505\pm285$ & $626\pm112$ & $827\pm220$ \\
$\sigma=0.1$ Hopper B2  & $\bm{2078\pm180}$ & $1178\pm87$ & $985\pm3$ & $579\pm141$ & $620\pm336$ \\
$\sigma=0.1$ Walker B1  & $\bm{1125\pm113}$ & $576\pm309$ & $610\pm212$ & $514\pm17$ & $436\pm24$ \\
$\sigma=0.1$ Walker B2  & $\bm{3141\pm300}$ & $2338\pm388$ & $2707\pm425$ & $1741\pm239$ & $1810\pm200$ \\
$\sigma=0.1$ HC B1  & $\bm{5746\pm29}$ & $\bm{5883\pm43}$ & $0\pm0$ & $\bm{5546\pm29}$ & $\bm{5573\pm35}$ \\
$\sigma=0.1$ HC B2  & $\bm{7212\pm43}$ & $\bm{7562\pm31}$ & $0\pm0$ & $6765\pm108$ & $\bm{6828\pm111}$ \\
$\sigma=0.5$ Hopper B1  & $\bm{2054\pm158}$ & $1145\pm300$ & $203\pm42$ & $919\pm52$ & $946\pm103$ \\
$\sigma=0.5$ Hopper B2  & $\bm{2623\pm282}$ & $1823\pm555$ & $241\pm239$ & $694\pm64$ & $818\pm112$ \\
$\sigma=0.5$ Walker B1  & $\bm{2522\pm51}$ & $1552\pm455$ & $1248\pm181$ & $2178\pm178$ & $2111\pm52$ \\
$\sigma=0.5$ Walker B2  & $\bm{3115\pm133}$ & $2785\pm123$ & $2302\pm630$ & $2483\pm94$ & $2364\pm228$ \\
$\sigma=0.5$ HC B1  & $1055\pm9$ & $\bm{1222\pm38}$ & $924\pm579$ & $570\pm35$ & $512\pm43$ \\
$\sigma=0.5$ HC B2  & $\bm{7173\pm120}$ & $5807\pm249$ & $-114\pm140$ & $\bm{6545\pm171}$ & $\bm{6668\pm93}$ \\
SAC Hopper B1  & $\bm{3296\pm105}$ & $2681\pm438$ & $1000\pm110$ & $2853\pm318$ & $2897\pm227$ \\
SAC Hopper B2  & $1831\pm915$ & $\bm{2134\pm917}$ & $1139\pm317$ & $\bm{2240\pm367}$ & $\bm{2063\pm168}$ \\
SAC Walker B1  & $\bm{2455\pm211}$ & $\bm{2408\pm84}$ & $-3\pm5$ & $1674\pm277$ & $1484\pm140$ \\
SAC Walker B2  & $\bm{4767\pm130}$ & $3794\pm398$ & $325\pm75$ & $2599\pm145$ & $2651\pm268$ \\
SAC HC B1  & $\bm{10143\pm77}$ & $8607\pm473$ & $7392\pm257$ & $8874\pm221$ & $9105\pm90$ \\
SAC HC B2  & $\bm{10772\pm59}$ & $\bm{10106\pm134}$ & $7217\pm273$ & $9523\pm164$ & $9488\pm136$ \\
SAC Ant B1  & $\bm{4284\pm64}$ & $\bm{4042\pm113}$ & $3452\pm128$ & $\bm{3986\pm112}$ & $\bm{4033\pm130}$ \\
SAC Ant B2  & $\bm{4946\pm148}$ & $\bm{4640\pm76}$ & $3712\pm236$ & $\bm{4618\pm111}$ & $\bm{4589\pm130}$ \\
SAC Humanoid B1  & $\bm{3852\pm430}$ & $1411\pm250$ & $0\pm0$ & $543\pm378$ & $589\pm121$ \\
SAC Humanoid B2  & $\bm{3565\pm153}$ & $1221\pm207$ & $0\pm0$ & $1216\pm826$ & $1033\pm257$ \\
					\bottomrule
				\end{tabular}
			\end{sc}
		\end{small}
	\end{center}
	\vskip -0.1in
\end{table*}

\subsection{Batch generation}
We provide experimental results for five MuJoCo environments: HalfCheetah-v2, Hopper-v2, Walker2d-v2, Ant-v2, and Humanoid-v2. 
To make the comparison as favorable as possible for BCQ and BEAR, we generate datasets using the same procedures proposed in the BCQ and BEAR papers, and compare the algorithms using those data sets. 
The BCQ \citep{fujimoto2019bcq} and BEAR \citep{kumar2019bear} papers both use batch datasets of one million samples, but generate the batches using different approaches. One important observation we make, which was not brought to light in previous batch DRL papers, is that batches generated with different seeds but with otherwise exactly the same algorithm can give drastically different results for batch DRL. Because of this, for every experimental scenario considered in this paper, we generate two batches, each generated with a different random seed. 

\subsubsection{Training batches}
We generate batches {\em while} training DDPG \citep{lillicrap2015ddpg} from scratch with exploration noise of $\sigma = 0.5$ for HalfCheetah-v2, Hopper-v2, and Walker2d-v2, as exactly done in the BCQ paper. We also generate batches with $\sigma = 0.1$ to study the robustness of tested algorithms with lower noise level. We also generate training batches for all five environments by training policies with adaptive Soft Actor Critic (SAC) \citep{haarnoja2018sacapps}. This gives six DDPG and five SAC scenarios. For each one, we generate two batches with different random seeds, giving a total of 22 ``training batches'' composed of non-expert data. These batches are the most important ones to measure the performance of a batch method, since they contain sub-optimal data obtained from the training process well before optimal performance is achieved (and in many cases using sub-optimal algorithms for training), which is difficult for vanilla behavioral cloning to use. 

\subsubsection{Execution batches}

In addition to training batches, we also study execution batches. We do a similar procedure as in the BEAR paper: first train SAC \citep{haarnoja2018sacapps} for a certain number of environment interactions, then {\em fixes the trained policy} and generates one million ``execution data points''. The BEAR paper generates batches with ``mediocre data'' where training is up to a mediocre performance, and with ``optimal data'' where training is up to near optimal performance. 
When generating the batches with the trained policy, the BEAR paper continues to include exploration noise, using the trained $\sigma(s)$ values in the policy network. 
Since after training, a test policy is typically deployed without exploration noise, we also consider noise-free data generation. The BEAR paper considers the same five MuJoCo environments considered here. This gives rise to 20 scenarios. For each one we generate two batches with random seeds, giving a total of 40 ``execution batches".

\begin{figure}[ht]
\vskip 0.1in
\centering
\begin{subfigure}{0.32\columnwidth}
 \centering
 \includegraphics[width=0.99\linewidth]{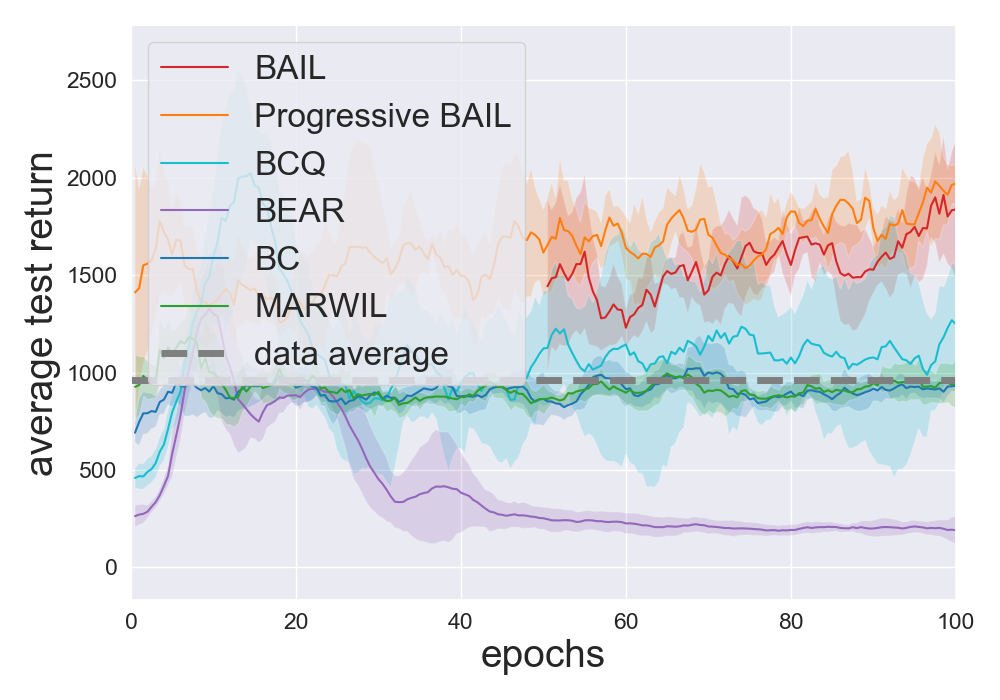}
 \caption{Hopper, batch 1}
\end{subfigure}
\begin{subfigure}{0.32\columnwidth}
 \centering
 \includegraphics[width=0.99\linewidth]{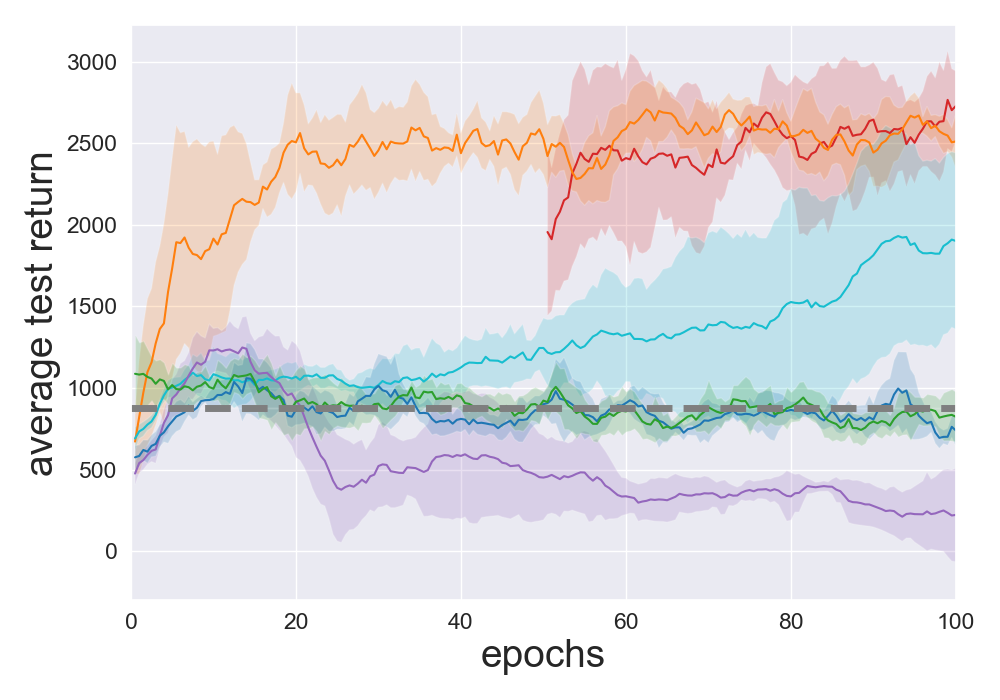}
 \caption{Hopper, batch 2}
\end{subfigure}
\begin{subfigure}{0.32\columnwidth}
 \centering
 \includegraphics[width=0.99\linewidth]{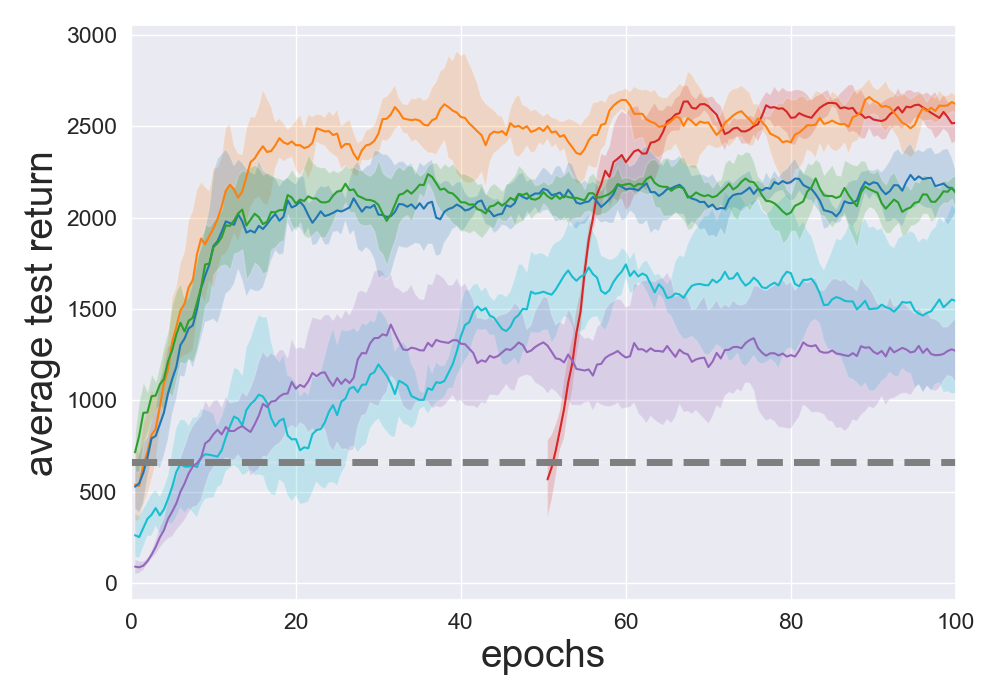}
 \caption{Walker2d, batch 1}
\end{subfigure}
\begin{subfigure}{0.32\columnwidth}
 \centering
 \includegraphics[width=0.99\linewidth]{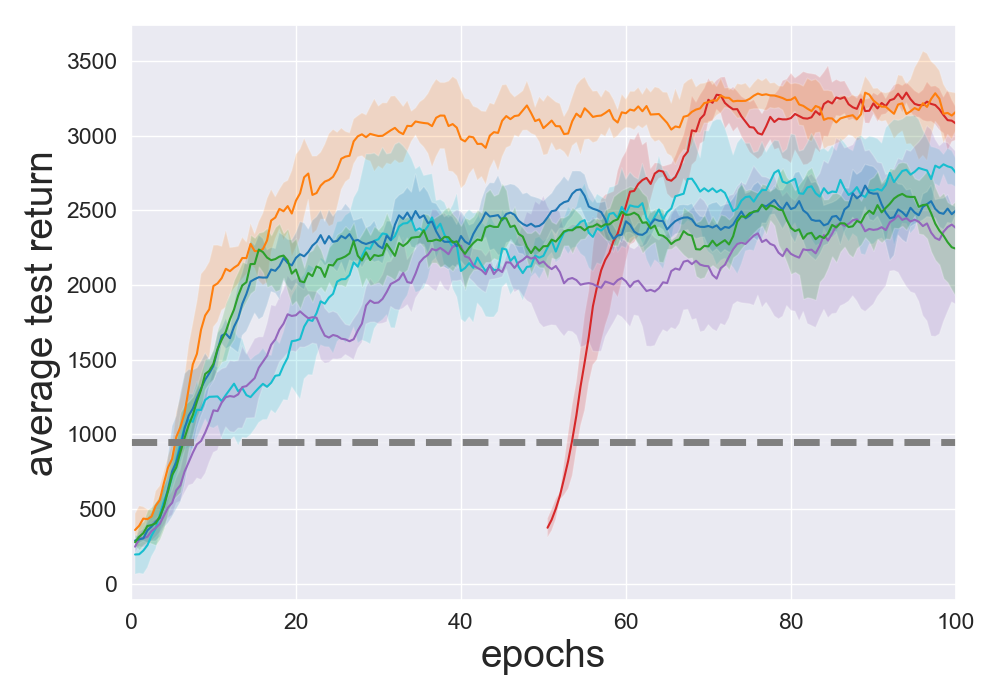}
 \caption{Walker2d, batch 2}
\end{subfigure}
\begin{subfigure}{0.32\columnwidth}
 \centering
 \includegraphics[width=0.99\linewidth]{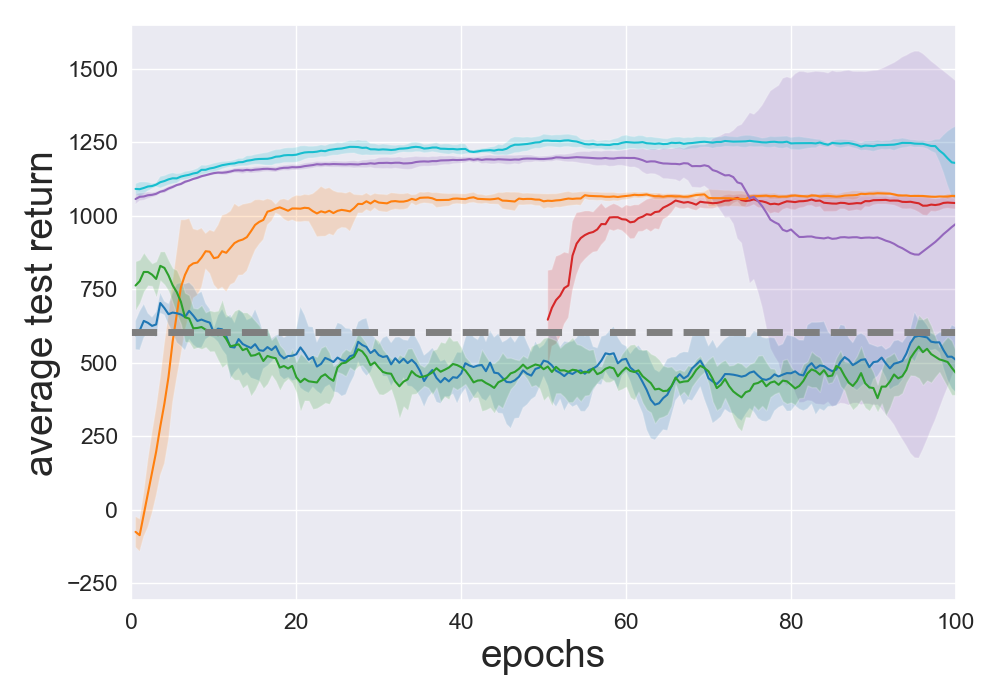}
 \caption{HalfCheetah, batch 1}
\end{subfigure}
\begin{subfigure}{0.32\columnwidth}
 \centering
 \includegraphics[width=0.99\linewidth]{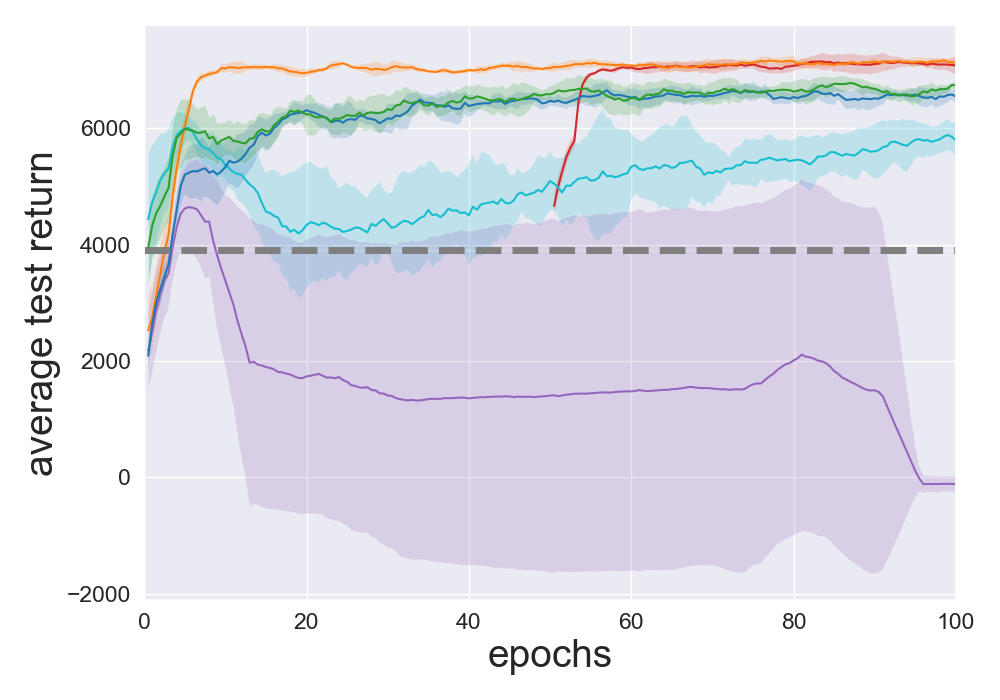}
 \caption{HalfCheetah, batch 2}
\end{subfigure}
\caption{Learning curves using DDPG training batches with $\sigma = 0.5$. }
\label{fig:bail_final_n_0-5}
\vskip -0.1in
\end{figure}

 \subsection{Performance comparison}

We now carefully compare the five algorithms. For a fair comparison, we keep all hyper-parameters fixed for all experiments, instead of fine-tuning for each one. For BCQ we use the authors' code with their default hyper-parameters. For BEAR we use the authors' code with  their version ``0'' with ``use ensemble variance'' set to False and employ the recommended hyper-parameters. Because the MARWIL code is not publicly available, we write our own code, and use neural networks the same size as in BAIL. In the supplementary material we provide more details on implementations, and explain how the comparisons are carefully and fairly done. 


For each algorithm, we train for 100 epochs (with each epoch consisting of one million data points). For each algorithm, after every 0.5 epochs, we run ten test episodes with the current policy to evaluate performance. We then repeat the procedure for five seeds to obtain the mean and confidence intervals shown in the learning curves.  Due to page length limitations, we cannot present the learning curves for all 62 datasets. Here we focus on the 6 DDPG training data batches with $\sigma = 0.5$ (corresponding to the datasets in the BCQ paper), and present the learning curves for the other batches in the supplementary material. However, we present summary results for all datasets in this section. 

Figure \ref{fig:bail_final_n_0-5} shows the learning curves for the 6 DDPG data sets over 100 epochs. As is commonly done, we present smoothed average performance and standard deviations. Note that for BAIL, all curves start at 50 epochs. This provides a fair comparison, since for BAIL we first use 50 epochs of data to train the upper envelopes and then use imitation learning to train the policy network. 
The horizontal grey dashed line indicates the average return of episodes contained in the batch. 

Table \ref{all-data-table} presents our main results, comparing the five algorithms for the 22 training batches. In batch DRL, since there is no interaction with the environment, one cannot use the policy parameters that provided the highest test returns to assess performance. So in Table \ref{all-data-table} , for each algorithm, we assume that the practitioner would use the policy obtained after 100 epochs of training. Since there can be significant variation in performance from one policy to the next during training, we calculate the average  performance across epochs 95.5 to 100 (i.e., averaged over the last ten tested policies). We do this for each of the five training seeds. We then report the average and standard deviation of these values across the five seeds. For each batch, all the algorithms that are within 10\% of the highest average value are considered winners and are indicated in bold.

From Table \ref{all-data-table}  we observe that for the training batches, BAIL is the clear winner.  BAIL wins for 20 of the 22 batches, and BCQ is in second place winning for only 8 of the 22 batches. These results show that BAIL is very robust for a wide variety of training datasets, including non-expert datasets, datasets generated as described in the BCQ paper, and for datasets with the more challenging Ant and Humanoid environments. 
To evaluate the average performance improvement of BAIL over BCQ, for each batch we take the ratio of the BAIL performance to the BCQ performance and then average over the 22 bathes. We also do the same for BC. With this metric, BAIL performs 42\% better than BCQ, and 101\% better than BC. BAIL is also more stable across seeds: The normalized standard deviations (standard deviation divided by average performance) of BAIL, averaged over the 22 batches, is about half that of BCQ. Because BAIL performs so well for training batches, BAIL can potentially be successfully used for growing batch DRL. 

We also note that BEAR occasionally performs very poorly. This is likely because we are using one set of recommended BEAR hyper-parameters for all environments, whereas the BEAR paper reports results using different hyper-parameters for different environments. We also note that for the MuJoCo environments, MARWIL performs similarly to BC. 

For the execution batches, the results are given in the supplementary materials. When BAIL uses the same hyper-parameters as for training batches (though fine-tuning will yield better results, we strive for a fair comparison), BC, MARWIL, BAIL, and BCQ have similar overall performance, with BC being the most robust and the overall winner. Comparing BAIL and BCQ, BAIL has slightly stronger average performance score, and BCQ has a few more wins.  It is no surprise that BC is the strongest here, since the execution batches are generated with a single fixed policy and are easy for BC to learn well. These results imply that the focus of future research on batch DRL should be on training batches, or other diverse datasets, since vanilla BC already works very well for fixed-policy datasets.

BAIL uses an upper envelope to select the ``best'' data points for training a policy network with imitation learning.  It is natural to ask how BAIL would perform when using the more naive approach of  selecting the best actions by simply selecting the same percentage of data points with the highest $G_i$ values, and also by constructing the value function with regression rather than with an upper envelope. These schemes do not do as well as BAIL by a wide margin (see supplementary material). 

Intuitively, BAIL can perform better than BCQ and BEAR because these policy-constraint methods rely on carefully tuned constraints to prevent the use of out-of-distribution actions. A loose constraint can cause extrapolation error to accumulate quickly, and a tight constraint will prevent the policy from choosing some of the good actions. BAIL, however, identifies and imitates the highest-performing actions in the dataset, thus avoiding the need to carefully tune such a constraint. 

\subsection{Comparison of run times for Batch DRL algorithms}
In our experiments we run all algorithms each for $100$ epochs for five seeds for each batch. 
For training with one seed,  it takes $1\sim 2$ hours for BAIL (including the time for upper envelope training and imitation learning), $12\sim 24$ hours for Progressive BAIL, $36\sim 72$ hours for BCQ and $60\sim 100$ hours for BEAR on a CPU node. Thus, roughly speaking, training BAIL is roughly 35 times faster than BCQ and 50 times faster than BEAR. 

\section{Conclusion}
In conclusion, our experimental results show that $(i)$ for the training data batches, BAIL is the clear winner, winning for 20 of 22 batches with a performance improvement of 42\% over BCQ and 101\% over BC;
$(ii)$ for the execution batches, vanilla BC does well with not much room for improvement, although BAIL and BCQ are almost as good and occasionally beat BC by a small amount;  $(iii)$  BAIL is computationally much faster than the Q-learning-based algorithms BCQ and BEAR. 

The results in this paper show that it is possible to achieve state-of-the art performance with a simple, computationally fast IL-based algorithm. BAIL is based on the notion of the ``upper envelope of the data'', which appears to be novel and may find applications in other machine-learning domains. One potential future research direction is to combine batch methods such as BAIL with exploration techniques to build robust online algorithms for better sample efficiency. Another potential direction is to develop methods that are more robust across different batches and hyperparameters and study what makes them robust. Such robustness can greatly improve computation time, and might be safer to work with when deployed to real-world systems.

\clearpage
\section*{Broader Impact}
This research may potentially lead to mechanisms for training robots and self-driving vehicles to perform complex tasks. Batch RL enables reusing the data collected by a policy to possibly improve the policy without further interactions with the environment. A batch RL algorithm can be deployed as part of a growing-batch algorithm, where the batch algorithm seeks a high-performing exploitation policy using the data in an experience replay buffer, combines this policy  with exploration to add fresh data to the buffer, and then repeats the whole process. Batch RL may also be necessary for learning a policy in safety-critical systems where a partially trained policy cannot be deployed online to collect data. Compared to policy constraint methods discussed in the paper, BAIL uses a much smaller amount of computation to achieve good performance, its efficiency means a smaller computation cost and less consumption of energy.

\begin{ack}
This research was partially supported by Nokia Bell Labs.
\end{ack}

\bibliographystyle{plainnat}
\bibliography{neurips_2020.bib}

\begin{thebibliography}{36}
\providecommand{\natexlab}[1]{#1}
\providecommand{\url}[1]{\texttt{#1}}
\expandafter\ifx\csname urlstyle\endcsname\relax
  \providecommand{\doi}[1]{doi: #1}\else
  \providecommand{\doi}{doi: \begingroup \urlstyle{rm}\Url}\fi

\bibitem[Agarwal et~al.(2019)Agarwal, Schuurmans, and
  Norouzi]{Agarwal2019offpolicydqn}
Rishabh Agarwal, Dale Schuurmans, and Mohammad Norouzi.
\newblock Striving for simplicity in off-policy deep reinforcement learning.
\newblock \emph{arXiv preprint arXiv:1907.04543}, 2019.

\bibitem[Bellemare et~al.(2013)Bellemare, Naddaf, Veness, and
  Bowling]{bellemare2013arcade}
Marc~G Bellemare, Yavar Naddaf, Joel Veness, and Michael Bowling.
\newblock The arcade learning environment: An evaluation platform for general
  agents.
\newblock \emph{Journal of Artificial Intelligence Research}, 47:\penalty0
  253--279, 2013.

\bibitem[Ernst et~al.(2005)Ernst, Geurts, and Wehenkel]{ernst2005tree}
Damien Ernst, Pierre Geurts, and Louis Wehenkel.
\newblock Tree-based batch mode reinforcement learning.
\newblock \emph{Journal of Machine Learning Research}, 6\penalty0
  (Apr):\penalty0 503--556, 2005.

\bibitem[Farajtabar et~al.(2018)Farajtabar, Chow, and
  Ghavamzadeh]{farajtabar2018more}
Mehrdad Farajtabar, Yinlam Chow, and Mohammad Ghavamzadeh.
\newblock More robust doubly robust off-policy evaluation.
\newblock \emph{arXiv preprint arXiv:1802.03493}, 2018.

\bibitem[Fu et~al.(2020)Fu, Kumar, Nachum, Tucker, and Levine]{fu2020d4rl}
Justin Fu, Aviral Kumar, Ofir Nachum, George Tucker, and Sergey Levine.
\newblock D4rl: Datasets for deep data-driven reinforcement learning.
\newblock \emph{arXiv preprint arXiv:2004.07219}, 2020.

\bibitem[Fujimoto et~al.(2018)Fujimoto, van Hoof, and Meger]{fujimoto2018td3}
Scott Fujimoto, Herke van Hoof, and Dave Meger.
\newblock Addressing function approximation error in actor-critic methods.
\newblock \emph{arXiv preprint arXiv:1802.09477}, 2018.

\bibitem[Fujimoto et~al.(2019{\natexlab{a}})Fujimoto, Conti, Ghavamzadeh, and
  Pineau]{fujimoto2019benchmarking}
Scott Fujimoto, Edoardo Conti, Mohammad Ghavamzadeh, and Joelle Pineau.
\newblock Benchmarking batch deep reinforcement learning algorithms.
\newblock \emph{arXiv preprint arXiv:1910.01708}, 2019{\natexlab{a}}.

\bibitem[Fujimoto et~al.(2019{\natexlab{b}})Fujimoto, Meger, and
  Precup]{fujimoto2019bcq}
Scott Fujimoto, David Meger, and Doina Precup.
\newblock Off-policy deep reinforcement learning without exploration.
\newblock \emph{arXiv preprint arXiv:1812.02900}, 2019{\natexlab{b}}.
\newblock URL \url{http://arxiv.org/abs/1812.02900}.

\bibitem[Goodfellow et~al.(2016)Goodfellow, Bengio, and
  Courville]{earlystop-dlbook2016}
Ian Goodfellow, Yoshua Bengio, and Aaron Courville.
\newblock \emph{Deep Learning}, chapter Regularization for Deep Learning.
\newblock MIT Press, 2016.
\newblock \url{http://www.deeplearningbook.org}.

\bibitem[Gretton et~al.(2012)Gretton, Borgwardt, Rasch, Sch{\"o}lkopf, and
  Smola]{gretton2012kernel}
Arthur Gretton, Karsten~M Borgwardt, Malte~J Rasch, Bernhard Sch{\"o}lkopf, and
  Alexander Smola.
\newblock A kernel two-sample test.
\newblock \emph{Journal of Machine Learning Research}, 13\penalty0
  (Mar):\penalty0 723--773, 2012.

\bibitem[Haarnoja et~al.(2018)Haarnoja, Zhou, Hartikainen, Tucker, Ha, Tan,
  Kumar, Zhu, Gupta, Abbeel, et~al.]{haarnoja2018sacapps}
Tuomas Haarnoja, Aurick Zhou, Kristian Hartikainen, George Tucker, Sehoon Ha,
  Jie Tan, Vikash Kumar, Henry Zhu, Abhishek Gupta, Pieter Abbeel, et~al.
\newblock Soft actor-critic algorithms and applications.
\newblock \emph{arXiv preprint arXiv:1812.05905}, 2018.

\bibitem[Irpan et~al.(2019)Irpan, Rao, Bousmalis, Harris, Ibarz, and
  Levine]{irpan2019off}
Alex Irpan, Kanishka Rao, Konstantinos Bousmalis, Chris Harris, Julian Ibarz,
  and Sergey Levine.
\newblock Off-policy evaluation via off-policy classification.
\newblock \emph{arXiv preprint arXiv:1906.01624}, 2019.

\bibitem[Jaques et~al.(2019)Jaques, Ghandeharioun, Shen, Ferguson, Lapedriza,
  Jones, Gu, and Picard]{jaques2019way}
Natasha Jaques, Asma Ghandeharioun, Judy~Hanwen Shen, Craig Ferguson, Agata
  Lapedriza, Noah Jones, Shixiang Gu, and Rosalind Picard.
\newblock Way off-policy batch deep reinforcement learning of implicit human
  preferences in dialog.
\newblock \emph{arXiv preprint arXiv:1907.00456}, 2019.

\bibitem[Jiang and Li(2015)]{jiang2015doubly}
Nan Jiang and Lihong Li.
\newblock Doubly robust off-policy value evaluation for reinforcement learning.
\newblock \emph{arXiv preprint arXiv:1511.03722}, 2015.

\bibitem[Kingma and Ba(2014)]{kingma2014adam}
Diederik~P Kingma and Jimmy Ba.
\newblock Adam: A method for stochastic optimization.
\newblock \emph{arXiv preprint arXiv:1412.6980}, 2014.

\bibitem[Kumar et~al.(2019)Kumar, Fu, Tucker, and Levine]{kumar2019bear}
Aviral Kumar, Justin Fu, George Tucker, and Sergey Levine.
\newblock Stabilizing off-policy q-learning via bootstrapping error reduction.
\newblock \emph{arXiv preprint arXiv:1906.00949}, 2019.
\newblock URL \url{http://arxiv.org/abs/1906.00949}.

\bibitem[Lange et~al.(2012)Lange, Gabel, and Riedmiller]{lange2012batch}
Sascha Lange, Thomas Gabel, and Martin Riedmiller.
\newblock Batch reinforcement learning.
\newblock In \emph{Reinforcement learning}, pages 45--73. Springer, 2012.

\bibitem[Levine et~al.(2020)Levine, Kumar, Tucker, and Fu]{levine2020offline}
Sergey Levine, Aviral Kumar, George Tucker, and Justin Fu.
\newblock Offline reinforcement learning: Tutorial, review, and perspectives on
  open problems.
\newblock \emph{arXiv preprint arXiv:2005.01643}, 2020.

\bibitem[Lillicrap et~al.(2015)Lillicrap, Hunt, Pritzel, Heess, Erez, Tassa,
  Silver, and Wierstra]{lillicrap2015ddpg}
Timothy~P Lillicrap, Jonathan~J Hunt, Alexander Pritzel, Nicolas Heess, Tom
  Erez, Yuval Tassa, David Silver, and Daan Wierstra.
\newblock Continuous control with deep reinforcement learning.
\newblock \emph{arXiv preprint arXiv:1509.02971}, 2015.

\bibitem[Lin(1992)]{lin1992self}
Long-Ji Lin.
\newblock Self-improving reactive agents based on reinforcement learning,
  planning and teaching.
\newblock \emph{Machine learning}, 8\penalty0 (3-4):\penalty0 293--321, 1992.

\bibitem[Luenberger and Ye(2008)]{David-linear-programming}
David~G. Luenberger and Yinyu Ye.
\newblock \emph{Linear and Nonlinear Programming}, chapter Penalty and Barrier
  Methods.
\newblock Springer, 2008.

\bibitem[Oh et~al.(2018)Oh, Guo, Singh, and Lee]{oh2018self}
Junhyuk Oh, Yijie Guo, Satinder Singh, and Honglak Lee.
\newblock Self-imitation learning.
\newblock \emph{arXiv preprint arXiv:1806.05635}, 2018.

\bibitem[Peng et~al.(2019)Peng, Kumar, Zhang, and Levine]{peng2019awr}
Xue~Bin Peng, Aviral Kumar, Grace Zhang, and Sergey Levine.
\newblock Advantage-weighted regression: Simple and scalable off-policy
  reinforcement learning.
\newblock \emph{arXiv preprint arXiv:1910.00177}, 2019.

\bibitem[Schulman et~al.(2015)Schulman, Levine, Abbeel, Jordan, and
  Moritz]{schulman2015trpo}
John Schulman, Sergey Levine, Pieter Abbeel, Michael Jordan, and Philipp
  Moritz.
\newblock Trust region policy optimization.
\newblock In \emph{International Conference on Machine Learning}, pages
  1889--1897, 2015.

\bibitem[Schulman et~al.(2017)Schulman, Wolski, Dhariwal, Radford, and
  Klimov]{schulman2017ppo}
John Schulman, Filip Wolski, Prafulla Dhariwal, Alec Radford, and Oleg Klimov.
\newblock Proximal policy optimization algorithms.
\newblock \emph{arXiv preprint arXiv:1707.06347}, 2017.

\bibitem[Strehl et~al.(2010)Strehl, Langford, Li, and
  Kakade]{strehl2010learning}
Alex Strehl, John Langford, Lihong Li, and Sham~M Kakade.
\newblock Learning from logged implicit exploration data.
\newblock In \emph{Advances in Neural Information Processing Systems}, pages
  2217--2225, 2010.

\bibitem[Sutton and Barto(2018)]{sutton2018reinforcement}
Richard~S Sutton and Andrew~G Barto.
\newblock \emph{Reinforcement learning: An introduction}.
\newblock MIT press, 2018.

\bibitem[Swaminathan and Joachims(2015)]{swaminathan2015batch}
Adith Swaminathan and Thorsten Joachims.
\newblock Batch learning from logged bandit feedback through counterfactual
  risk minimization.
\newblock \emph{Journal of Machine Learning Research}, 16\penalty0
  (1):\penalty0 1731--1755, 2015.

\bibitem[Thomas and Brunskill(2016)]{thomas2016data}
Philip Thomas and Emma Brunskill.
\newblock Data-efficient off-policy policy evaluation for reinforcement
  learning.
\newblock In \emph{International Conference on Machine Learning}, pages
  2139--2148, 2016.

\bibitem[Van~Hasselt et~al.(2016)Van~Hasselt, Guez, and Silver]{van2016ddqn}
Hado Van~Hasselt, Arthur Guez, and David Silver.
\newblock Deep reinforcement learning with double q-learning.
\newblock In \emph{AAAI}, volume~2, page~5. Phoenix, AZ, 2016.

\bibitem[Vinyals et~al.(2019)Vinyals, Babuschkin, Czarnecki, Mathieu, Dudzik,
  Chung, Choi, Powell, Ewalds, Georgiev, et~al.]{vinyals2019alphastar}
Oriol Vinyals, Igor Babuschkin, Wojciech~M Czarnecki, Micha{\"e}l Mathieu,
  Andrew Dudzik, Junyoung Chung, David~H Choi, Richard Powell, Timo Ewalds,
  Petko Georgiev, et~al.
\newblock Grandmaster level in starcraft ii using multi-agent reinforcement
  learning.
\newblock \emph{Nature}, 575\penalty0 (7782):\penalty0 350--354, 2019.

\bibitem[Vuong et~al.(2018)Vuong, Zhang, and Ross]{vuong2018spu}
Quan Vuong, Yiming Zhang, and Keith~W Ross.
\newblock Supervised policy update for deep reinforcement learning.
\newblock \emph{arXiv preprint arXiv:1805.11706}, 2018.

\bibitem[Wang et~al.(2020)Wang, Wu, Vuong, and Ross]{wang2019towards}
Che Wang, Yanqiu Wu, Quan Vuong, and Keith Ross.
\newblock Towards simplicity in deep reinforcement learning: Streamlined
  off-policy learning.
\newblock \emph{ICML}, 2020.

\bibitem[Wang et~al.(2018)Wang, Xiong, Han, Liu, Zhang, et~al.]{wang2018marwil}
Qing Wang, Jiechao Xiong, Lei Han, Han Liu, Tong Zhang, et~al.
\newblock Exponentially weighted imitation learning for batched historical
  data.
\newblock In \emph{Advances in Neural Information Processing Systems}, pages
  6288--6297, 2018.

\bibitem[Wu et~al.(2019)Wu, Tucker, and Nachum]{wu2019behavior}
Yifan Wu, George Tucker, and Ofir Nachum.
\newblock Behavior regularized offline reinforcement learning.
\newblock \emph{arXiv preprint arXiv:1911.11361}, 2019.

\bibitem[Zhang et~al.(2017)Zhang, Bengio, Hardt, Recht, and
  Vinyals]{zhang2017uat}
Chiyuan Zhang, Samy Bengio, Moritz Hardt, Benjamin Recht, and Oriol Vinyals.
\newblock Understanding deep learning requires rethinking generalization.
\newblock \emph{ICLR}, 2017.

\end{thebibliography}

\clearpage
\appendix
\section{Proofs of Theorems}
\subsection{Proof of Theorem 4.1}
\begin{proof}
Part $(1)$: For any $\lambda \geq 0$ and $\phi=(w, b)$ define
\begin{equation}
\label{objective}
 J^\lambda(\phi) = \sum_{i=1}^m [V_{\phi}(s_i)-G_i]^2+\lambda\|w\|^2
\end{equation}
Note that for any $\phi$ of the form $\phi = (0,b)$, we have $V_{(0,b)}(s) = b$ for all $s$ and $J^{\lambda}(0,b) = \sum_{i=1}^m (b- G_i)^2$ for all $\lambda \geq 0$. 
Also define
\begin{displaymath}
G^* := \max \limits_{1\le i \le m} \{G_i \}
\end{displaymath}
and $\hat{\phi} = (0,G^*)$.
Note that $\hat{\phi}$ is feasible for the constrained optimization problem. It therefore follows that for any $\lambda \geq 0$:
\begin{equation}
\label{H_bound}
J^\lambda(\phi^\lambda) \leq J^\lambda(\hat{\phi}) = \sum \limits_{i=1}^m (G^* - G_i)^2 := H^*
\end{equation}

We first show that $\lim \limits_{\lambda \rightarrow \infty} w^\lambda = 0$. To proceed with a proof by contradiction, assume that this is not true.
There then exists an $\epsilon > 0$ such that for 
any $\lambda \geq 0$ there exists some $\lambda' \ge \lambda$ such that $\|w^{\lambda'}\|^2 > \epsilon$. Choosing $\lambda =  H^*/\epsilon$, we have for some $\lambda'$:
\begin{equation}
    J^{\lambda'}(\phi^{\lambda'}) \ge \lambda'\|w^{\lambda'}\|^2 > \lambda \cdot \epsilon = H^*
\end{equation}
But this contradicts (\ref{H_bound}), establishing 
$\lim \limits_{\lambda \rightarrow \infty} w^\lambda = 0$. 

Next, we show $\lim \limits_{\lambda \rightarrow \infty} b^\lambda 
= G^*$. To prove this, we will show $\bar{b} := \limsup \limits_{\lambda \rightarrow \infty}b^\lambda = G^*$ and also $\tilde{b} := \liminf \limits_{\lambda \rightarrow \infty} b^\lambda = G^*$. First, consider a subsequence $\{b^{\lambda_n}\}$ such that $\lim \limits_{n \rightarrow \infty}b^{\lambda_n} = \tilde{b}$. 
Due to the continuity of $\phi \rightarrow V_{\phi}(s)$ and $\lim \limits_{\lambda \rightarrow \infty}w^\lambda = 0$, we have
\begin{equation}
\label{nn_limit1}
    \lim_{n \rightarrow \infty}V_{\phi^{\lambda^n}}(s) = V_{(0,\tilde{b})}(s) = \tilde{b} \qquad \forall{s}
\end{equation}
Moreover, since $\phi^{\lambda^n}$ has to satisfy the constraints, we also have
\begin{equation}
\label{constrain1}
    V_{\phi^{\lambda^n}}(s_j) \ge G_j = G^*
\end{equation}
where $j = \argmax \limits_i G_i$.
Therefore, combining (\ref{nn_limit1}) and (\ref{constrain1}) yields
\begin{equation}
\label{liminf}
    \liminf_{\lambda \rightarrow \infty}b^\lambda \ge G^*
\end{equation}

Similarly, consider another subsequence $\{b^{\lambda_k}\}$ such that $\lim \limits_{k \rightarrow \infty} b^{\lambda_k} = \bar{b}$.
Again, we have
\begin{equation}
\label{nn_limit2}
    \lim_{k \rightarrow \infty} V_{\phi^{\lambda_k}}(s) = 
    \bar{b}
    \qquad \forall{s}
\end{equation}
We have from (\ref{H_bound}) that
\begin{equation}
    J^{\lambda_k}(\phi^{\lambda_k}) \leq H^*
\end{equation}
Letting $k \rightarrow \infty$ gives
\begin{equation}
    \sum_{i=1}^m 
    (\bar{b} - G_i)^2
    \le \sum_{i=1}^m (G^*-G_i)^2
\end{equation}
which implies that
\begin{equation}
\label{limsup}
    \bar{b} = \limsup_{\lambda \rightarrow \infty} b^\lambda \le G^*
\end{equation}
Therefore, combining (\ref{liminf}) and (\ref{limsup}) together, we have finally shown that $\lim \limits_{\lambda \rightarrow \infty} b^\lambda = G^*$. As we have previously shown $\lim \limits_{\lambda \rightarrow \infty}w^\lambda = 0$, it follows that
\begin{equation}
    \lim_{\lambda \rightarrow \infty} V_{\phi^\lambda}(s) = \max_{1 \le i \le m}\{G_i\}, \qquad \forall{s}
\end{equation}

Part $(2)$: For the case of $\lambda=0$, notice that we  have finitely many inputs $s_i$ to feed into the neural network. Therefore, this is a typical problem regarding the {\em finite-sample expressivity} of neural networks, and the proof directly follows from the work done in  \cite{zhang2017uat}.
\end{proof}

\subsection{Proof of Theorem 4.2}
\begin{proof}
Let $J^\lambda(\phi) = \sum_{i=1}^m [V_{\phi}(s_i)-G_i]^2+\lambda\|w\|^2$ be the loss function that defines the $\lambda$-regularized upper envelope. We notice that the penalty loss function $L^K(\phi)$ takes the form:
\begin{equation}
    L^K(\phi) = J^\lambda(\phi)+(K-1) \sum_{i=1}^m  (\max\{0, G_i-V_{\phi}(s_i)\})^2
\end{equation}
The theorem then directly follows from the standard convergence theorem for penalty functions \cite{David-linear-programming}.
\end{proof}

\newpage
\section{Algorithmic Implementation}

\subsection{Pseudo-Code and Early Stopping Scheme for Upper Envelope Training}

BAIL includes a regularization scheme to prevent over-fitting when generating the upper envelope. We refer to it as an ``early stopping scheme'' because the key idea is to return to the parameter values which gave the lowest validation error (see Section 7.8 of \citet{earlystop-dlbook2016}). In our implementation, we initialize two upper envelope networks with parameters $\phi$ and $\phi^\prime$, where  $\phi$ is trained using the penalty loss, and $\phi^\prime$  records the parameters with the lowest validation error encountered so far. The procedure is done as follows: After every epoch, we calculate the validation loss $L_\phi$ as the penalty loss over all the data in the validation set $\mathcal{B}_v$. We compare this validation loss $L_\phi$ to $L_{\phi^\prime}$, which is the minimum validation loss encountered so far (throughout the history of training). If $L_\phi < L_{\phi^\prime}$, we set $\phi^\prime \leftarrow \phi$. If $L_\phi > L_{\phi^\prime}$, we count the number of consecutive times this occurs.  
The training parameters $\phi$ are returned to $\phi^\prime$ once there are $C$ consecutive times with $L_\phi > L_{\phi^\prime}$. We use $C=4$ in practice.

\begin{algorithm}[!b]
\caption{BAIL}
\label{alg:static_bail_complete}
\begin{algorithmic}
    \STATE Initialize upper envelope parameters $\phi,\phi^\prime$, 
    policy parameters $\theta$. Obtain batch data $\mathcal{B}$. 
    Randomly split data into training set $\mathcal{B}_t$ and validation set $\mathcal{B}_v$ for the upper envelope.
    \STATE Compute return $G_i$ for each data point $i$ in $\mathcal{B}$.
    \STATE Obtain upper envelope by minimizing the loss $L^K(\phi)$:
    \FOR{$j=1,\dots,J$}
    \STATE Sample a mini-batch  $B$ from $\mathcal{B}$.
    \STATE Update $\phi$ using the gradient: 
    $\nabla_{\phi} \sum_{i \in B} ( V_{\phi}(s_i) - G_i )^2$ $\{ \mathbbm{1}_{(V_{\phi}(s_i) >  G_i)} + K \mathbbm{1}_{(V_{\phi}(s_i) <  G_i)} \} + \lambda \|\phi\|^2$
    \IF {time to do validation for the upper envelope}
    \STATE Compute validation loss on $B_v$
    \STATE Update $\phi$ and $\phi^\prime$ according to the validation loss
    \ENDIF
    \ENDFOR
    \STATE Select data point $i$ if $G_i > x V_\phi(s_i)$, where $x$ is such that $p\%$ of data in $\mathcal{B}$ are selected. Let $\mathcal{U}$ be the set of selected data points.
    \FOR{$l=1,\dots,L$}
    \STATE Sample a mini-batch $U$ of data from $\mathcal{U}$.
    \STATE Update $\theta$ using the gradient: 
    $\nabla_{\theta} \sum_{i \in U} (\pi_\theta(s_i) - a_i )^2$
    \ENDFOR
\end{algorithmic}
\end{algorithm}

\begin{algorithm}[!b]
\caption{Progressive BAIL}
\label{alg:progressive_bail_complete}
\begin{algorithmic}
    \STATE Initialize upper envelope parameters $\phi, \phi^\prime$, policy parameters $\theta$. 
    \STATE Obtain batch data $\mathcal{B}$. Randomly split data into training set $\mathcal{B}_t$ and validation set $\mathcal{B}_v$ for the upper envelope.
    \STATE Compute return $G_i$ for each data point $i$ in $\mathcal{B}$.
    \FOR{$l=1,\dots,L$}
    \STATE Sample a mini-batch of data $B$ from the batch $\mathcal{B}_t$.
    \STATE Update $\phi$ using the gradient: 
    $\nabla_{\phi} \sum_{i \in B_t} ( V_{\phi}(s_i) - G_i )^2$ $\{ \mathbbm{1}_{(V_{\phi}(s_i) >  G_i)} + K \mathbbm{1}_{(V_{\phi}(s_i) <  G_i)} \} + \lambda \|\phi\|^2$
    \IF {time to validate}
    \STATE Compute validation loss on $B_v$
    \STATE Update $\phi$ and $\phi^\prime$ according to validation loss
    \ENDIF
    \STATE Select data point $i$ if $G_i > x V_\phi(s_i)$, where $x$ is such that $p\%$ of data in $B$ are selected. Let $U$ be the set of selected data points.
    \STATE Update $\theta$ using the gradient: 
    $\nabla_{\theta} \sum_{i \in U} (\pi_\theta(s_i) - a_i )^2$
    \ENDFOR
\end{algorithmic}
\end{algorithm}

\newpage
The pseudo-code for BAIL and Progressive BAIL, which include the early stopping scheme, are presented in Algorithms  \ref{alg:static_bail_complete} and \ref{alg:progressive_bail_complete}. Note BAIL has two for loops in series, whereas Progressive BAIL has only one for loop. 

\subsection{Hyper-parameters of BAIL}
BAIL and  Progressive BAIL use the same hyper-parameters except for the selection percentage $p$. Details are provided in Table \ref{bail_hyper-table}.

\begin{table}[htbp]
\caption{BAIL hyper-parameters}
\label{bail_hyper-table}
\begin{center}
\begin{tabular}{ll}
\toprule 
\multicolumn{1}{c}{\bf Parameter}  &\multicolumn{1}{c}{\bf Value} \\
\midrule
discount rate $\gamma$    & $0.99$ \\
horizon $T$  & $1000$ \\
training set size   &$0.8 \cdot |\mathcal{B}|$  \\
validation set size   &$0.2 \cdot |\mathcal{B}|$  \\
optimizer         & Adam \citep{kingma2014adam}\\
percentage $p\%$ & 30\% for BAIL  \\
                 & 25\% for Progressive BAIL\\
                 
\hline
\textbf{upper envelope network} \\
structure & $128 \times 128$ hidden units, ReLU activation\\
learning rate        &$3 \cdot 10^{-3}$ \\
penalty loss coefficient $K$ & 1000\\
\hline
\textbf{policy network} \\
structure & $400 \times 300$ hidden units, ReLU activation\\
learning rate        &$1 \cdot 10^{-3}$ \\
\bottomrule
\end{tabular}
\end{center}
\end{table}

\section{Experimental Details}

This paper compares BAIL (our algorithm) with four other baselines: BC, BCQ, BEAR, and MARWIL. We use five MuJoCo environments, including Humanoid, which is the most challenging of the MuJoCo environments, and is not attempted in most other papers on batch DRL. 

\subsection{Hyper-parameter consistency}

When designing RL algorithms, it is desirable that they generalize over new, unforeseen environments and tasks. Therefore, consistent with common practice for online reinforcement learning \citep{schulman2015trpo, schulman2017ppo, vuong2018spu,lillicrap2015ddpg,fujimoto2018td3, wang2019towards}, when evaluating any given algorithm, we use the same hyper-parameters for all environments and all batches.  The BCQ paper \cite{fujimoto2019bcq} also uses the same hyper-parameters for all experiments. 

Alternatively, one could optimize the hyper-parameters for each environment separately. Not only is this not standard practice, but to make a fair comparison across all algorithms, this would require, for {\em each} of the five algorithms, performing a separate hyper-parameter search for {\em each} of the five environments. 

\subsection{Reproduction of the Baseline Algorithms}

In our submission, we went the extra mile to make a fair comparison to other batch RL algorithms. We are therefore confident about properly using the authors' BCQ and BEAR code, and fairly reproducing MARWIL for the MuJocO benchmark.

\textbf{BCQ} We use the authors’ code and recommended hyper-parameters. In the BCQ paper, the “final buffer” batches are where the BCQ algorithm shines the most; therefore, included in our training batches are batches for which we used exactly the same “final buffer” experimental set-up. In our terminology, this corresponds to DDPG training batches with sigma = 0.5. Looking at the BCQ final-buffer results in Figure 2 and Table 1, we see that they are consistent with the results in Figure 2a in the BCQ paper. 

\begin{figure}[ht]
\vskip 0.2in
\begin{center}
\centerline{\includegraphics[width=0.25\columnwidth]{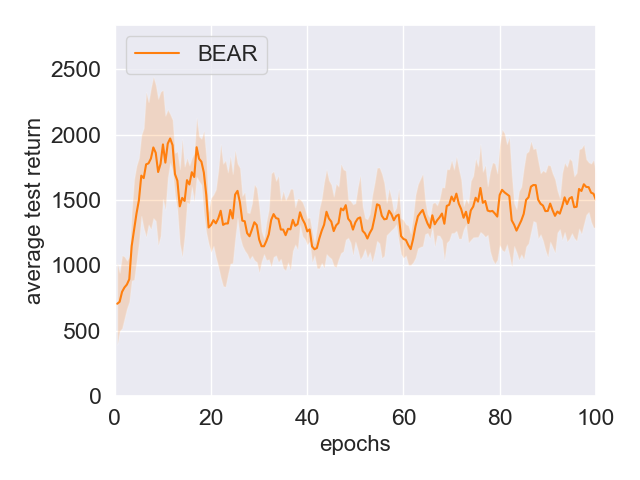}}
\caption{Our results when we apply BEAR to the authors' dataset. This figure matches Figure 3 in \citet{kumar2019bear}.}
\label{fig:bear_reproduce}
\end{center}
\vskip -0.2in
\end{figure}

\textbf{BEAR} To ensure that we are running the BEAR code properly, we obtained a dataset directly from the BEAR authors and ran the BEAR algorithm with a specific set of hyper-parameters among their recommendations. Specifically, we used their version "0” with “use ensemble variance” set to False and employ Laplacian kernel.
The dataset provided by the authors was for Hopper-v2 with mediocre performance. The performance we obtained is shown in Figure \ref{fig:bear_reproduce}, which fully matches the Hopper-v2 case in Figure 3 in \citep{kumar2019bear}. 
Also, we observed that for some of our batches, we obtained very similar results to what is shown in the BEAR paper. 

However, the results shown in Tables 1 and 2 show that BEAR can sometimes have poor performance, much worse than what is shown in \citep{kumar2019bear,wu2019behavior}. This is because in \citep{kumar2019bear,wu2019behavior}, hyper-parameters are optimized separately for each of the MuJoCo environments. In this paper, as discussed above, for each algorithm we use one set of hyper-parameters. In the case of BEAR, we use one of their recommended hyper-parameter settings for all environments and batches, namely, their version "0” with ``use ensemble variance'' set to False and employ the Laplacian kernel.


\textbf{MARWIL} The authors of MARWIL do not provide an open-source implementation of their algorithm. Furthermore, experiments in \cite{wang2018marwil} are carried out on environments like HFO and TORCS which are considerably different from MuJoCo. We replicate all implementation details discussed in MARWIL, except that we use the same 
network architectures used for BCQ, BEAR and BAIL to ensure a fair comparison.
We use the same augmentation heuristic for the returns as we use in BAIL. 
We use the recommended hyper-parameters given by the MARWIL authors. 

\subsection{Common Hyper-parameters across all batch RL algorithms}

\textbf{Network size}
A common feature among all the batch DRL algorithms is that they have a policy neural network. BCQ and BEAR both have an architecture consisting of $400 \times 300$ hidden units with ReLU activation units. We use exactly the same network architecture for the policy network for BAIL and Progressive BAIL. 
For the IL-based algorithms, we also use this same policy network architecture.

\textbf{Learning rate} All algorithms considered in our experiments use the same learning rate of $1 \cdot 10^{-3}$ for the policy network, which is also the default in BCQ and BEAR.

\subsection{Evaluation methodology employed for all batch RL algorithms}
To evaluate the performance of the current policy during training, we run ten episodes of test runs with the current policy and record the average of the returns. This is done with the same frequency for each algorithm considered in our experiments.

For a test episode, we sometimes encounter an error signal from the MuJoCo environment, and thus are not able to continue the episode. 
In these cases, we assign a zero value to the return for the terminated episode.
In Tables 1 and 2 of the paper, there are a few entries with zero mean and zero standard deviation. These zeros are due to repeatedly encountering this error signal for the test runs using different seeds, with each test run getting a zero value for the return. This happens for BEAR in several batches, which is likely because we are not using different hyper-parameters for each environment.

\newpage
\section{Ablation studies for BAIL}
\subsection{Augmented return versus oracle performance}

\begin{figure}[ht]
\vskip 0.1in
\centering
\begin{subfigure}{0.24\columnwidth}
 \centering
 \includegraphics[width=0.99\linewidth]{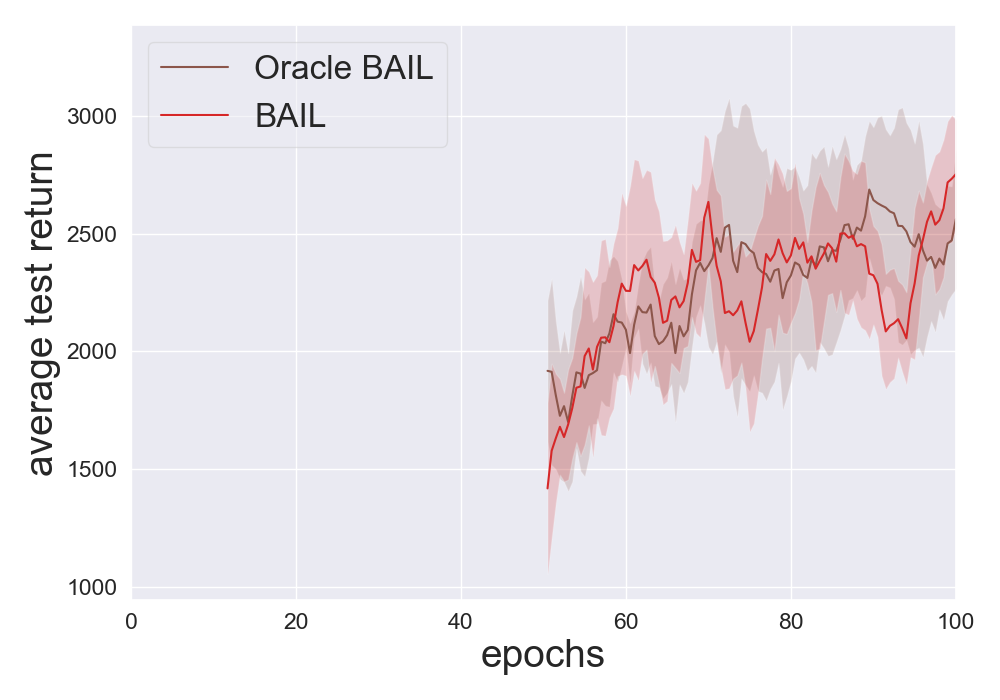}
 \caption{DDPG training batch with $\sigma = 0.5$}
\end{subfigure}
\begin{subfigure}{0.24\columnwidth}
 \centering
 \includegraphics[width=0.99\linewidth]{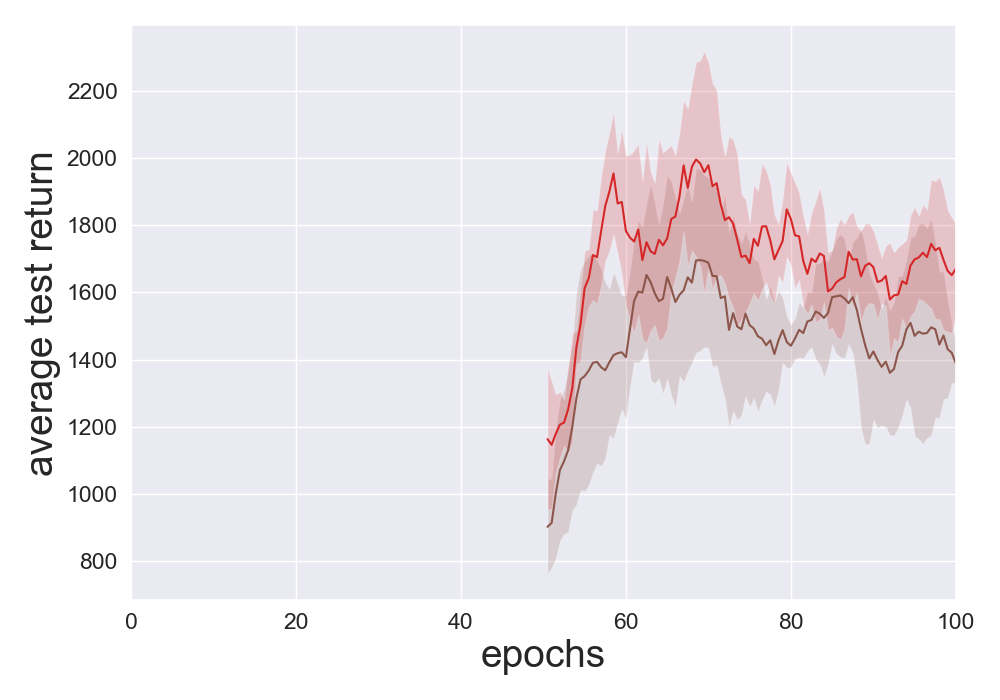}
 \caption{DDPG training batch with $\sigma = 0.1$}
\end{subfigure}
\begin{subfigure}{0.24\columnwidth}
 \centering
 \includegraphics[width=0.99\linewidth]{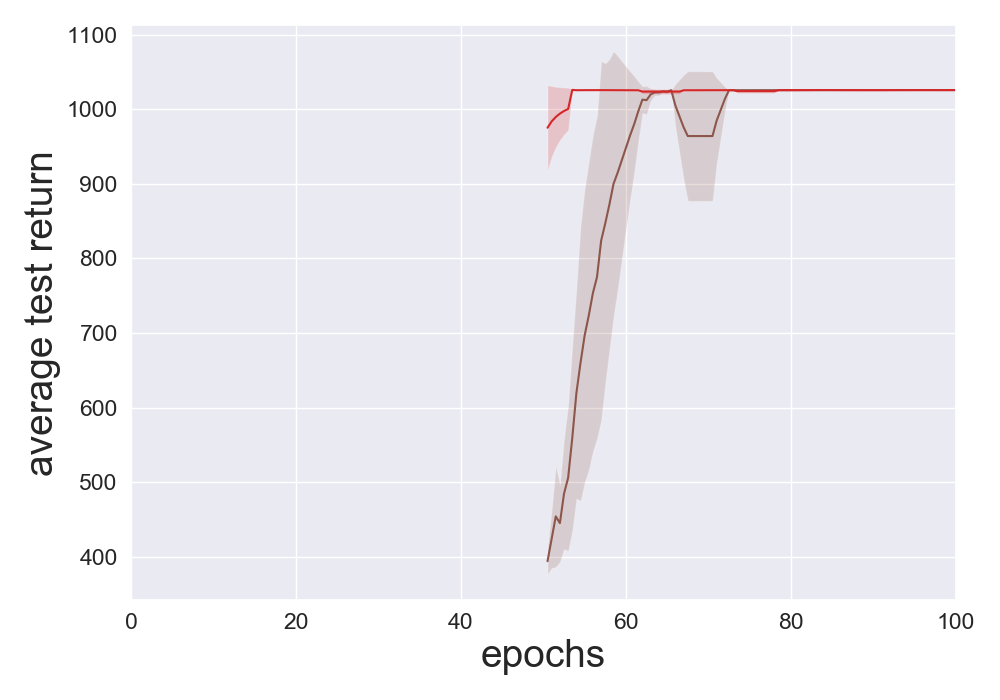}
 \caption{SAC mediocre execution\\ batch with $\sigma = 0$}
\end{subfigure}
\begin{subfigure}{0.24\columnwidth}
 \centering
 \includegraphics[width=0.99\linewidth]{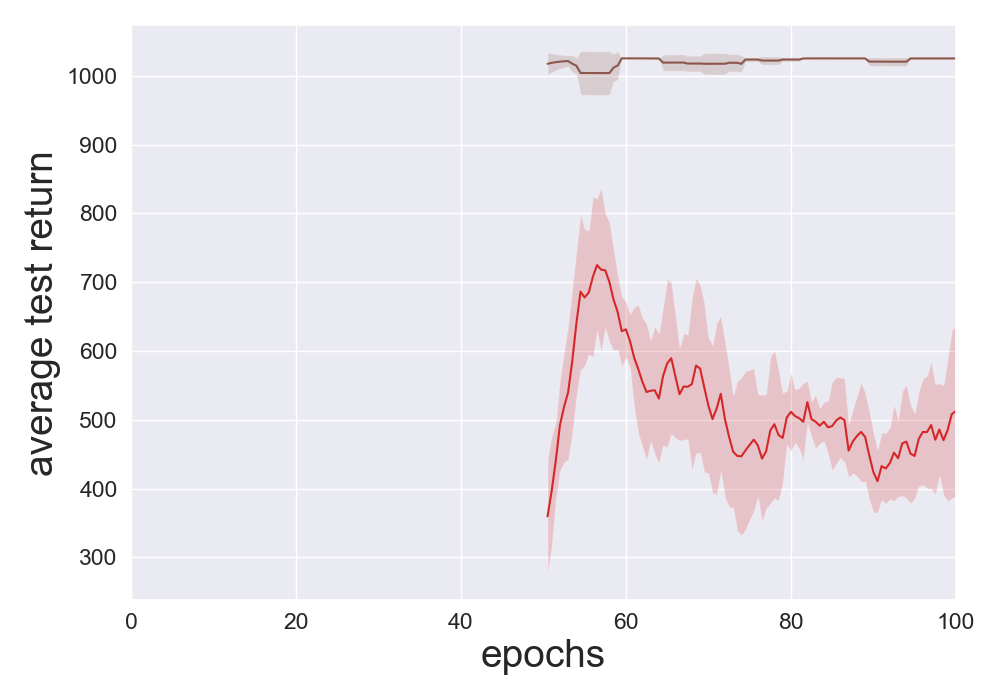}
 \caption{SAC mediocre execution\\ batch with  $\sigma = \sigma(s)$}
\end{subfigure}
\begin{subfigure}{0.24\columnwidth}
 \centering
 \includegraphics[width=0.99\linewidth]{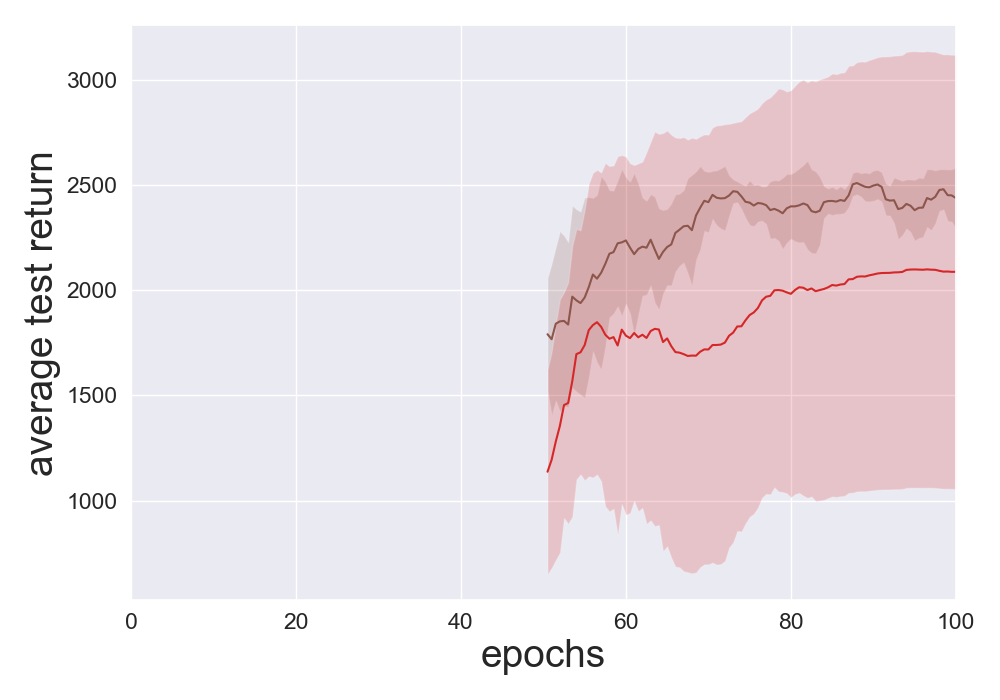}
 \caption{SAC optimal execution\\ batch with $\sigma = 0$}
\end{subfigure}
\begin{subfigure}{0.24\columnwidth}
 \centering
 \includegraphics[width=0.99\linewidth]{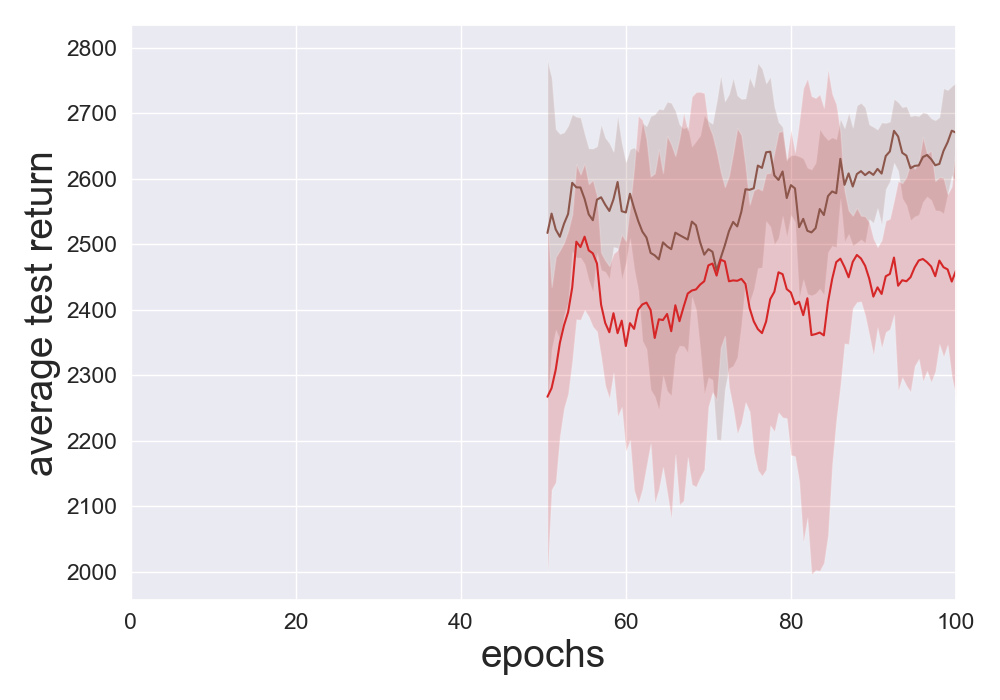}
 \caption{SAC optimal execution\\ batch with $\sigma = \sigma(s)$}
\end{subfigure}
\begin{subfigure}{0.24\columnwidth}
 \centering
 \includegraphics[width=0.99\linewidth]{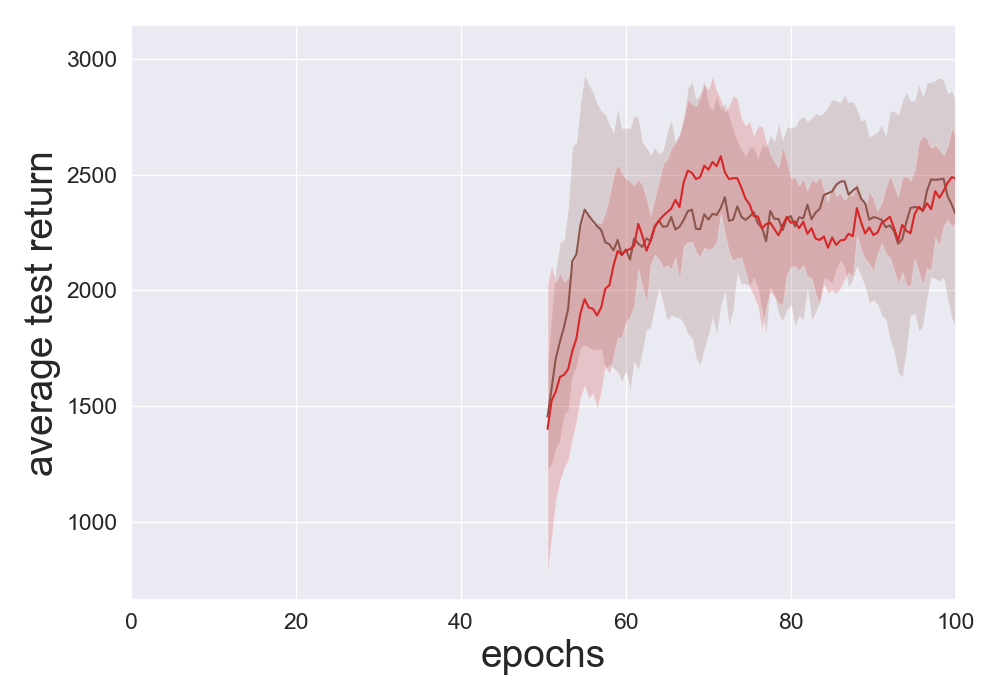}
 \caption{SAC training batch}
\end{subfigure}
\caption{Augmented Returns versus Oracle Performance. All learning curves are for the Hopper-v2 environment. The x-axis ranges from 50 to 100 epochs since this comparison involves only BAIL. The results show that the augmentation heuristic typically achieves oracle-level performance.}
\label{fig:oracles_complete}
\vskip -0.1in
\end{figure}

To validate our heuristic for the augmented returns, we compute oracle returns by letting episodes run up to 2000 time steps. In this manner, every return is calculated with at least 1000 actual rewards, and is therefore essentially exact due to discounting.
Figure \ref{fig:oracles_complete} compares the performance of BAIL using our augmentation heuristic and BAIL using the oracle for Hopper-v2 for seven diverse batches. 
The results show that our augmentation heuristic typically achieves oracle-level performance. We conclude that our augmentation heuristic is a satisfactory method for addressing continual environments such as MuJoCo, which is also confirmed with its good performance shown in Table \ref{execution-table} . 

\newpage
\subsection{Ablation study for data selection}

\begin{figure}[ht]
\vskip 0.1in
\centering
\begin{subfigure}{0.24\columnwidth}
 \centering
 \includegraphics[width=0.99\linewidth]{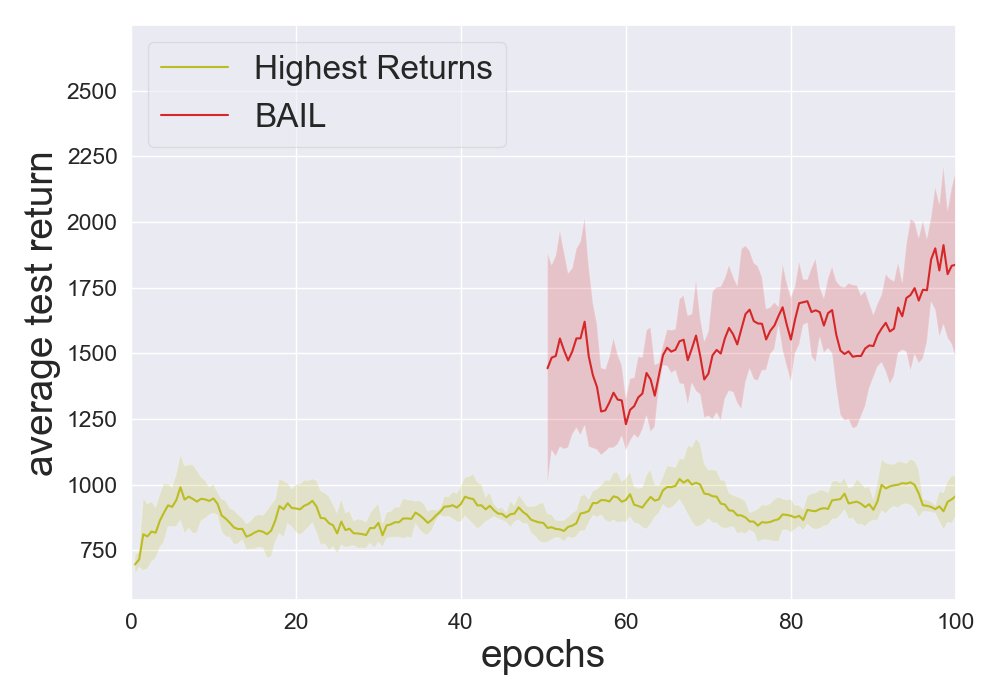}
 \caption{Hopper $\sigma=0.5$ 1st}
\end{subfigure}
\begin{subfigure}{0.24\columnwidth}
 \centering
 \includegraphics[width=0.99\linewidth]{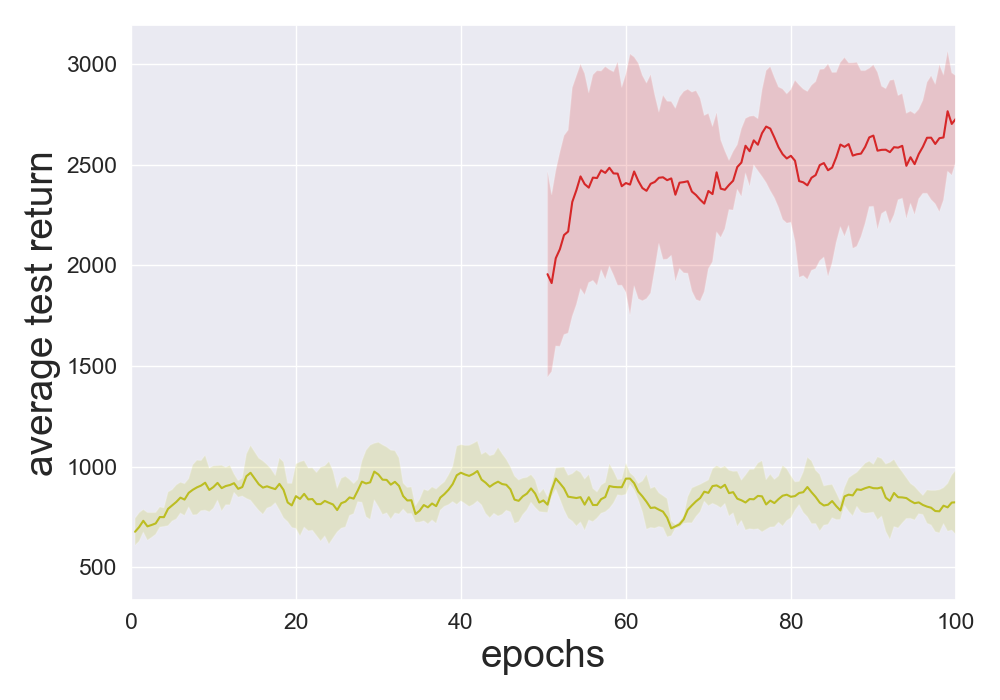}
 \caption{Hopper $\sigma=0.5$ 2nd}
\end{subfigure}
\begin{subfigure}{0.24\columnwidth}
 \centering
 \includegraphics[width=0.99\linewidth]{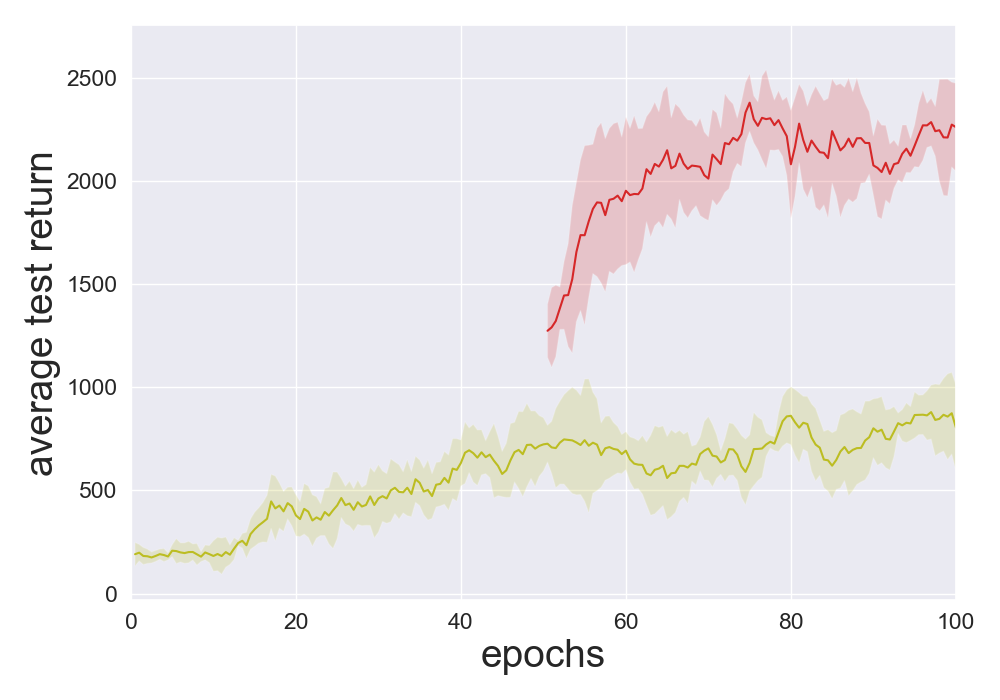}
 \caption{Hopper $\sigma=0.1$ 1st}
\end{subfigure}
\begin{subfigure}{0.24\columnwidth}
 \centering
 \includegraphics[width=0.99\linewidth]{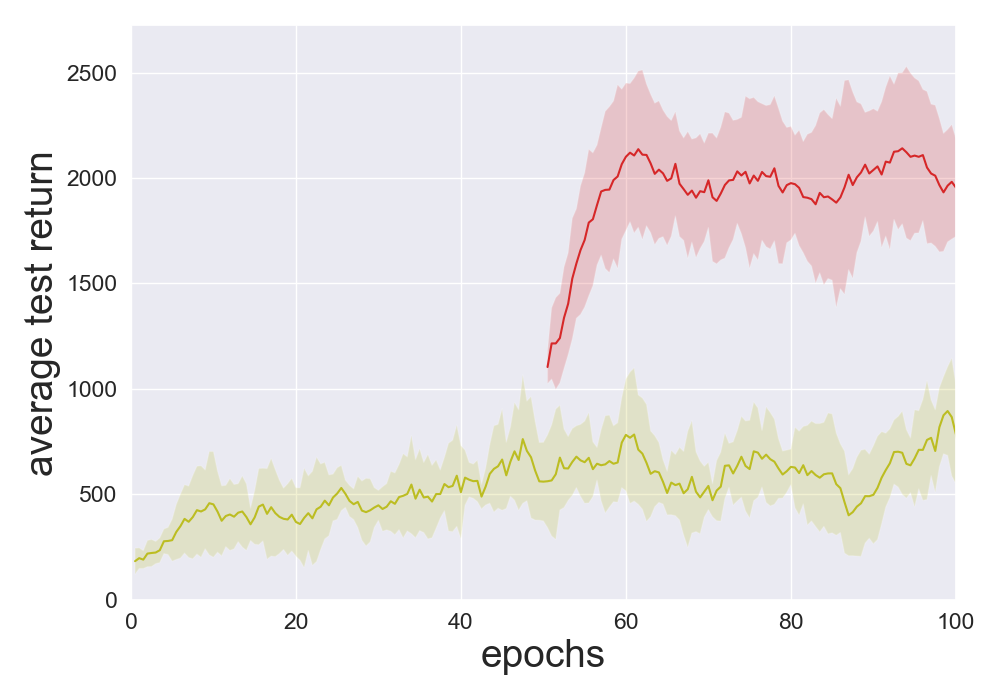}
 \caption{Hopper $\sigma=0.1$ 2nd}
\end{subfigure}
\begin{subfigure}{0.24\columnwidth}
 \centering
 \includegraphics[width=0.99\linewidth]{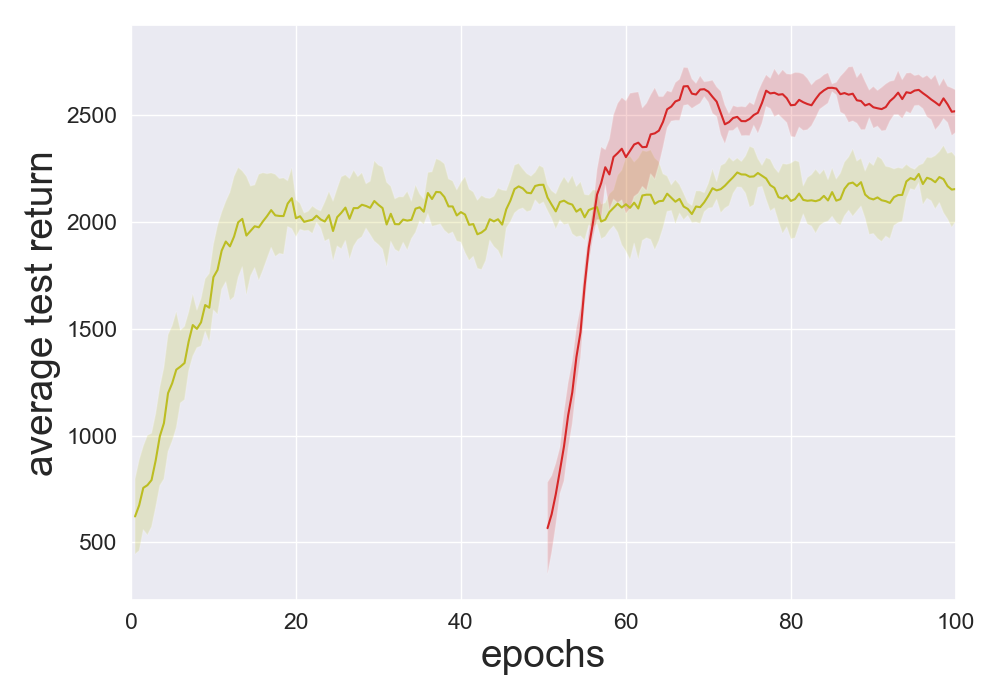}
 \caption{Walker2d $\sigma=0.5$ 1st}
\end{subfigure}
\begin{subfigure}{0.24\columnwidth}
 \centering
 \includegraphics[width=0.99\linewidth]{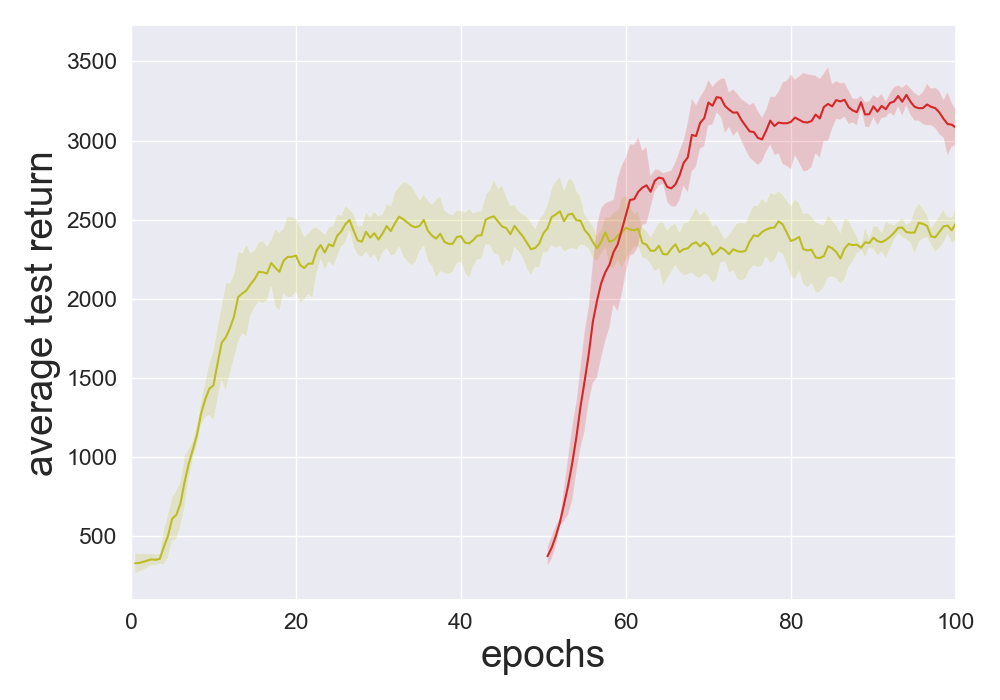}
 \caption{Walker2d $\sigma=0.5$ 2nd}
\end{subfigure}
\begin{subfigure}{0.24\columnwidth}
 \centering
 \includegraphics[width=0.99\linewidth]{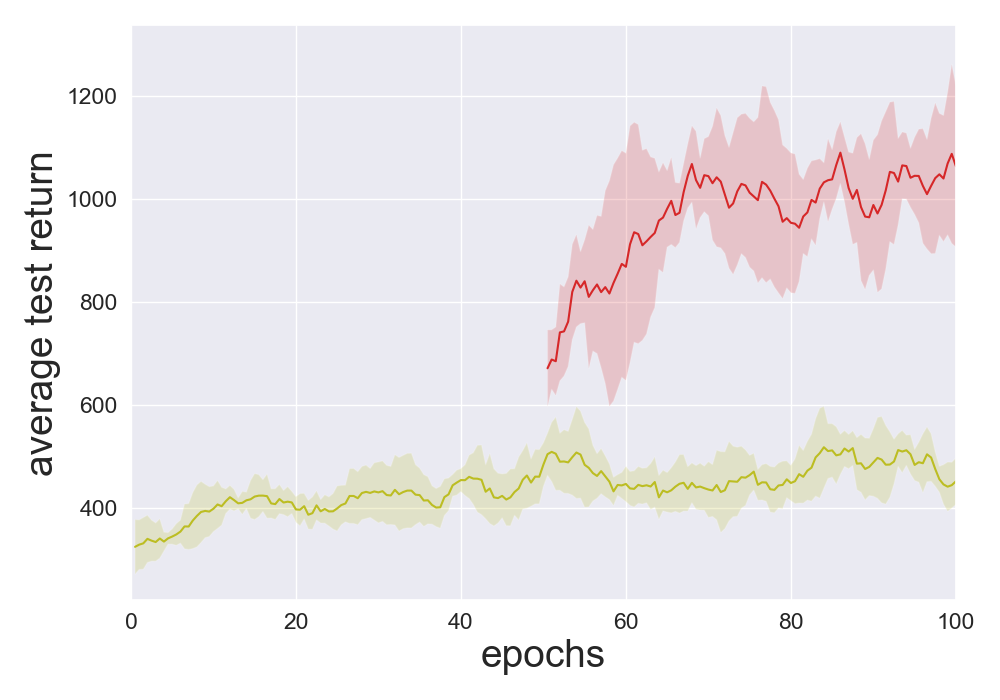}
 \caption{Walker2d $\sigma=0.1$ 1st}
\end{subfigure}
\begin{subfigure}{0.24\columnwidth}
 \centering
 \includegraphics[width=0.99\linewidth]{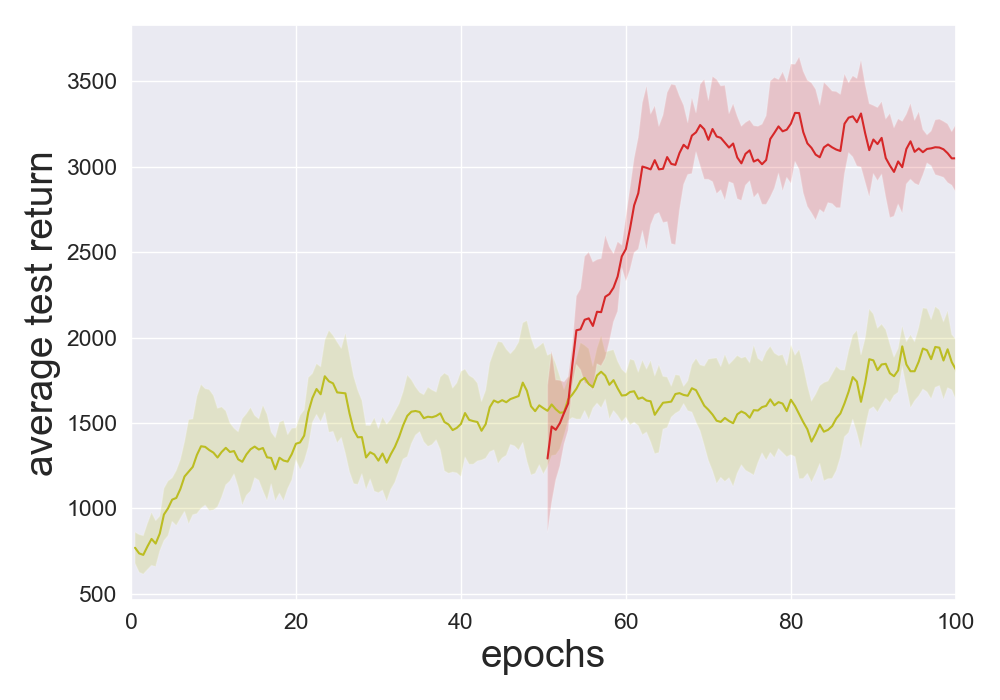}
 \caption{Walker2d $\sigma=0.1$ 2nd}
\end{subfigure}
\begin{subfigure}{0.24\columnwidth}
 \centering
 \includegraphics[width=0.99\linewidth]{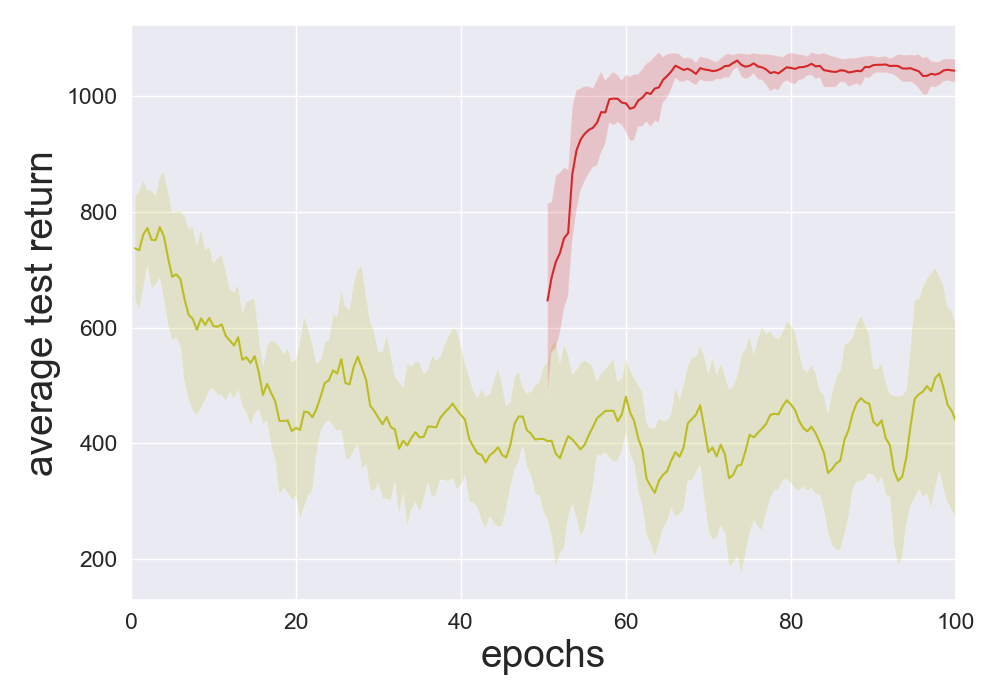}
 \caption{HalfCheetah $\sigma=0.5$ 1st}
\end{subfigure}
\begin{subfigure}{0.24\columnwidth}
 \centering
 \includegraphics[width=0.99\linewidth]{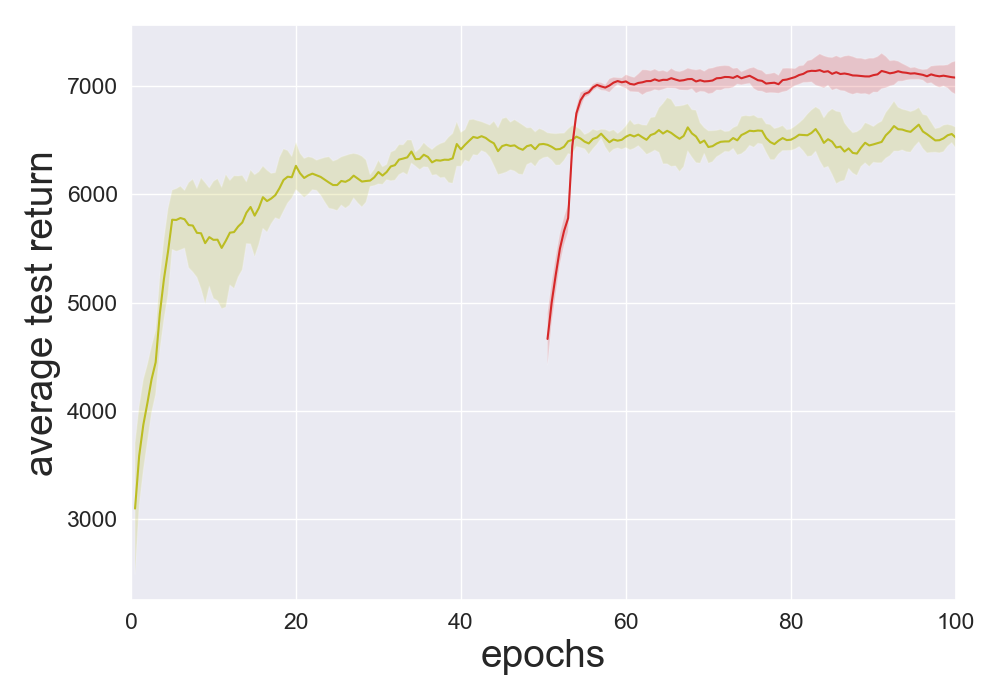}
 \caption{HalfCheetah $\sigma=0.5$ 2nd}
\end{subfigure}
\begin{subfigure}{0.24\columnwidth}
 \centering
 \includegraphics[width=0.99\linewidth]{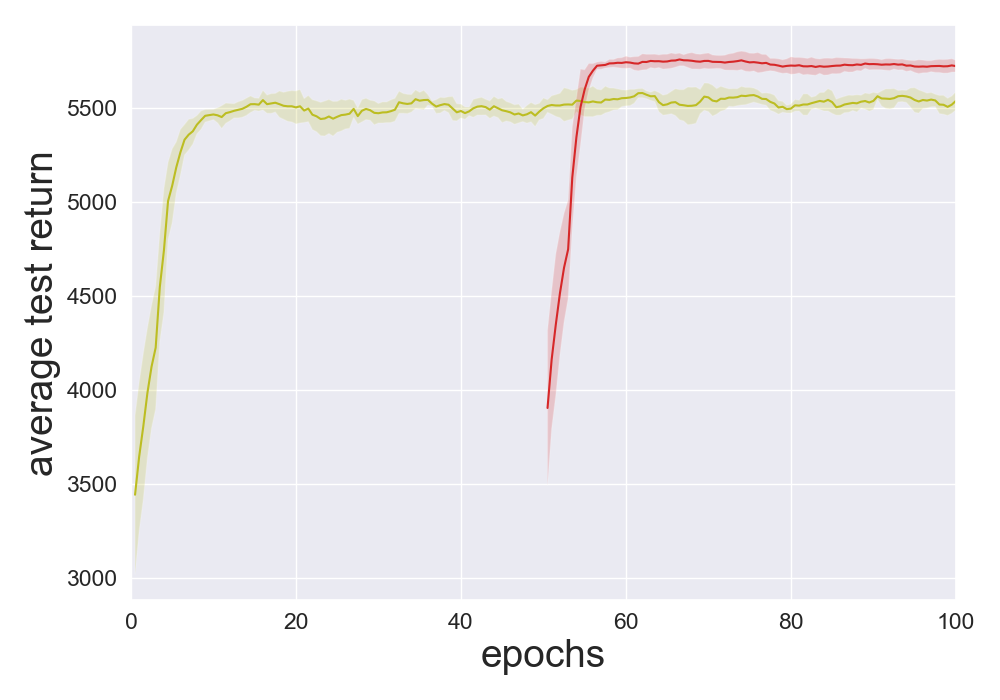}
 \caption{HalfCheetah $\sigma=0.1$ 1st}
\end{subfigure}
\begin{subfigure}{0.24\columnwidth}
 \centering
 \includegraphics[width=0.99\linewidth]{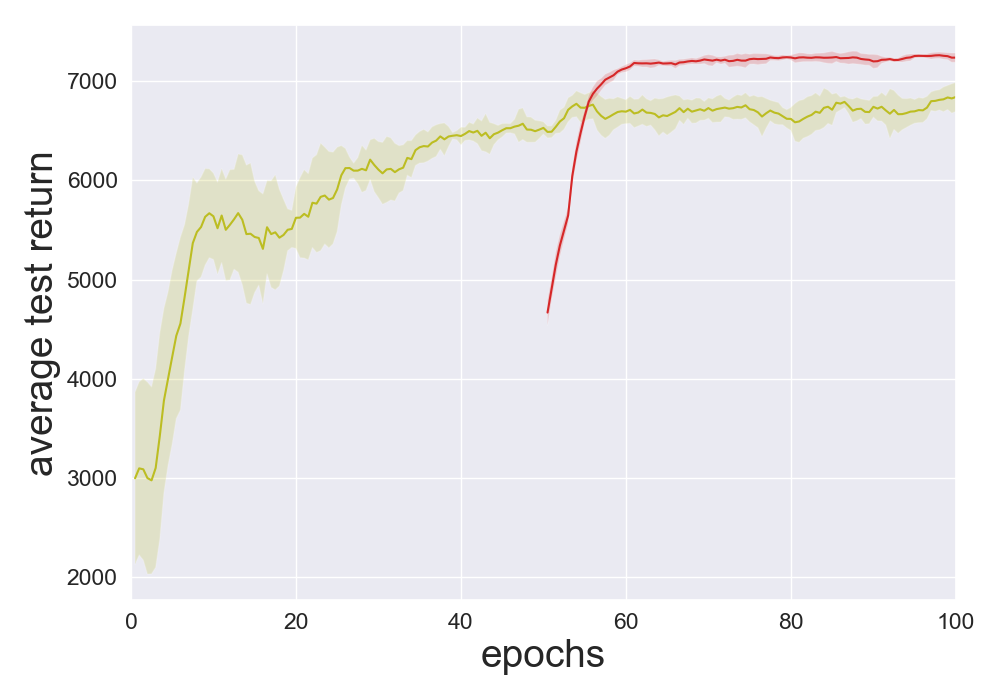}
 \caption{HalfCheetah $\sigma=0.1$ 2nd}
\end{subfigure}
\caption{Ablation study for data selection. The figure compares BAIL with the algorithm that simply chooses the state-action pairs with the highest returns (without using an upper envelope). The learning curves show that the upper envelope is critical components of BAIL. }
\label{fig:ablation_returns_complete}
\vskip -0.1in
\end{figure}

BAIL uses an upper envelope to select the ``best'' data points for training a policy network with imitation learning.  It is natural to ask how BAIL would perform when using the more naive approach of  selecting the best actions by simply selecting the same percentage of data points with the highest $G_i$ values.  
Figure \ref{fig:ablation_returns_complete} compares BAIL with the algorithm that simply chooses the state-action pairs with the highest returns (without using an upper envelope). The learning curves show that the upper envelope is a critical component of BAIL.

\newpage
\subsection{Ablation study using standard regression instead of an upper envelope}

\begin{figure}[ht]
\vskip 0.1in
\centering
\begin{subfigure}{0.24\columnwidth}
 \centering
 \includegraphics[width=0.99\linewidth]{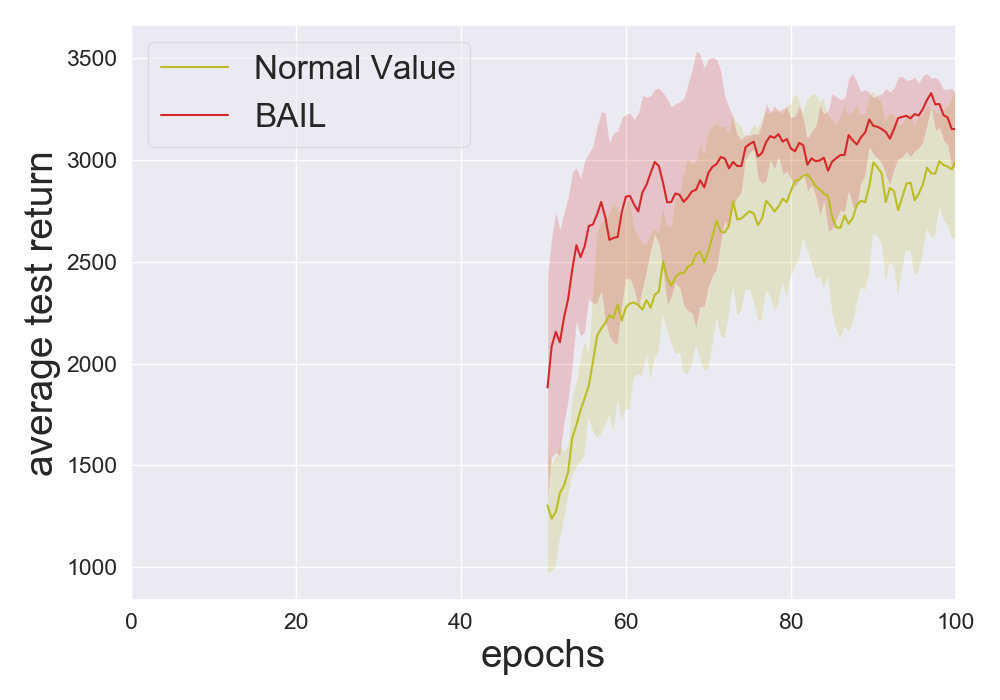}
 \caption{Training SAC,\\ Hopper}
\end{subfigure}
\begin{subfigure}{0.24\columnwidth}
 \centering
 \includegraphics[width=0.99\linewidth]{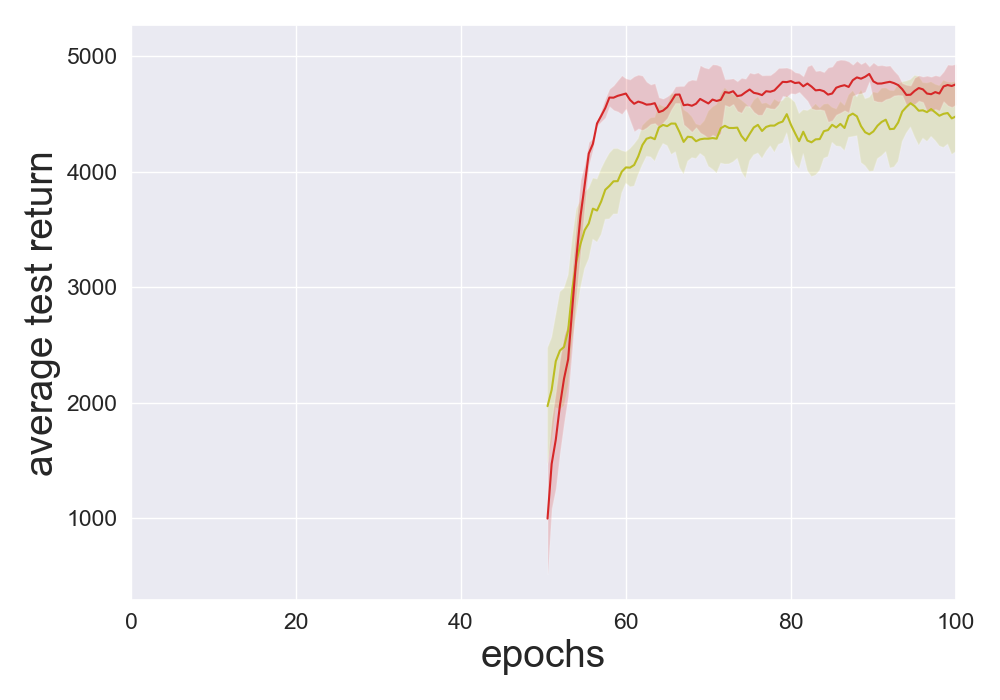}
 \caption{Training SAC,\\ Walker2d}
\end{subfigure}
\begin{subfigure}{0.24\columnwidth}
 \centering
 \includegraphics[width=0.99\linewidth]{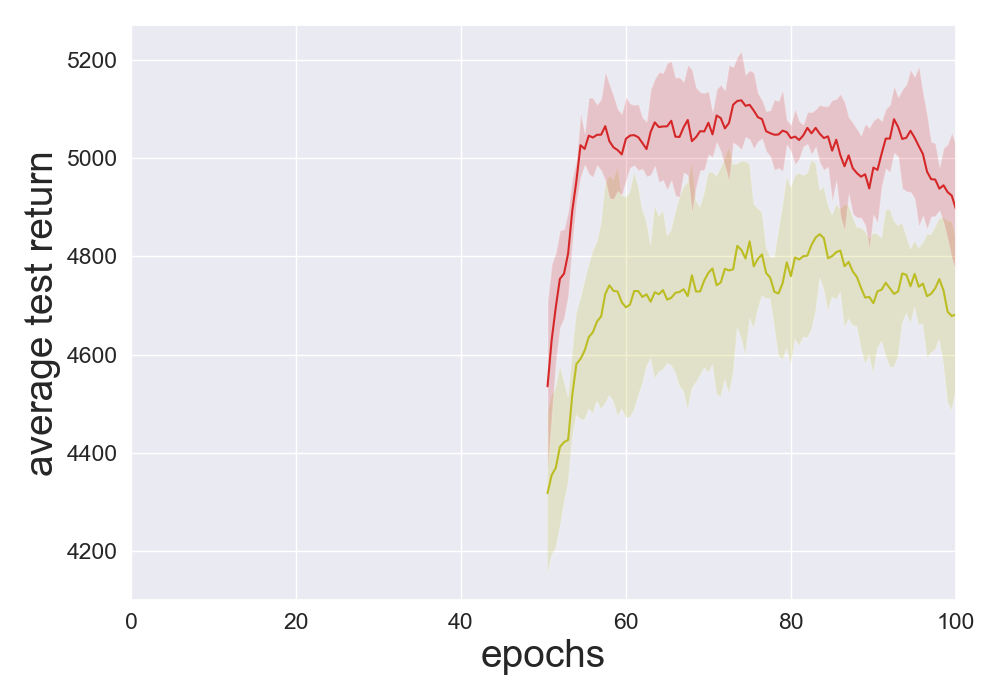}
 \caption{Training SAC,\\ Ant}
\end{subfigure}
\begin{subfigure}{0.24\columnwidth}
 \centering
 \includegraphics[width=0.99\linewidth]{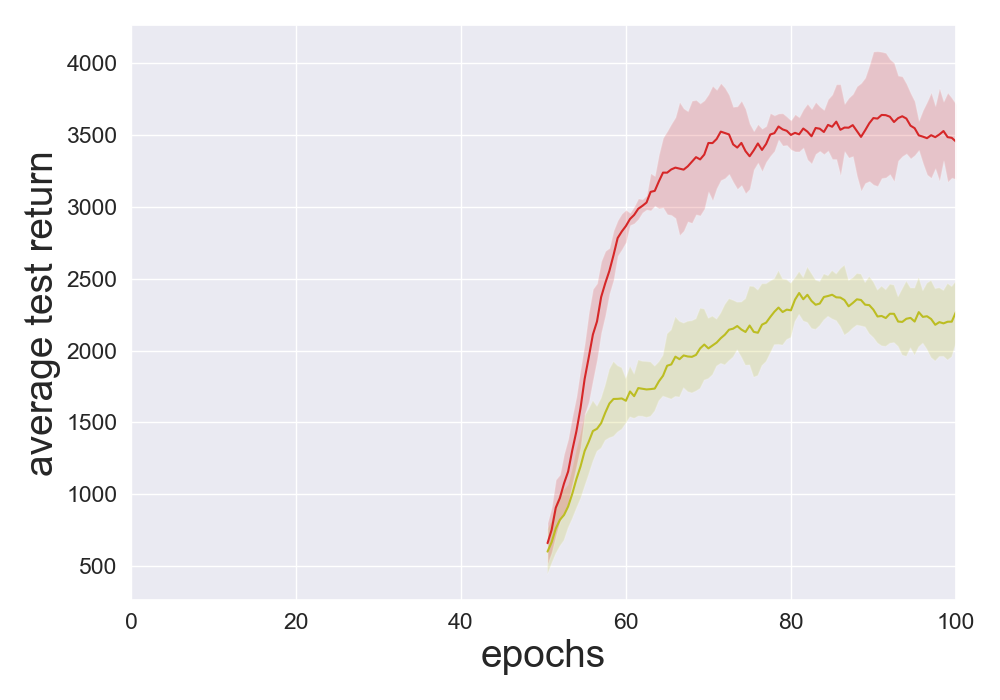}
 \caption{Training SAC,\\ Humanoid}
\end{subfigure}

\begin{subfigure}{0.24\columnwidth}
 \centering
 \includegraphics[width=0.99\linewidth]{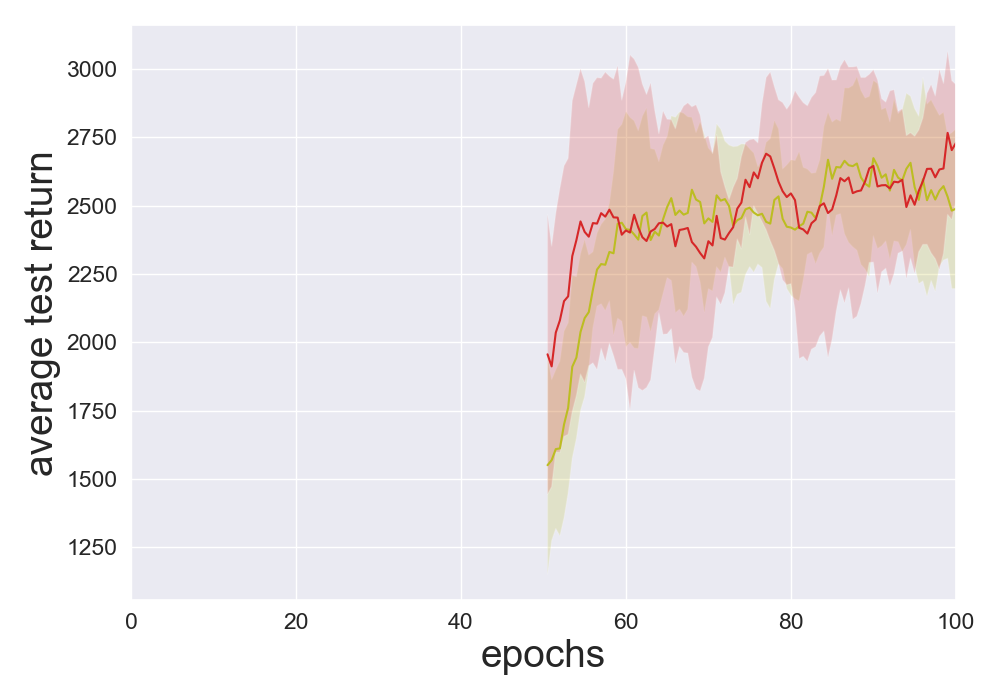}
 \caption{Training DDPG\\ $\sigma=0.5$, Hopper}
\end{subfigure}
\begin{subfigure}{0.24\columnwidth}
 \centering
 \includegraphics[width=0.99\linewidth]{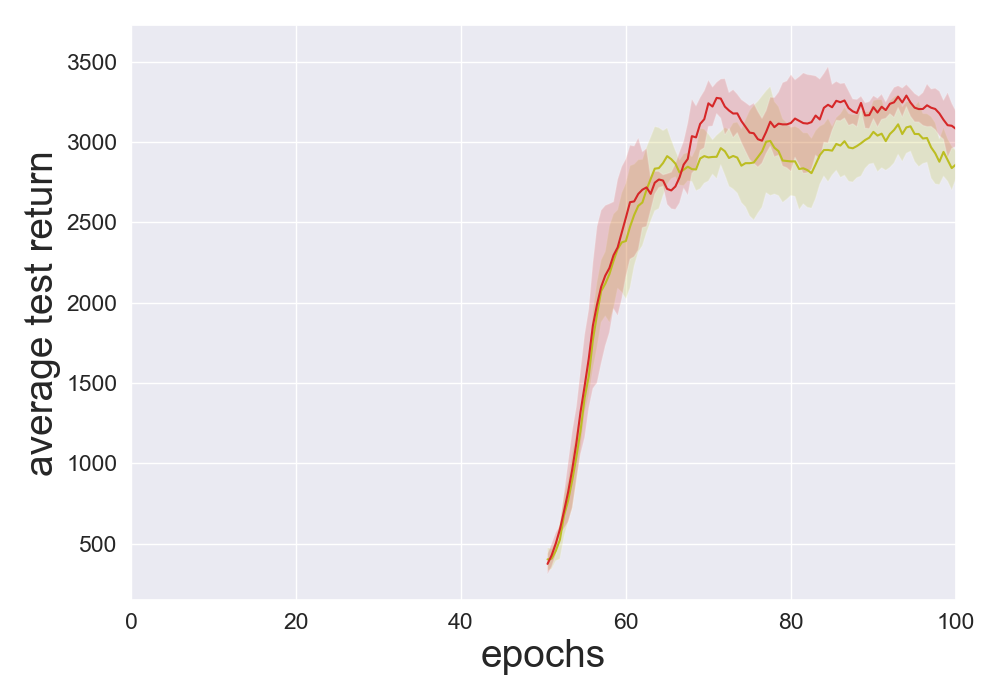}
 \caption{Training DDPG\\ $\sigma=0.5$, Walker2d}
\end{subfigure}
\begin{subfigure}{0.24\columnwidth}
 \centering
 \includegraphics[width=0.99\linewidth]{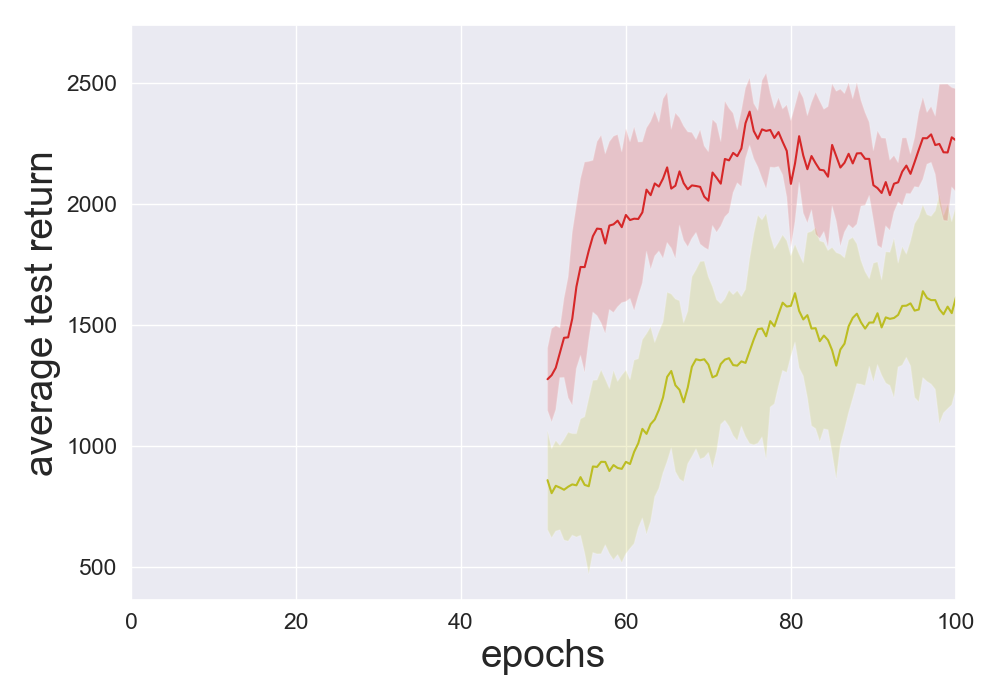}
 \caption{Training DDPG\\ $\sigma=0.1$, Hopper}
\end{subfigure}
\begin{subfigure}{0.24\columnwidth}
 \centering
 \includegraphics[width=0.99\linewidth]{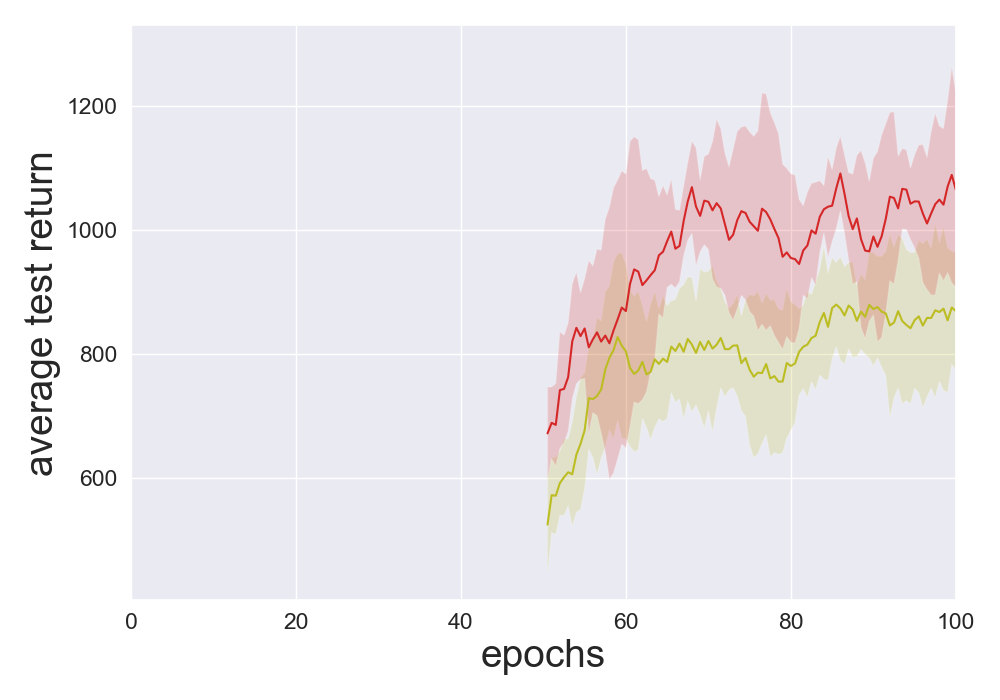}
 \caption{Training DDPG\\ $\sigma=0.1$, Walker2d}
\end{subfigure}
\caption{Ablation study using standard regression instead of an upper envelope. The figure compares BAIL with the more naive scheme of using standard regression in place of an upper envelope.  The learning curves show that the upper envelope is a critical component of BAIL. }
\label{fig:ablation_normalvalue_complete}
\vskip -0.1in
\end{figure}

Figure \ref{fig:ablation_normalvalue_complete} compares BAIL with the more naive scheme of using standard regression in place of an upper envelope.  The learning curves show that the upper envelope is a critical component of BAIL.

\section{Performance for execution batches}

As discussed in the main body of the paper, we also performed experiments for execution batches. Once again, for a given algorithm, we use the same hyper-parameters for all environments and batches (training ane execution). 
We see from Table \ref{execution-table} that 
BC, MARWIL, BAIL, and BCQ have similar overall performance, with BC and MARWIL having the highest number of wins and also being slightly stronger in terms of average performance. MARWIL has one more win compared to BC, but slightly lower average performance. Comparing BAIL and BCQ, BAIL has a slightly stronger average performance score, and BCQ has a few more wins. 

\begin{table*}[t]
\centering
	\caption{Performance of Five Batch DRL Algorithms for 40 different execution datasets.} 
	\label{execution-table}
	\vskip 0.15in
	\begin{center}
		\begin{small}
			\begin{sc}
				\begin{tabular}{lccccr}
					\toprule
					Environment & Bail & BCQ & BEAR & BC & MARWIL\\
                    \midrule
M $\sigma=0$ Hopper B1  & $\bm{1026\pm0}$ & $901\pm132$ & $4\pm1$ & $\bm{1026\pm0}$ & $\bm{1026\pm0}$ \\
M $\sigma=0$ Hopper B2  & $696\pm233$ & $805\pm312$ & $19\pm23$ & $\bm{977\pm0}$ & $\bm{977\pm1}$ \\
M $\sigma=0$ Walker B1  & $437\pm20$ & $\bm{525\pm45}$ & $380\pm194$ & $444\pm16$ & $439\pm17$ \\
M $\sigma=0$ Walker B2  & $\bm{500\pm12}$ & $\bm{554\pm29}$ & $\bm{546\pm28}$ & $489\pm15$ & $\bm{504\pm4}$ \\
M $\sigma=0$ HC B1  & $\bm{4057\pm69}$ & $\bm{4255\pm150}$ & $\bm{4470\pm96}$ & $\bm{4032\pm72}$ & $\bm{4073\pm55}$ \\
M $\sigma=0$ HC B2  & $\bm{4013\pm12}$ & $\bm{4438\pm25}$ & $\bm{4395\pm31}$ & $\bm{3998\pm4}$ & $\bm{3999\pm6}$ \\
M $\sigma=0$ Ant B1  & $753\pm9$ & $\bm{996\pm52}$ & $734\pm43$ & $730\pm7$ & $732\pm11$ \\
M $\sigma=0$ Ant B2  & $738\pm4$ & $\bm{994\pm12}$ & $\bm{988\pm30}$ & $708\pm11$ & $725\pm7$ \\
M $\sigma=0$ Humanoid B1  & $\bm{4313\pm139}$ & $3108\pm510$ & $0\pm0$ & $\bm{4507\pm481}$ & $\bm{4521\pm156}$ \\
M $\sigma=0$ Humanoid B2  & $\bm{4053\pm252}$ & $2906\pm226$ & $0\pm0$ & $\bm{3994\pm530}$ & $\bm{3940\pm165}$ \\
M $\sigma=\sigma(s)$ Hopper B1  & $375\pm52$ & $881\pm155$ & $0\pm0$ & $\bm{1026\pm0}$ & $\bm{1026\pm0}$ \\
M $\sigma=\sigma(s)$ Hopper B2  & $254\pm102$ & $\bm{961\pm25}$ & $3\pm7$ & $\bm{977\pm0}$ & $\bm{977\pm0}$ \\
M $\sigma=\sigma(s)$ Walker B1  & $384\pm21$ & $399\pm21$ & $\bm{507\pm7}$ & $369\pm10$ & $359\pm15$ \\
M $\sigma=\sigma(s)$ Walker B2  & $\bm{512\pm24}$ & $\bm{517\pm19}$ & $\bm{515\pm30}$ & $\bm{527\pm12}$ & $\bm{532\pm5}$ \\
M $\sigma=\sigma(s)$ HC B1  & $4744\pm19$ & $\bm{5500\pm12}$ & $\bm{5443\pm21}$ & $4415\pm25$ & $4439\pm59$ \\
M $\sigma=\sigma(s)$ HC B2  & $4123\pm19$ & $\bm{4712\pm40}$ & $\bm{4824\pm51}$ & $3928\pm18$ & $3936\pm18$ \\
M $\sigma=\sigma(s)$ Ant B1  & $790\pm9$ & $\bm{1068\pm12}$ & $\bm{1161\pm32}$ & $775\pm7$ & $774\pm15$ \\
M $\sigma=\sigma(s)$ Ant B2  & $781\pm6$ & $\bm{1089\pm29}$ & $\bm{1150\pm18}$ & $768\pm5$ & $761\pm6$ \\
M $\sigma=\sigma(s)$ Humanoid B1  & $1375\pm387$ & $489\pm87$ & $0\pm0$ & $\bm{1947\pm901}$ & $\bm{1963\pm264}$ \\
M $\sigma=\sigma(s)$ Humanoid B2  & $1309\pm372$ & $816\pm177$ & $0\pm0$ & $\bm{3021\pm1042}$ & $\bm{2976\pm241}$ \\
O $\sigma=0$ Hopper B1  & $\bm{2602\pm5}$ & $1976\pm383$ & $1904\pm321$ & $\bm{2594\pm8}$ & $\bm{2603\pm4}$ \\
O $\sigma=0$ Hopper B2  & $\bm{3046\pm34}$ & $\bm{3014\pm47}$ & $2202\pm410$ & $\bm{3071\pm10}$ & $\bm{3050\pm22}$ \\
O $\sigma=0$ Walker B1  & $\bm{2735\pm26}$ & $2409\pm235$ & $877\pm1077$ & $\bm{2646\pm133}$ & $\bm{2691\pm121}$ \\
O $\sigma=0$ Walker B2  & $\bm{3019\pm6}$ & $\bm{3019\pm45}$ & $0\pm0$ & $\bm{3014\pm5}$ & $\bm{3013\pm5}$ \\
O $\sigma=0$ HC B1  & $\bm{11265\pm243}$ & $10405\pm275$ & $1755\pm1142$ & $\bm{11674\pm90}$ & $\bm{11661\pm49}$ \\
O $\sigma=0$ HC B2  & $\bm{11360\pm265}$ & $\bm{10792\pm209}$ & $1139\pm960$ & $\bm{11797\pm29}$ & $\bm{11691\pm96}$ \\
O $\sigma=0$ Ant B1  & $\bm{4901\pm65}$ & $\bm{4646\pm179}$ & $1756\pm2151$ & $\bm{4881\pm74}$ & $\bm{4933\pm74}$ \\
O $\sigma=0$ Ant B2  & $\bm{4975\pm108}$ & $\bm{4734\pm100}$ & $0\pm0$ & $\bm{5041\pm29}$ & $\bm{4974\pm52}$ \\
O $\sigma=0$ Humanoid B1  & $4872\pm895$ & $4884\pm641$ & $0\pm0$ & $\bm{5462\pm124}$ & $\bm{5503\pm1}$ \\
O $\sigma=0$ Humanoid B2  & $\bm{5320\pm125}$ & $\bm{5362\pm54}$ & $0\pm0$ & $\bm{5413\pm64}$ & $\bm{5413\pm29}$ \\
O $\sigma=\sigma(s)$ Hopper B1  & $2359\pm153$ & $\bm{2650\pm99}$ & $1962\pm300$ & $1952\pm85$ & $2012\pm101$ \\
O $\sigma=\sigma(s)$ Hopper B2  & $\bm{2035\pm217}$ & $1678\pm113$ & $1461\pm75$ & $\bm{2063\pm95}$ & $\bm{2092\pm100}$ \\
O $\sigma=\sigma(s)$ Walker B1  & $2834\pm120$ & $\bm{3386\pm196}$ & $\bm{3278\pm128}$ & $2024\pm131$ & $1987\pm114$ \\
O $\sigma=\sigma(s)$ Walker B2  & $\bm{3200\pm16}$ & $\bm{3375\pm12}$ & $2100\pm1715$ & $\bm{3091\pm15}$ & $\bm{3090\pm10}$ \\
O $\sigma=\sigma(s)$ HC B1  & $10258\pm1255$ & $\bm{10928\pm215}$ & $694\pm651$ & $\bm{11659\pm75}$ & $\bm{11663\pm44}$ \\
O $\sigma=\sigma(s)$ HC B2  & $\bm{10882\pm634}$ & $\bm{11755\pm97}$ & $1470\pm1211$ & $\bm{11871\pm57}$ & $\bm{11819\pm78}$ \\
O $\sigma=\sigma(s)$ Ant B1  & $\bm{4981\pm91}$ & $\bm{4878\pm117}$ & $3462\pm1740$ & $\bm{5000\pm79}$ & $\bm{4992\pm86}$ \\
O $\sigma=\sigma(s)$ Ant B2  & $\bm{5067\pm83}$ & $\bm{5054\pm157}$ & $0\pm0$ & $\bm{5079\pm55}$ & $\bm{5124\pm47}$ \\
O $\sigma=\sigma(s)$ Humanoid B1  & $2129\pm381$ & $1715\pm637$ & $0\pm0$ & $\bm{3514\pm1195}$ & $\bm{3180\pm503}$ \\
O $\sigma=\sigma(s)$ Humanoid B2  & $4328\pm569$ & $1970\pm512$ & $0\pm0$ & $\bm{4875\pm885}$ & $\bm{4772\pm272}$ \\
					\bottomrule
				\end{tabular}
			\end{sc}
		\end{small}
	\end{center}
	\vskip -0.1in
\end{table*}

\clearpage

\section{Learning Curves for all 62 Batches}
\subsection{DDPG training batches}

\begin{figure}[ht]
\vskip 0.1in
\centering
\begin{subfigure}{0.24\columnwidth}
 \centering
 \includegraphics[width=0.99\linewidth]{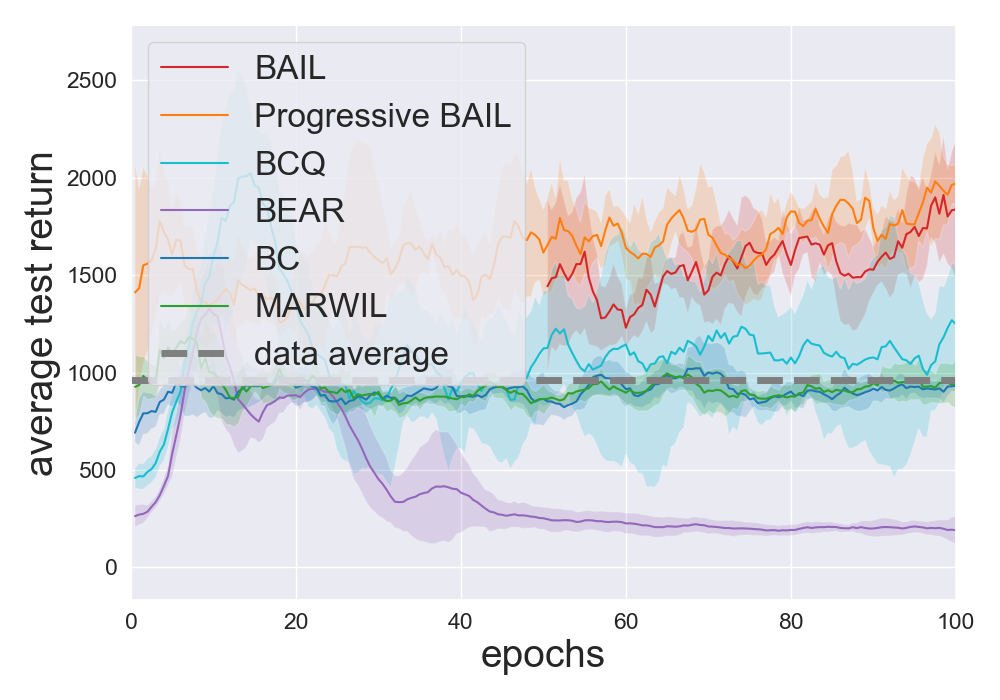}
 \caption{Hopper, batch 1}
\end{subfigure}
\begin{subfigure}{0.24\columnwidth}
 \centering
 \includegraphics[width=0.99\linewidth]{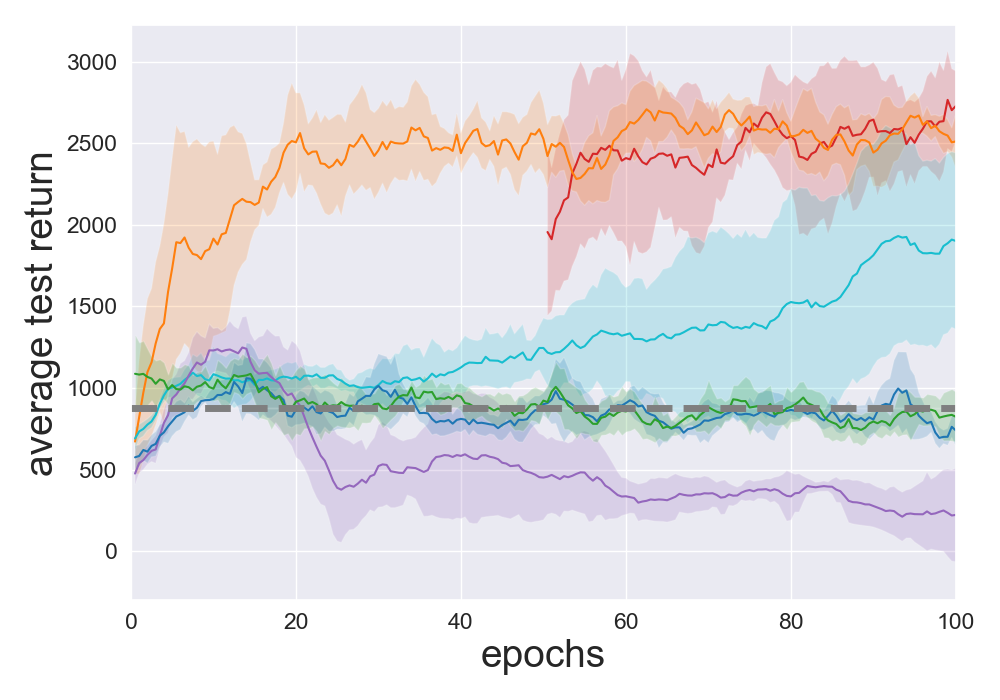}
 \caption{Hopper, batch 2}
\end{subfigure}
\begin{subfigure}{0.24\columnwidth}
 \centering
 \includegraphics[width=0.99\linewidth]{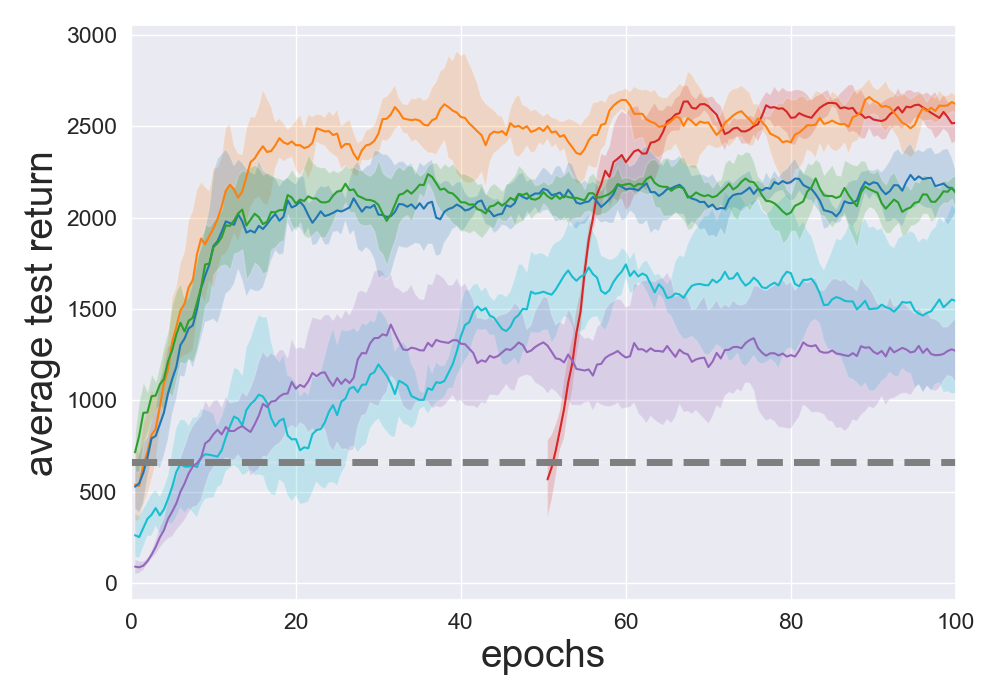}
 \caption{Walker2d, batch 1}
\end{subfigure}
\begin{subfigure}{0.24\columnwidth}
 \centering
 \includegraphics[width=0.99\linewidth]{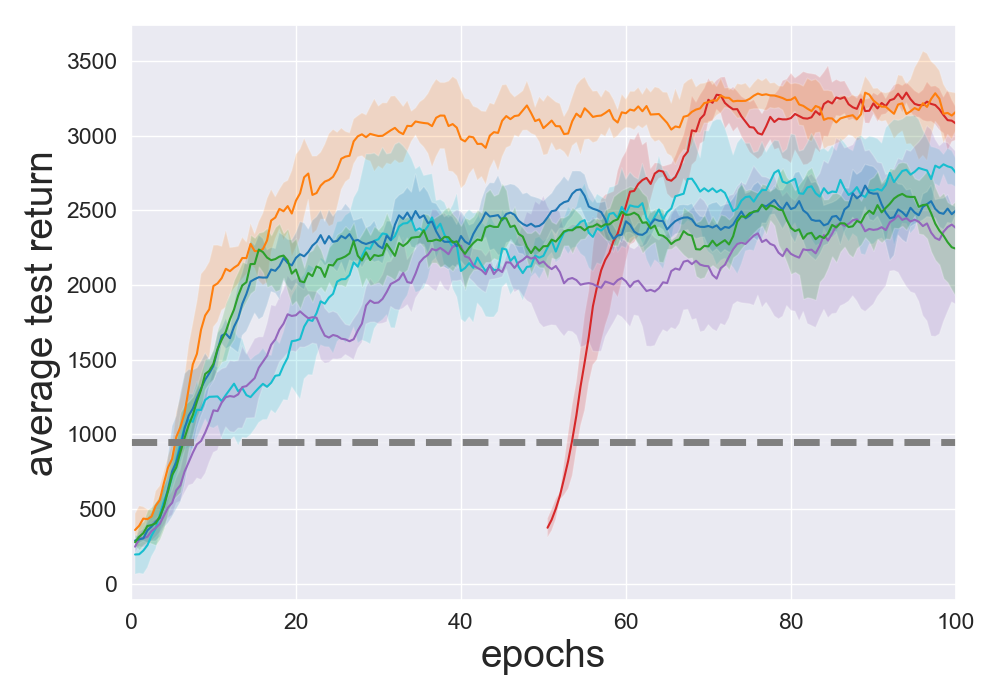}
 \caption{Walker2d, batch 2}
\end{subfigure}
\begin{subfigure}{0.24\columnwidth}
 \centering
 \includegraphics[width=0.99\linewidth]{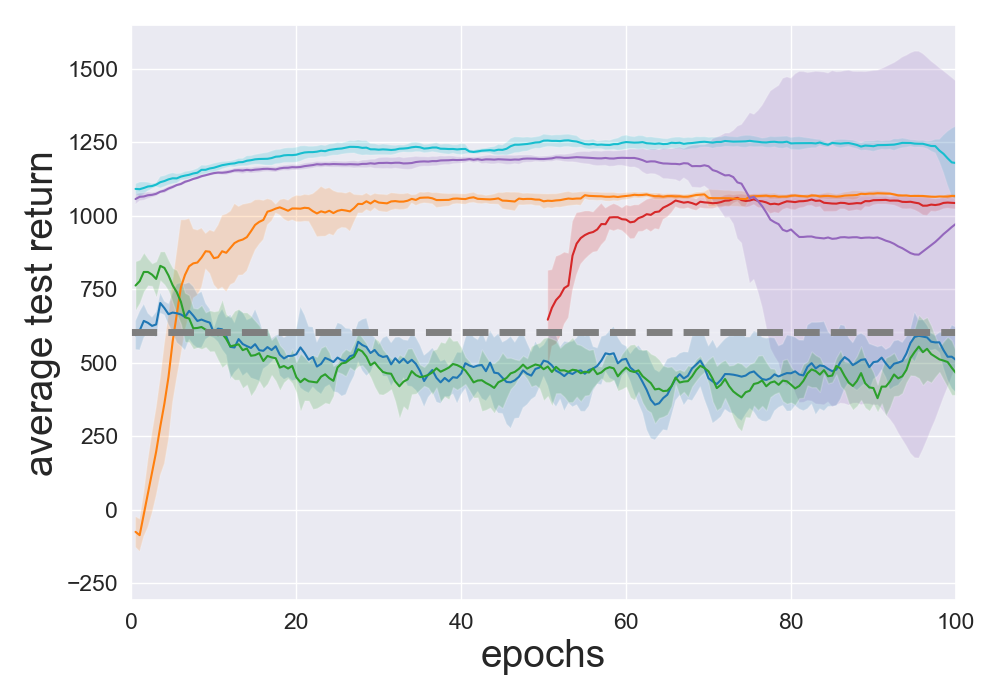}
 \caption{HalfCheetah, batch 1}
\end{subfigure}
\begin{subfigure}{0.24\columnwidth}
 \centering
 \includegraphics[width=0.99\linewidth]{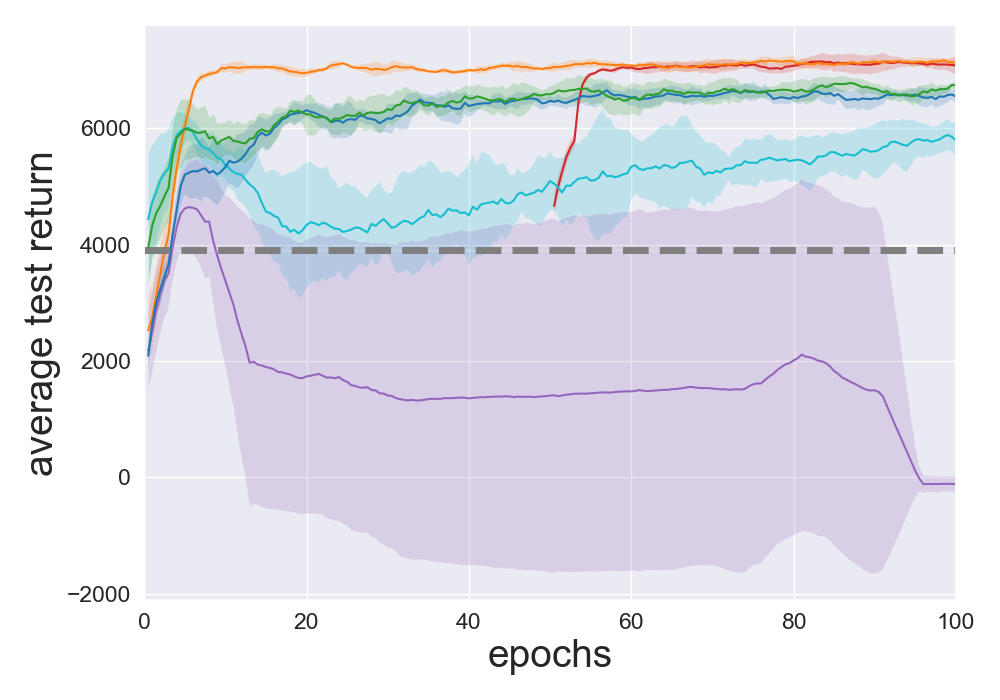}
 \caption{HalfCheetah, batch 2}
\end{subfigure}
\caption{Performance of batch DRL algorithms on DDPG training batches with $\sigma=0.5$. The policy networks for all algorithms are trained for 100 epochs except BAIL, which is trained for 50 epochs after training the upper envelope for 50 epochs.}
\label{fig:bail_train_n_0-5}
\vskip -0.1in
\end{figure}

\begin{figure}[ht]
\vskip 0.1in
\centering
\begin{subfigure}{0.24\columnwidth}
 \centering
 \includegraphics[width=0.99\linewidth]{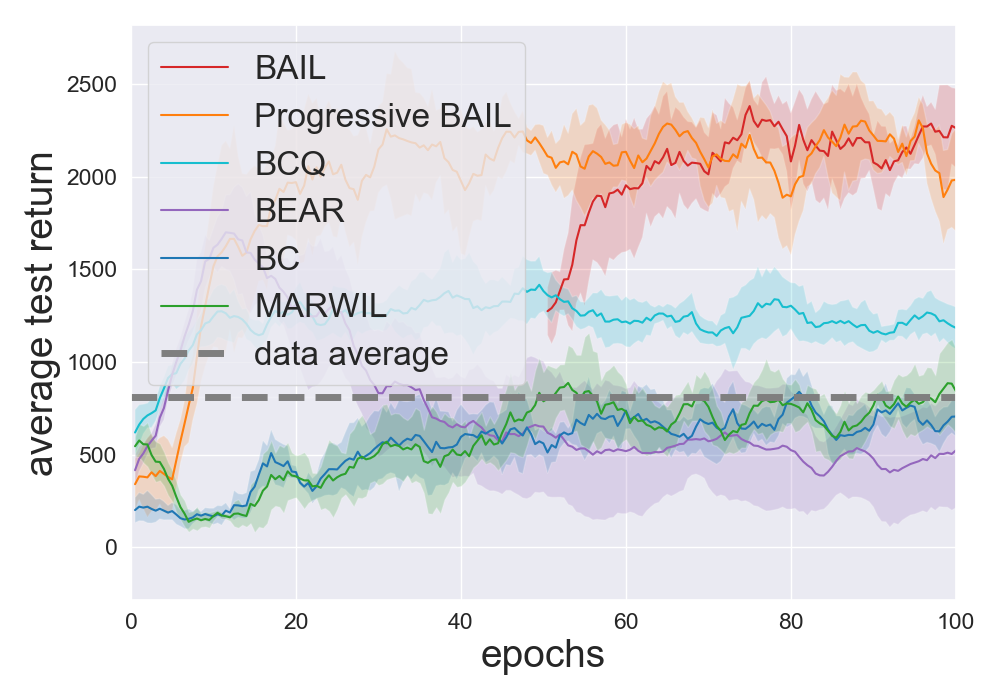}
 \caption{Hopper, batch 1}
\end{subfigure}
\begin{subfigure}{0.24\columnwidth}
 \centering
 \includegraphics[width=0.99\linewidth]{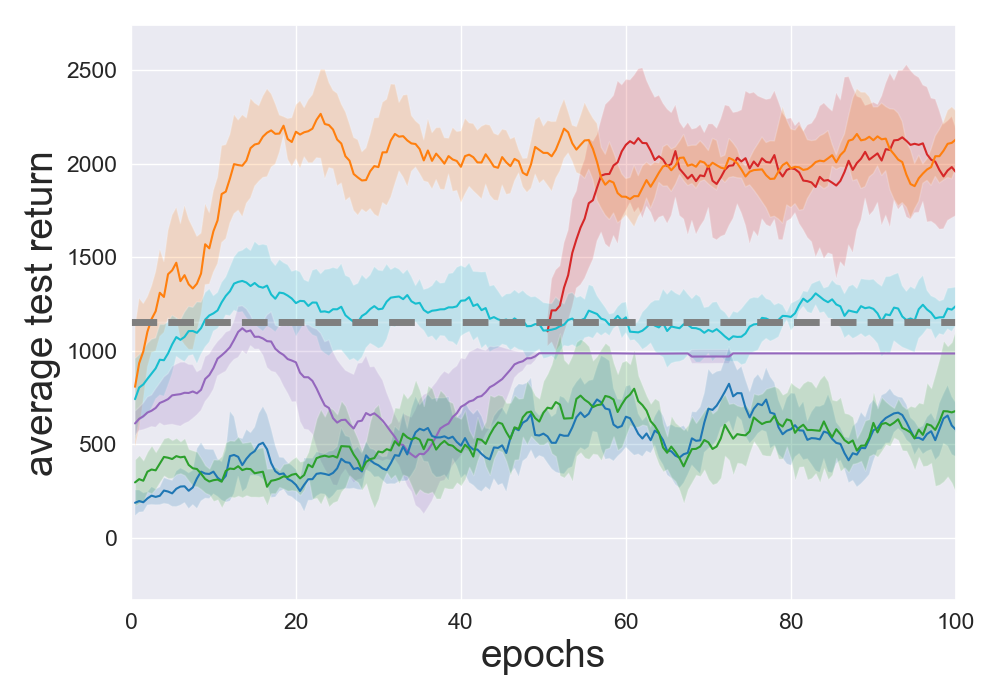}
 \caption{Hopper, batch 2}
\end{subfigure}
\begin{subfigure}{0.24\columnwidth}
 \centering
 \includegraphics[width=0.99\linewidth]{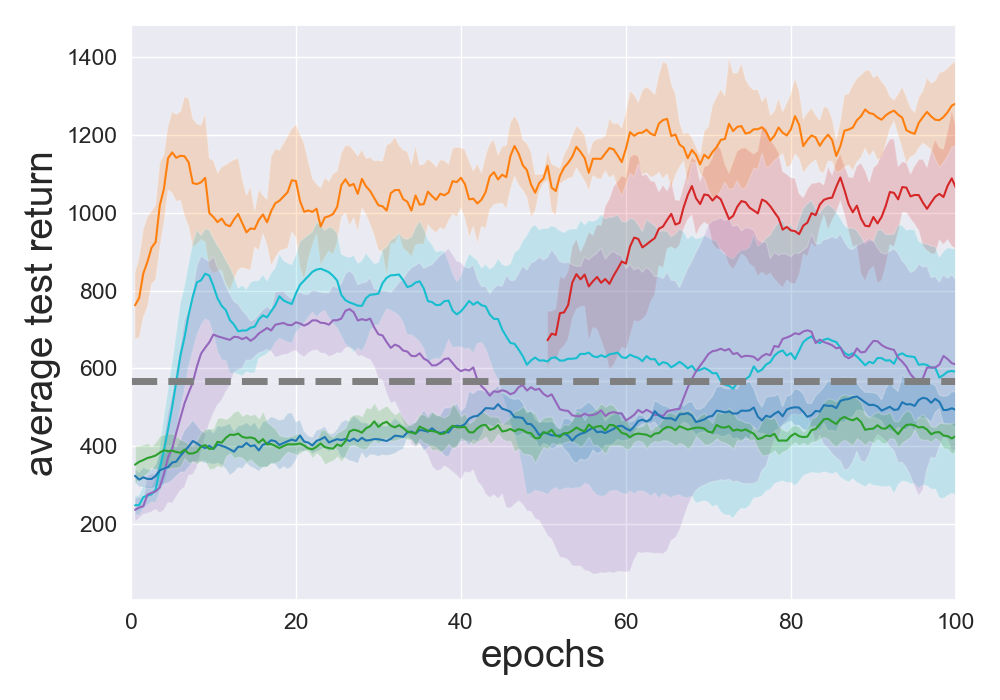}
 \caption{Walker2d, batch 1}
\end{subfigure}
\begin{subfigure}{0.24\columnwidth}
 \centering
 \includegraphics[width=0.99\linewidth]{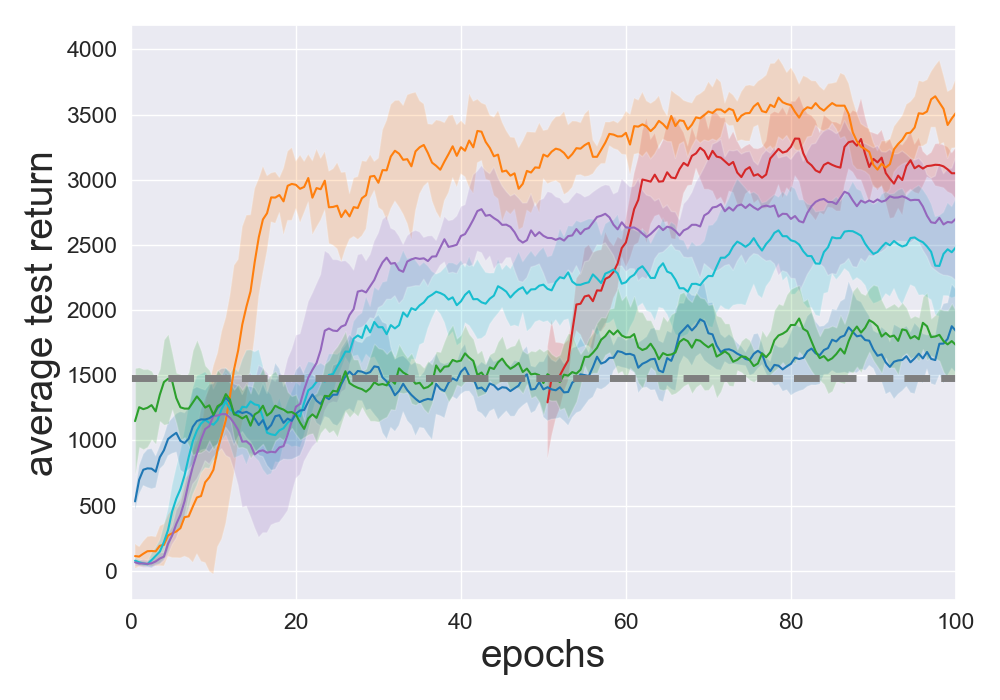}
 \caption{Walker2d, batch 2}
\end{subfigure}
\begin{subfigure}{0.24\columnwidth}
 \centering
 \includegraphics[width=0.99\linewidth]{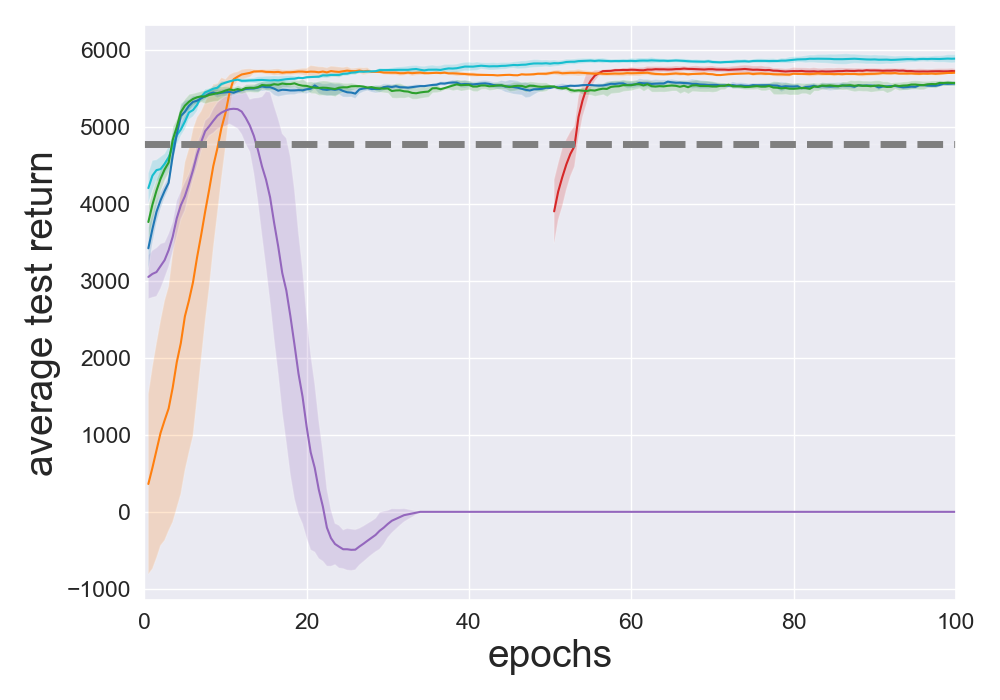}
 \caption{HalfCheetah, batch 1}
\end{subfigure}
\begin{subfigure}{0.24\columnwidth}
 \centering
 \includegraphics[width=0.99\linewidth]{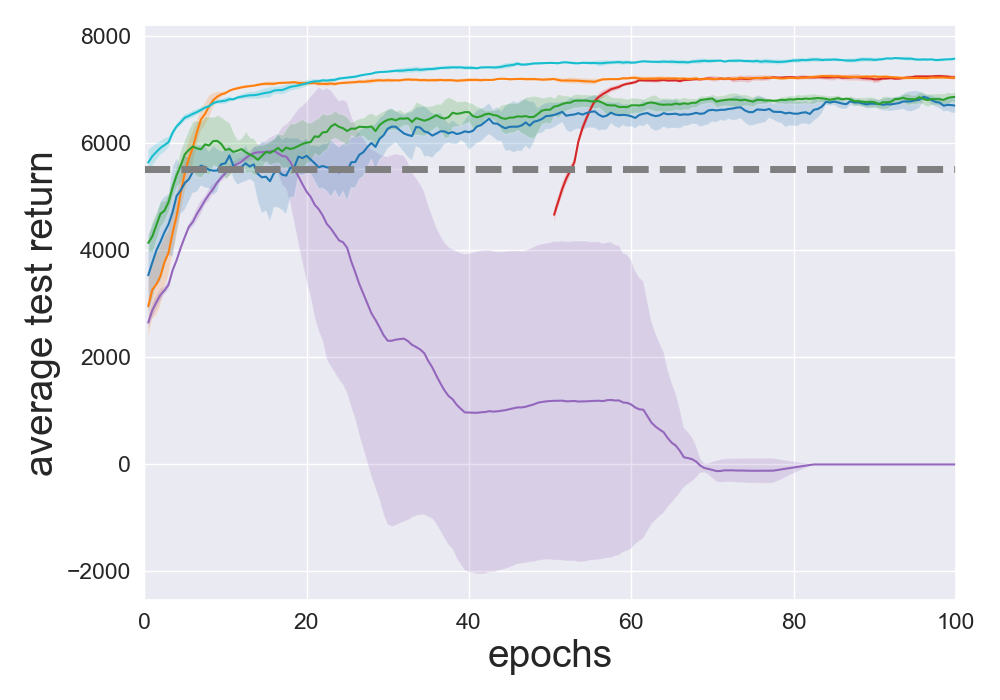}
 \caption{HalfCheetah, batch 2}
\end{subfigure}
\caption{Performance of batch DRL algorithms on DDPG training batches with $\sigma=0.1$. The policy networks for all algorithms are trained for 100 epochs except BAIL, which is trained for 50 epochs after training the upper envelope for 50 epochs.}
\label{fig:bail_final_n_0-1}
\vskip -0.1in
\end{figure}

\newpage
\subsection{SAC training batches}

\begin{figure}[ht]
\vskip 0.1in
\centering
\begin{subfigure}{0.24\columnwidth}
 \centering
 \includegraphics[width=0.99\linewidth]{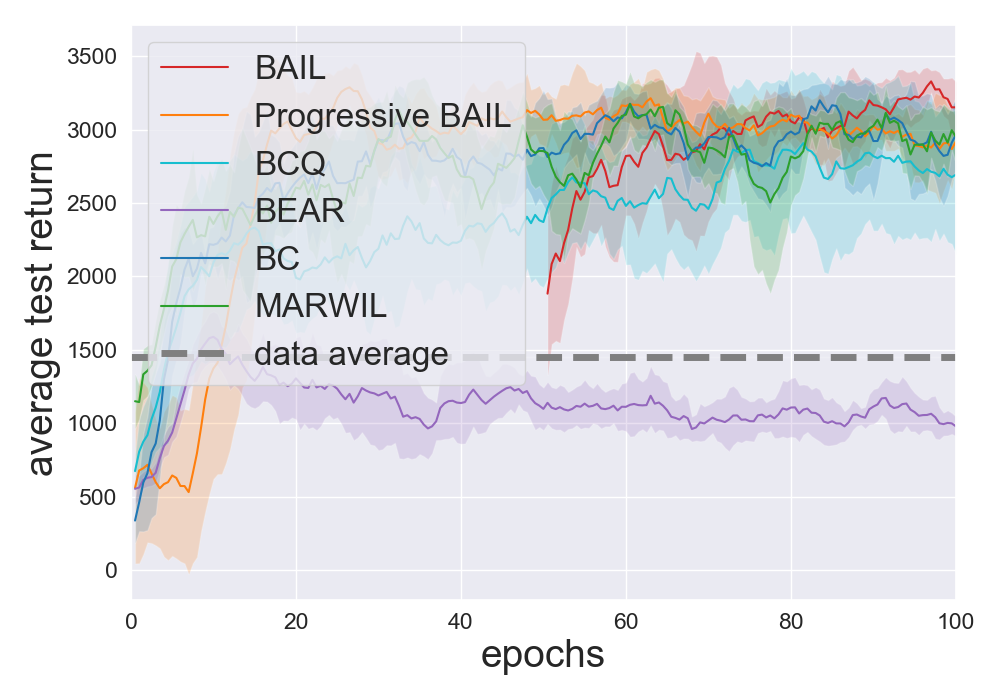}
 \caption{Hopper, batch 1}
\end{subfigure}
\begin{subfigure}{0.24\columnwidth}
 \centering
 \includegraphics[width=0.99\linewidth]{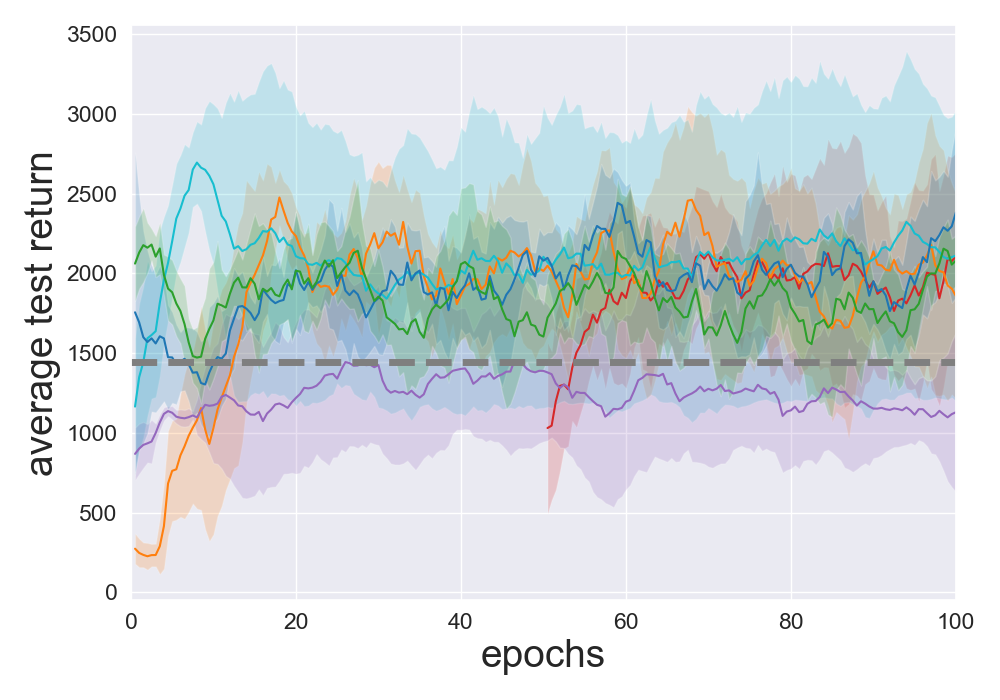}
 \caption{Hopper, batch 2}
\end{subfigure}
\begin{subfigure}{0.24\columnwidth}
 \centering
 \includegraphics[width=0.99\linewidth]{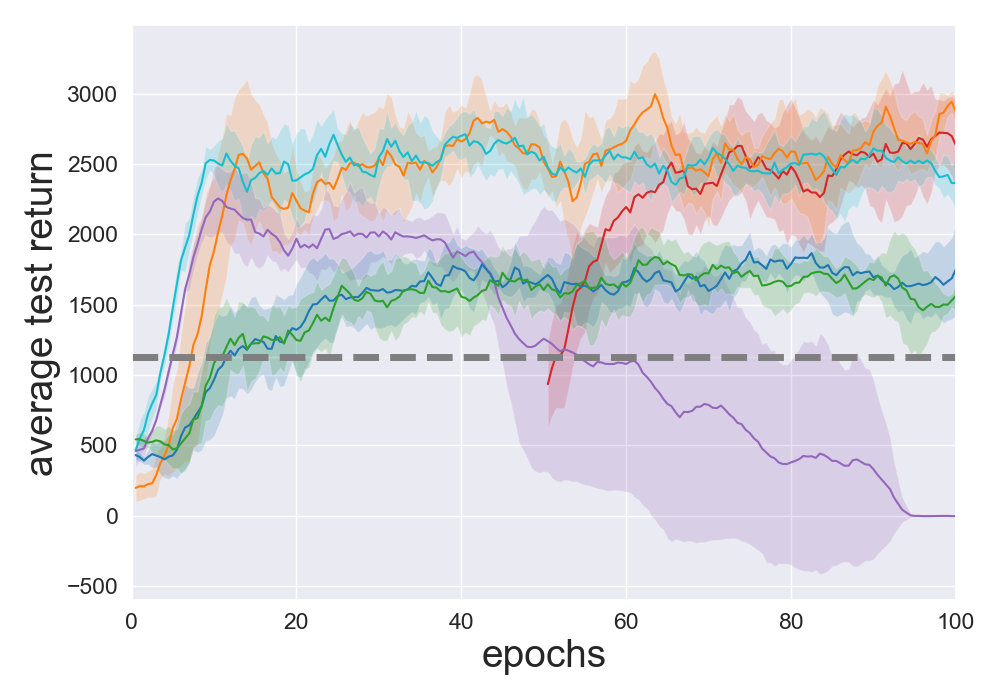}
 \caption{Walker2d, batch 1}
\end{subfigure}
\begin{subfigure}{0.24\columnwidth}
 \centering
 \includegraphics[width=0.99\linewidth]{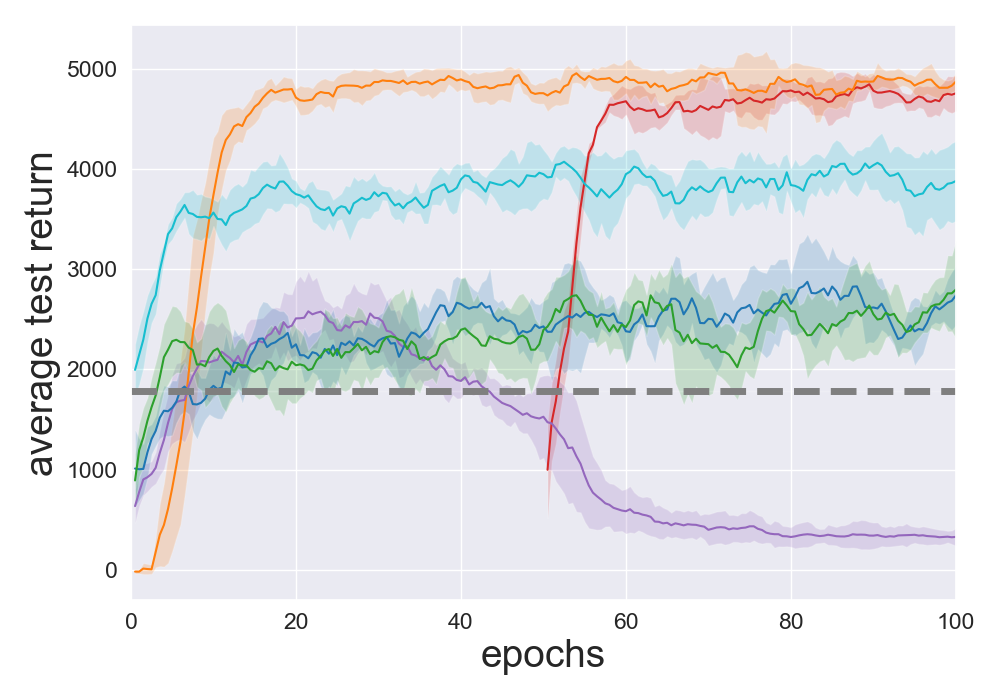}
 \caption{Walker2d, batch 2}
\end{subfigure}
\begin{subfigure}{0.24\columnwidth}
 \centering
 \includegraphics[width=0.99\linewidth]{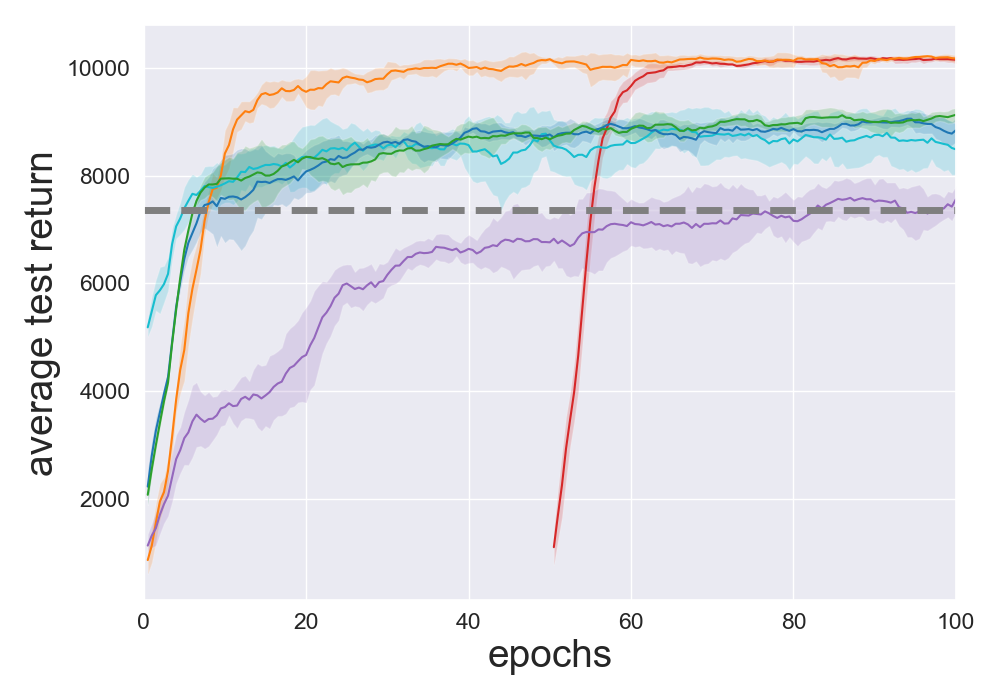}
 \caption{HalfCheetah, batch 1}
\end{subfigure}
\begin{subfigure}{0.24\columnwidth}
 \centering
 \includegraphics[width=0.99\linewidth]{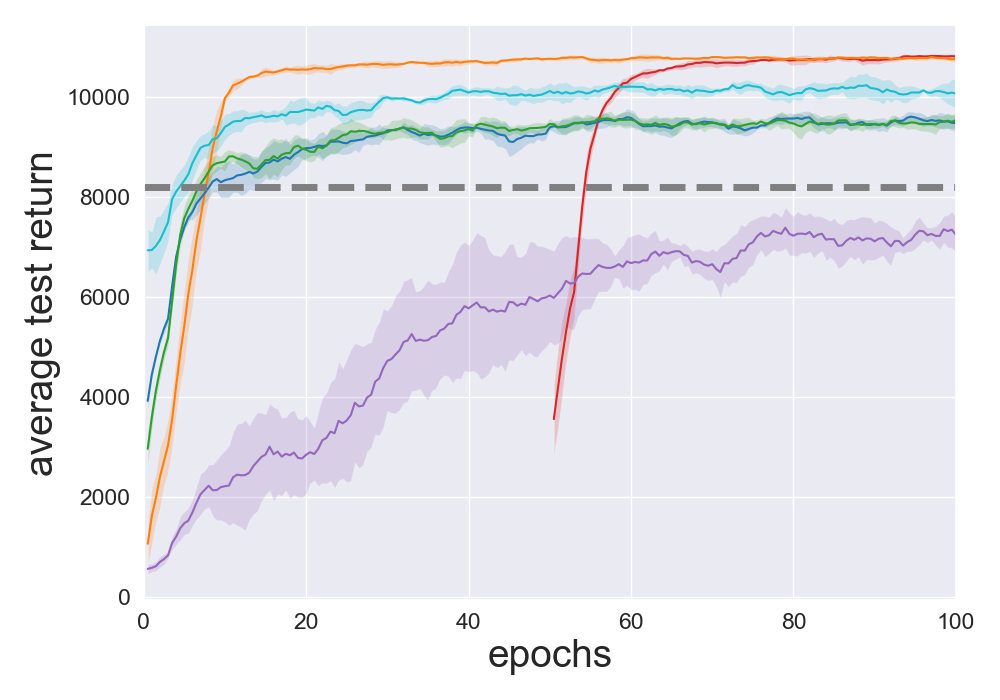}
 \caption{HalfCheetah, batch 2}
\end{subfigure}
\begin{subfigure}{0.24\columnwidth}
 \centering
 \includegraphics[width=0.99\linewidth]{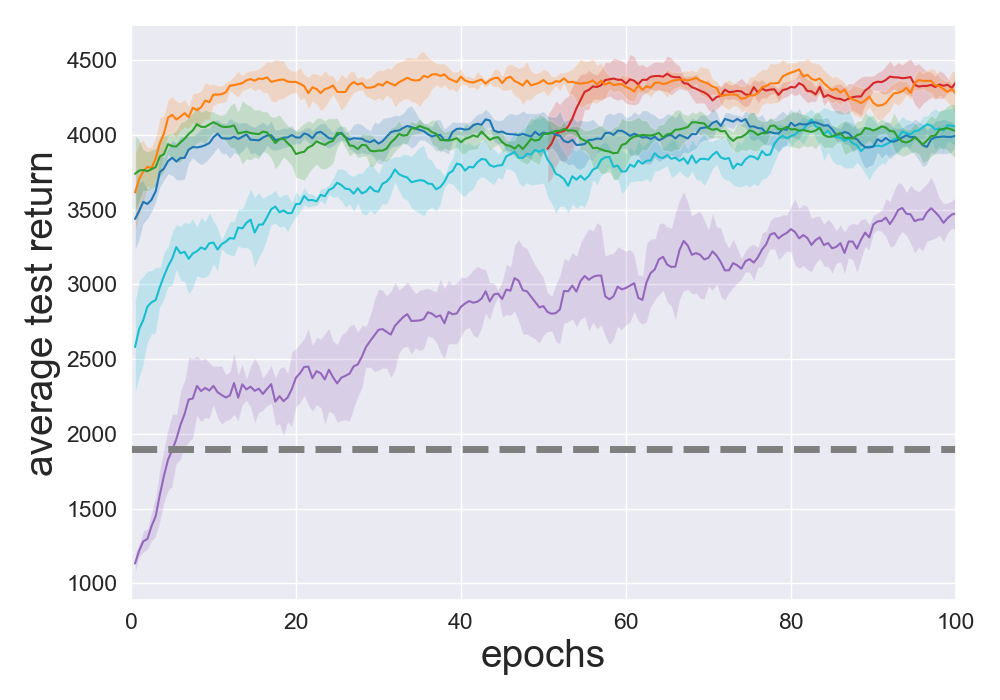}
 \caption{Ant, batch 1}
\end{subfigure}
\begin{subfigure}{0.24\columnwidth}
 \centering
 \includegraphics[width=0.99\linewidth]{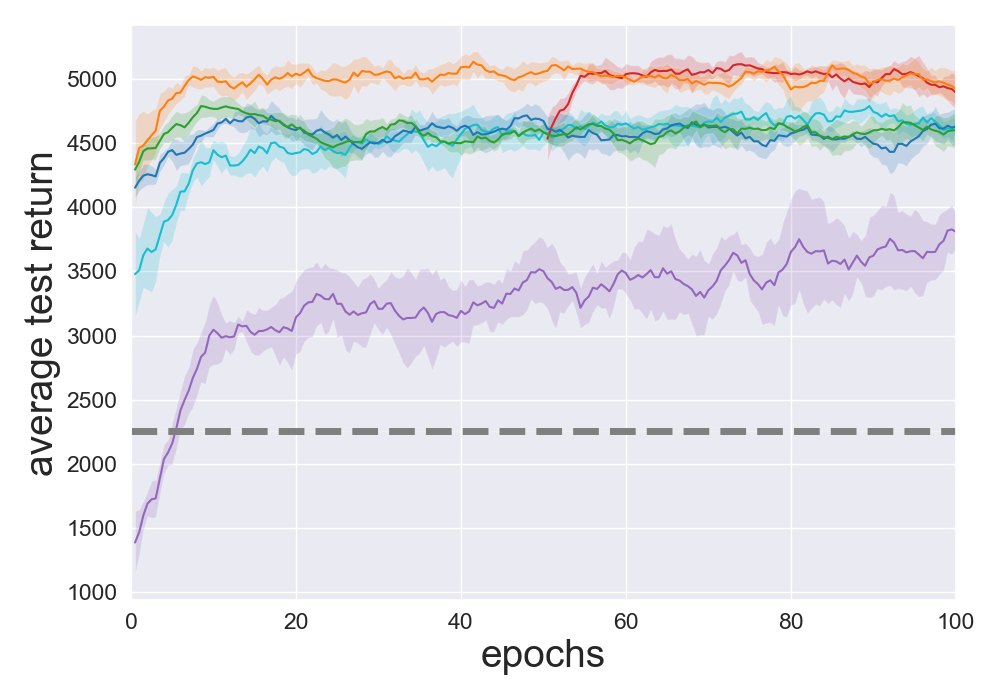}
 \caption{Ant, batch 2}
\end{subfigure}
\caption{Performance of batch DRL algorithms on SAC training batches. The policy networks for all algorithms are trained for 100 epochs except  BAIL, which is trained for 50 epochs after training the upper envelope for 50 epochs.}
\label{fig:bail_sac_training}
\vskip -0.1in
\end{figure}

\newpage
\subsection{SAC mediocre execution batches}

\begin{figure}[ht]
\vskip 0.1in
\centering
\begin{subfigure}{0.24\columnwidth}
 \centering
 \includegraphics[width=0.99\linewidth]{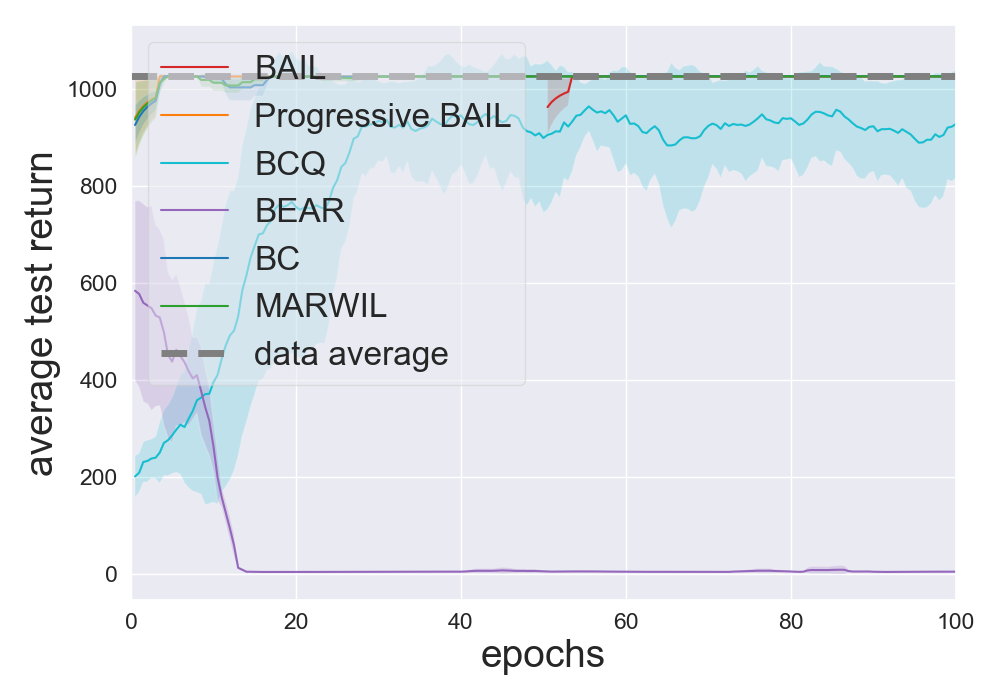}
 \caption{Hopper, batch 1}
\end{subfigure}
\begin{subfigure}{0.24\columnwidth}
 \centering
 \includegraphics[width=0.99\linewidth]{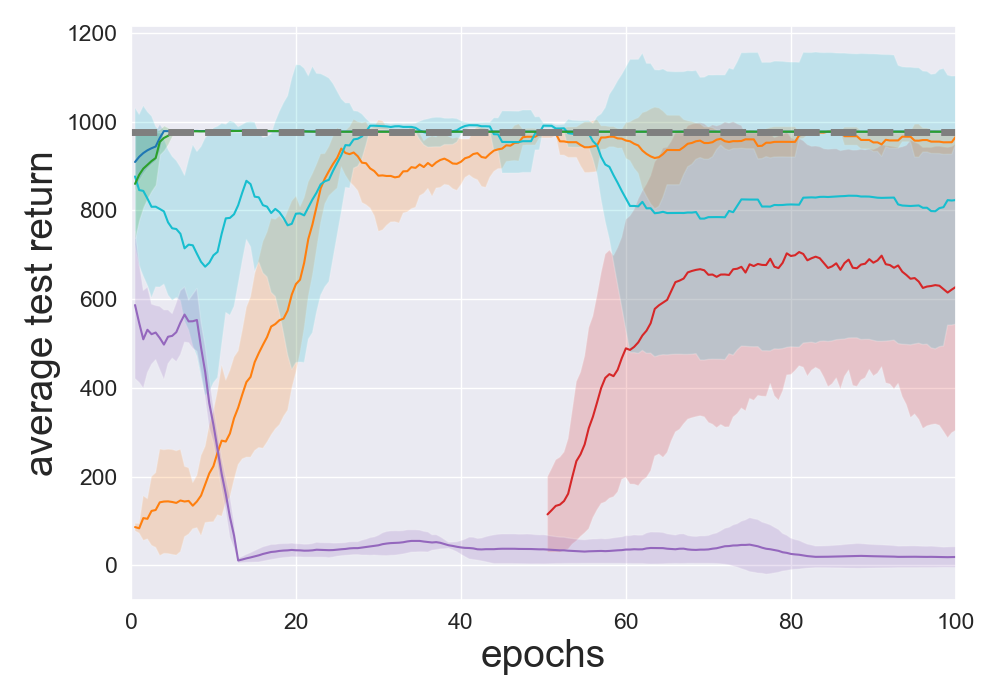}
 \caption{Hopper, batch 2}
\end{subfigure}
\begin{subfigure}{0.24\columnwidth}
 \centering
 \includegraphics[width=0.99\linewidth]{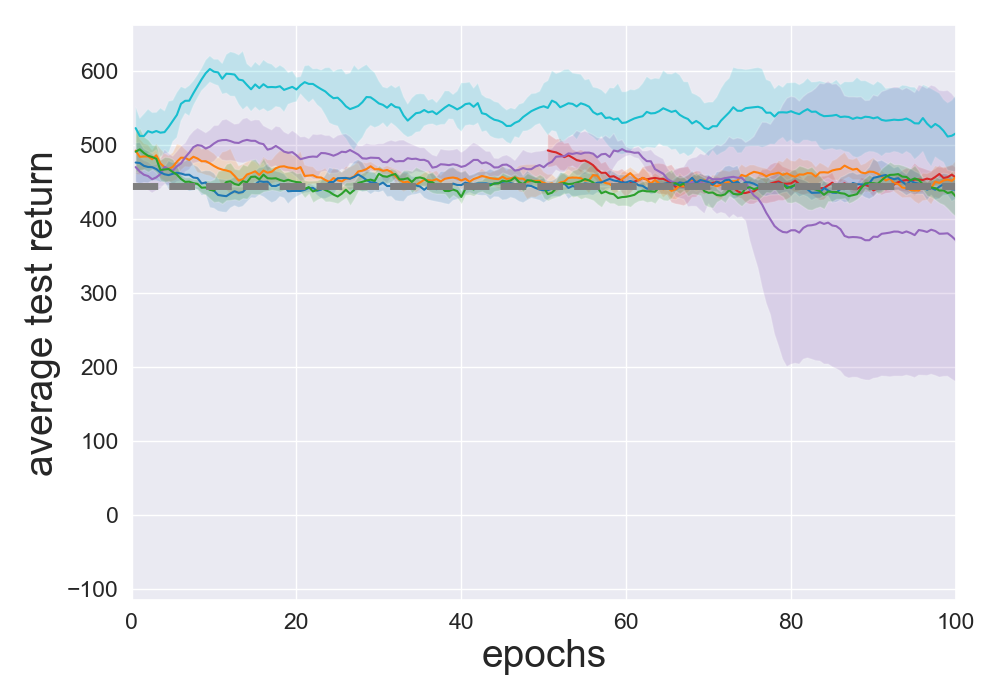}
 \caption{Walker2d, batch 1}
\end{subfigure}
\begin{subfigure}{0.24\columnwidth}
 \centering
 \includegraphics[width=0.99\linewidth]{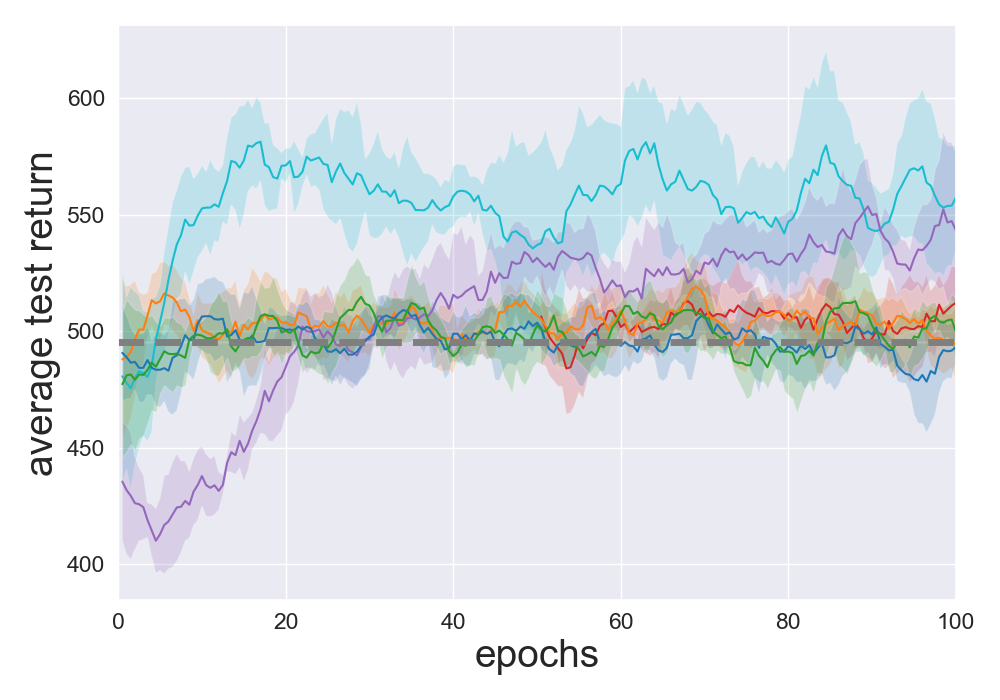}
 \caption{Walker2d, batch 2}
\end{subfigure}
\begin{subfigure}{0.24\columnwidth}
 \centering
 \includegraphics[width=0.99\linewidth]{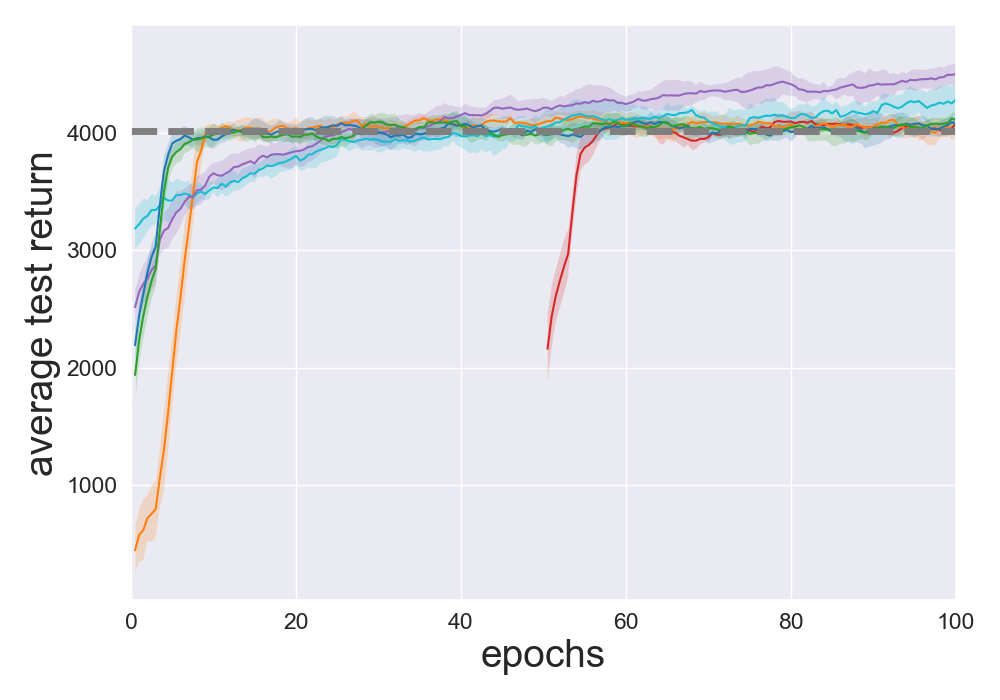}
 \caption{HalfCheetah, batch 1}
\end{subfigure}
\begin{subfigure}{0.24\columnwidth}
 \centering
 \includegraphics[width=0.99\linewidth]{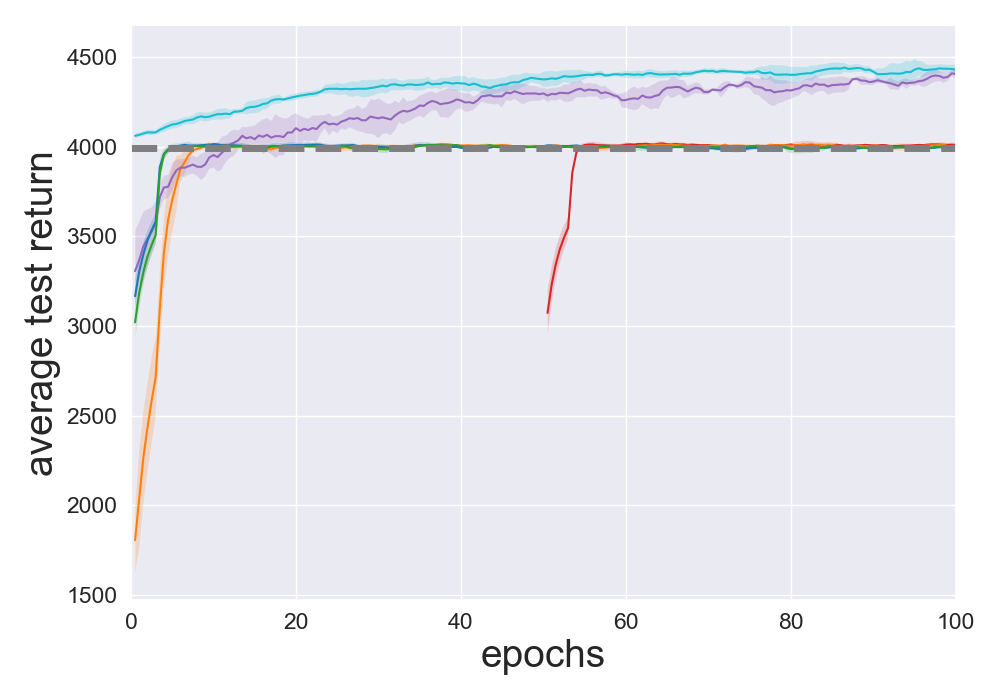}
 \caption{HalfCheetah, batch 2}
\end{subfigure}
\begin{subfigure}{0.24\columnwidth}
 \centering
 \includegraphics[width=0.99\linewidth]{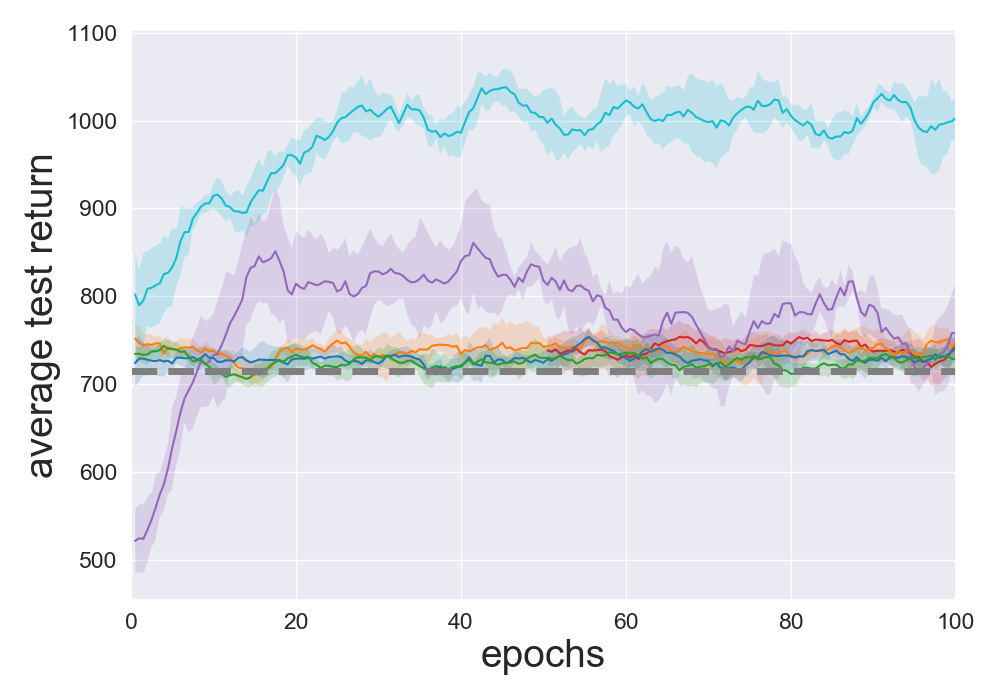}
 \caption{Ant, batch 1}
\end{subfigure}
\begin{subfigure}{0.24\columnwidth}
 \centering
 \includegraphics[width=0.99\linewidth]{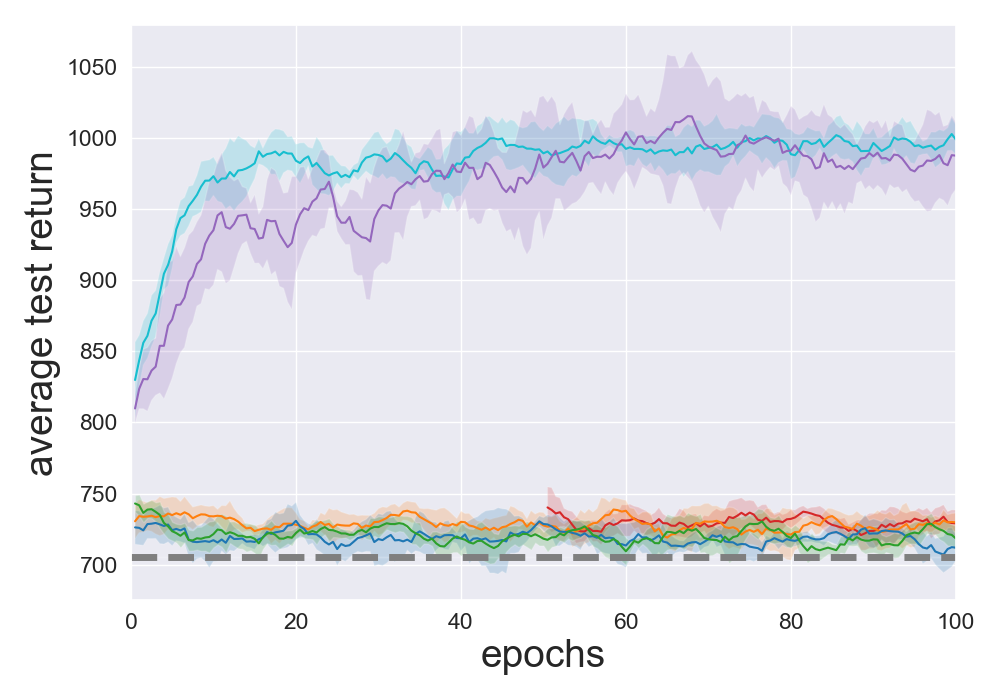}
 \caption{Ant, batch 2}
\end{subfigure}
\caption{Performance of batch DRL algorithms on SAC mediocre execution batches with $\sigma = 0$. The policy networks for all algorithms are trained for 100 epochs except  BAIL, which is trained for 50 epochs after training the upper envelope for 50 epochs.}
\label{fig:bail_mediocre_no_sigma}
\vskip -0.1in
\end{figure}

\begin{figure}[ht]
\vskip 0.1in
\centering
\begin{subfigure}{0.24\columnwidth}
 \centering
 \includegraphics[width=0.99\linewidth]{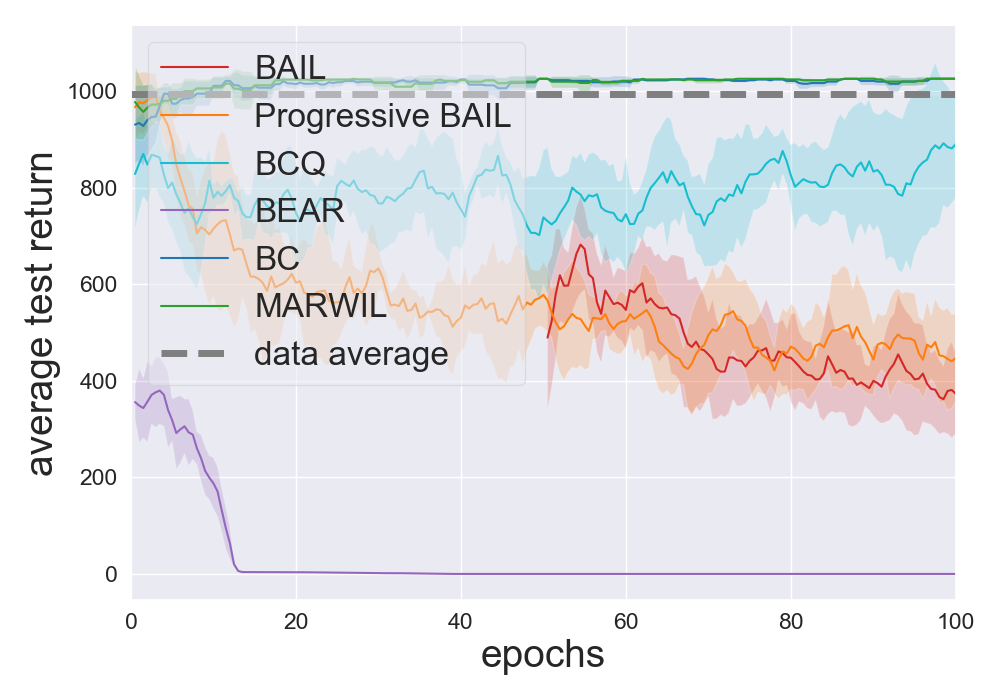}
 \caption{Hopper, batch 1}
\end{subfigure}
\begin{subfigure}{0.24\columnwidth}
 \centering
 \includegraphics[width=0.99\linewidth]{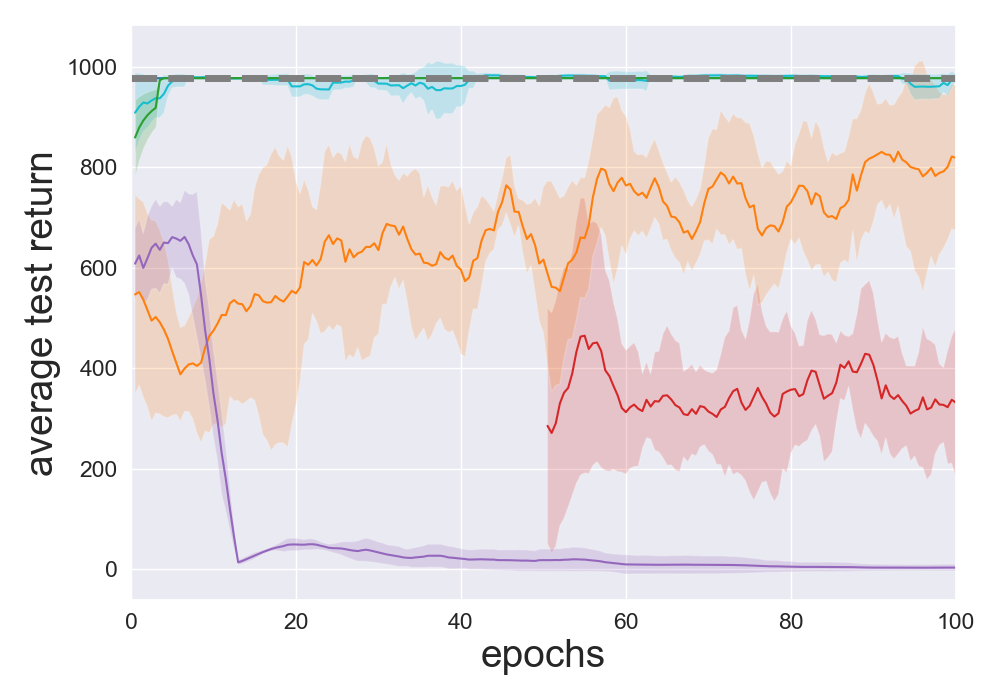}
 \caption{Hopper, batch 2}
\end{subfigure}
\begin{subfigure}{0.24\columnwidth}
 \centering
 \includegraphics[width=0.99\linewidth]{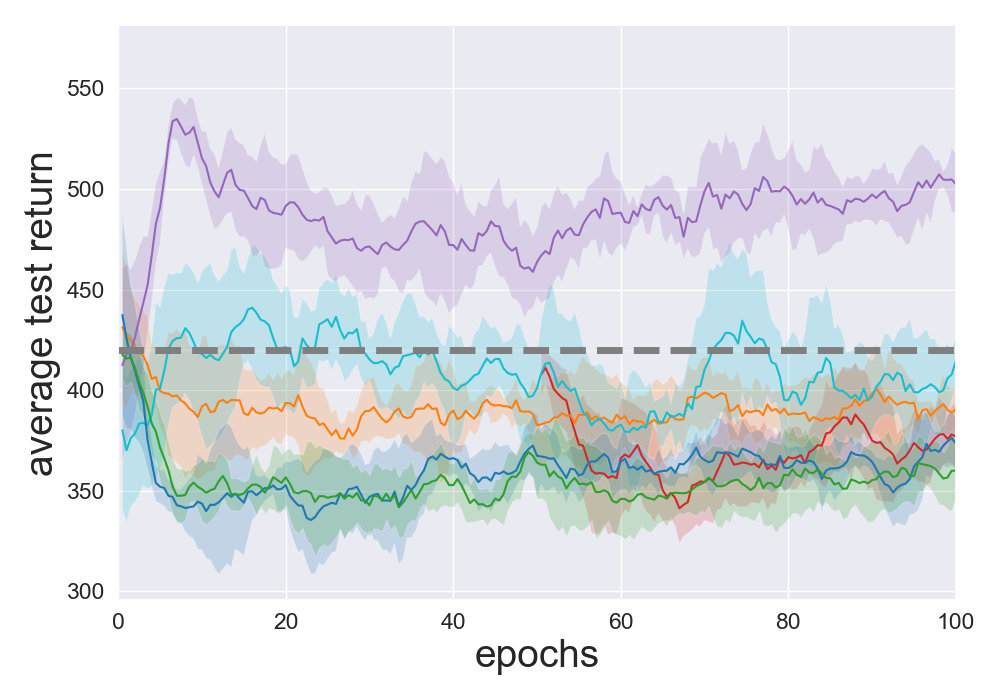}
 \caption{Walker2d, batch 1}
\end{subfigure}
\begin{subfigure}{0.24\columnwidth}
 \centering
 \includegraphics[width=0.99\linewidth]{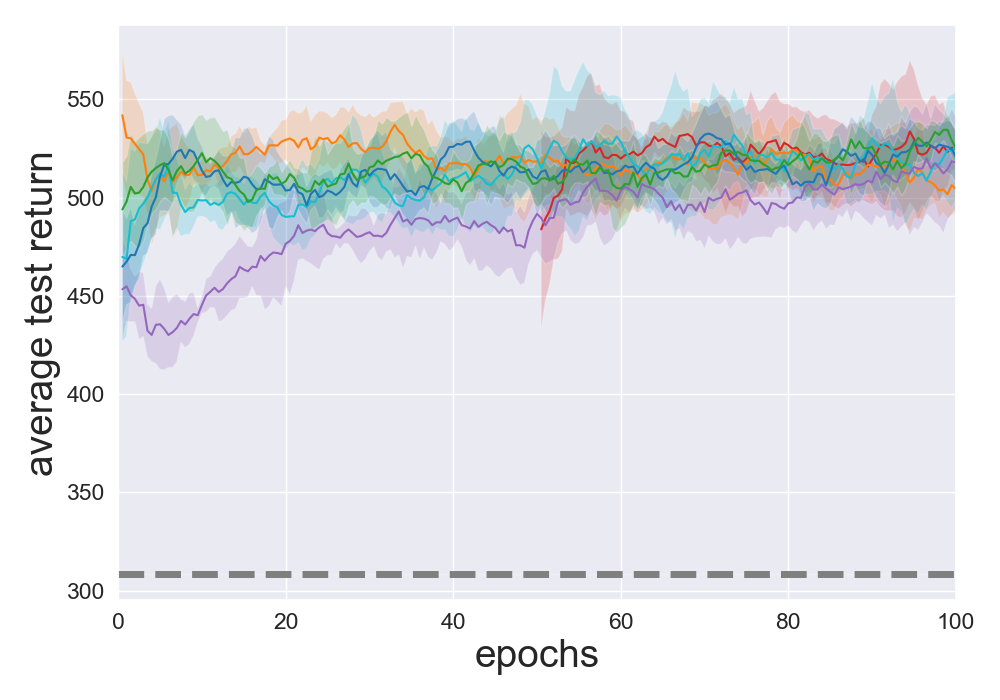}
 \caption{Walker2d, batch 2}
\end{subfigure}
\begin{subfigure}{0.24\columnwidth}
 \centering
 \includegraphics[width=0.99\linewidth]{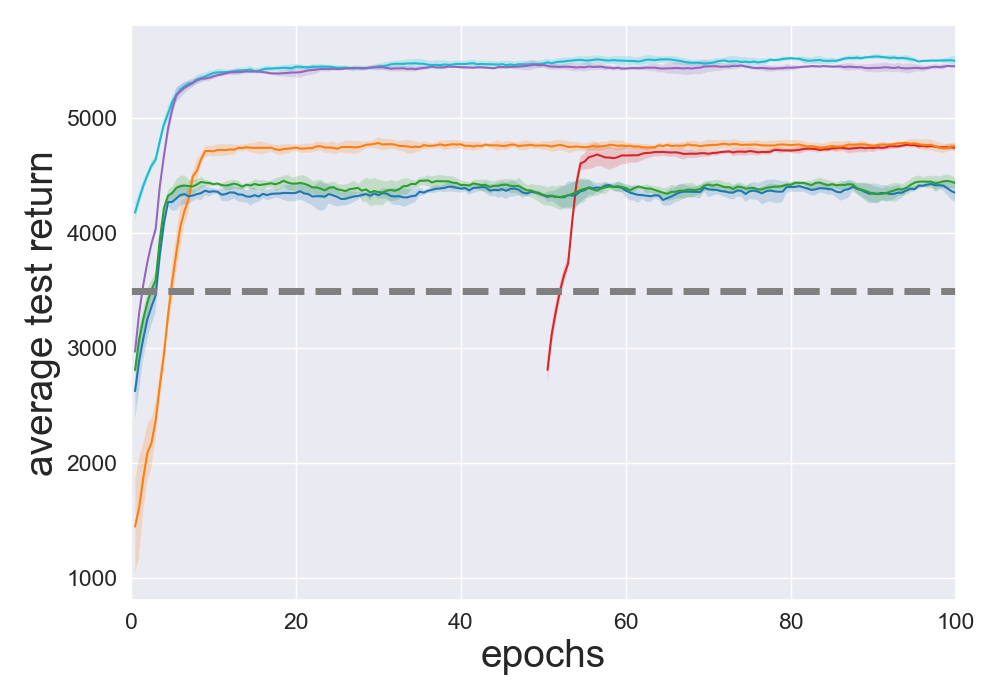}
 \caption{HalfCheetah, batch 1}
\end{subfigure}
\begin{subfigure}{0.24\columnwidth}
 \centering
 \includegraphics[width=0.99\linewidth]{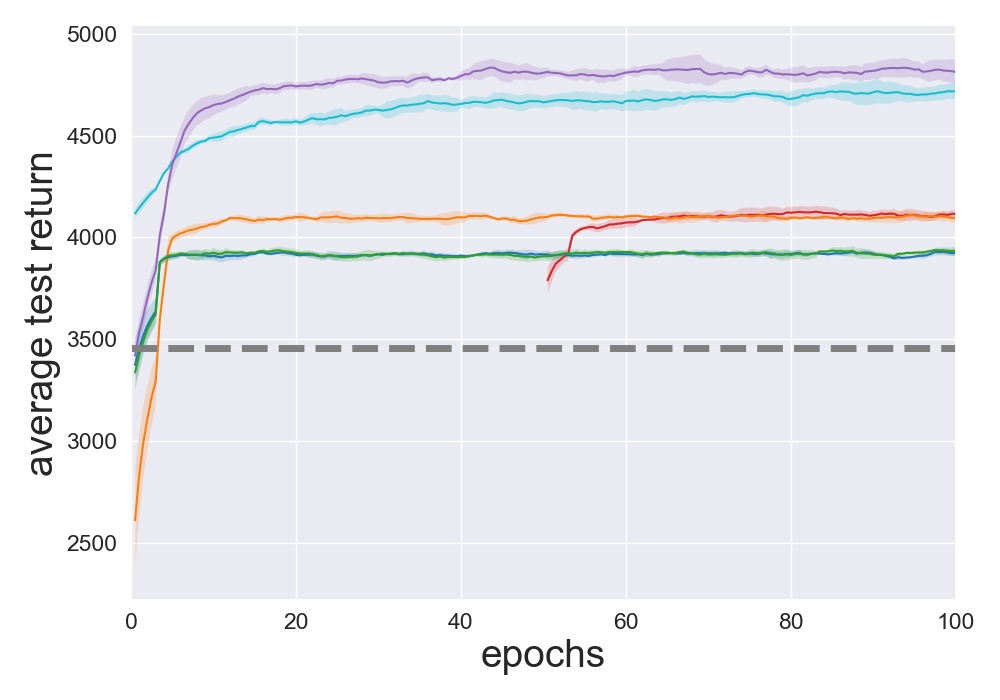}
 \caption{HalfCheetah, batch 2}
\end{subfigure}
\begin{subfigure}{0.24\columnwidth}
 \centering
 \includegraphics[width=0.99\linewidth]{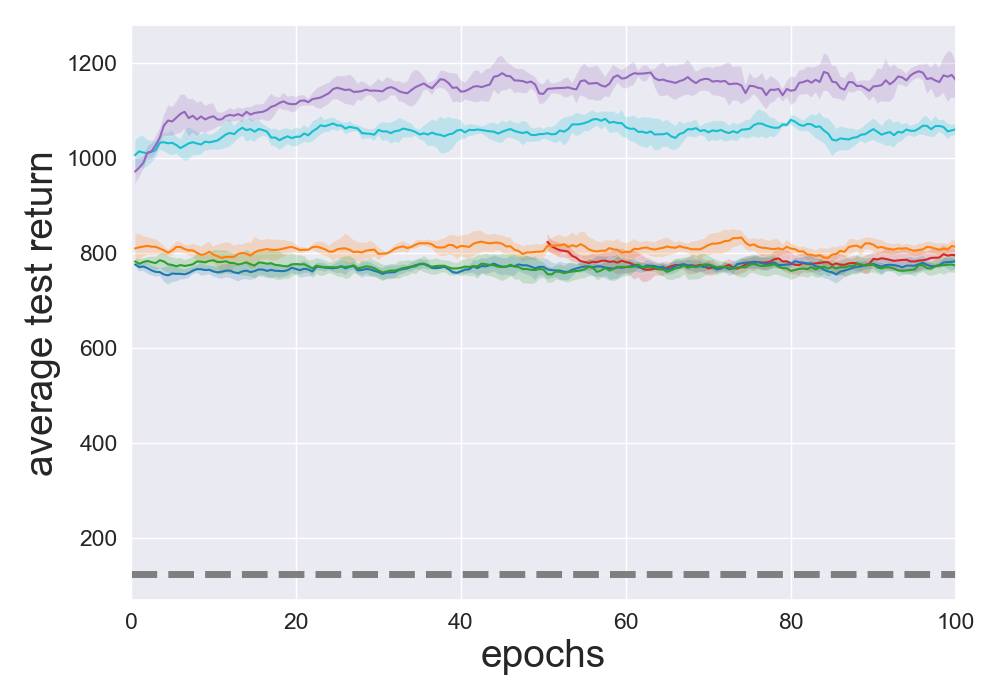}
 \caption{Ant, batch 1}
\end{subfigure}
\begin{subfigure}{0.24\columnwidth}
 \centering
 \includegraphics[width=0.99\linewidth]{figures/nips-6algo_sac_mediocre_B2_sch_ant_smooth-10.png}
 \caption{Ant, batch 2}
\end{subfigure}
\caption{Performance of batch DRL algorithms on SAC mediocre execution batches with $\sigma = \sigma(s)$. The policy networks for all algorithms are trained for 100 epochs except BAIL, which is trained for 50 epochs after training the upper envelope for 50 epochs.}
\label{fig:bail_mediocre_with_sigma}
\vskip -0.1in
\end{figure}

\newpage
\subsection{SAC optimal execution batches}

\begin{figure}[ht]
\vskip 0.1in
\centering
\begin{subfigure}{0.24\columnwidth}
 \centering
 \includegraphics[width=0.99\linewidth]{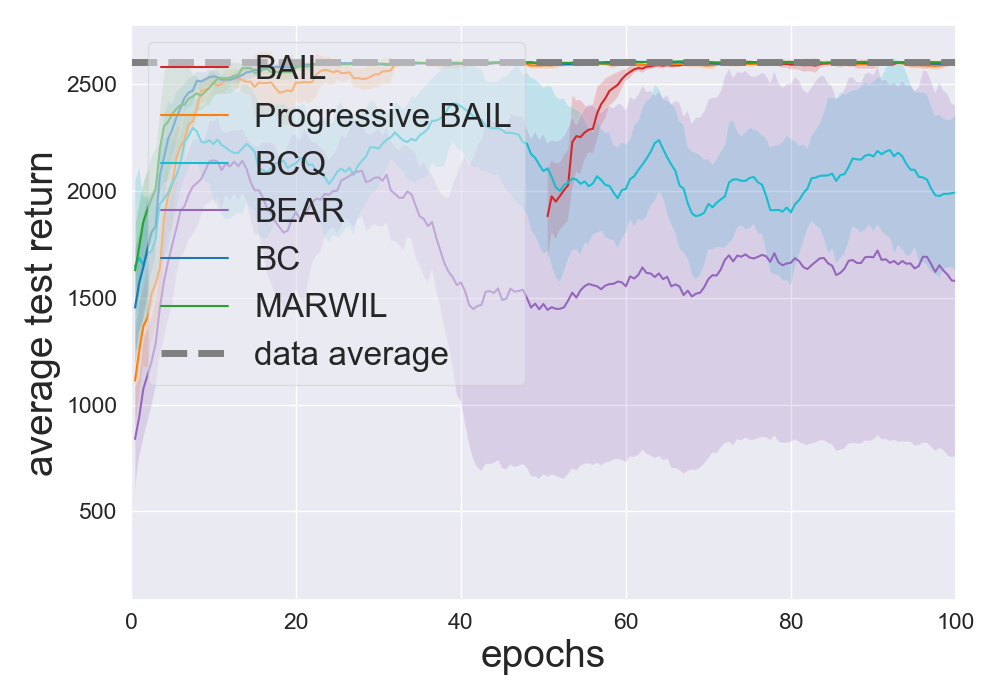}
 \caption{Hopper, batch 1}
\end{subfigure}
\begin{subfigure}{0.24\columnwidth}
 \centering
 \includegraphics[width=0.99\linewidth]{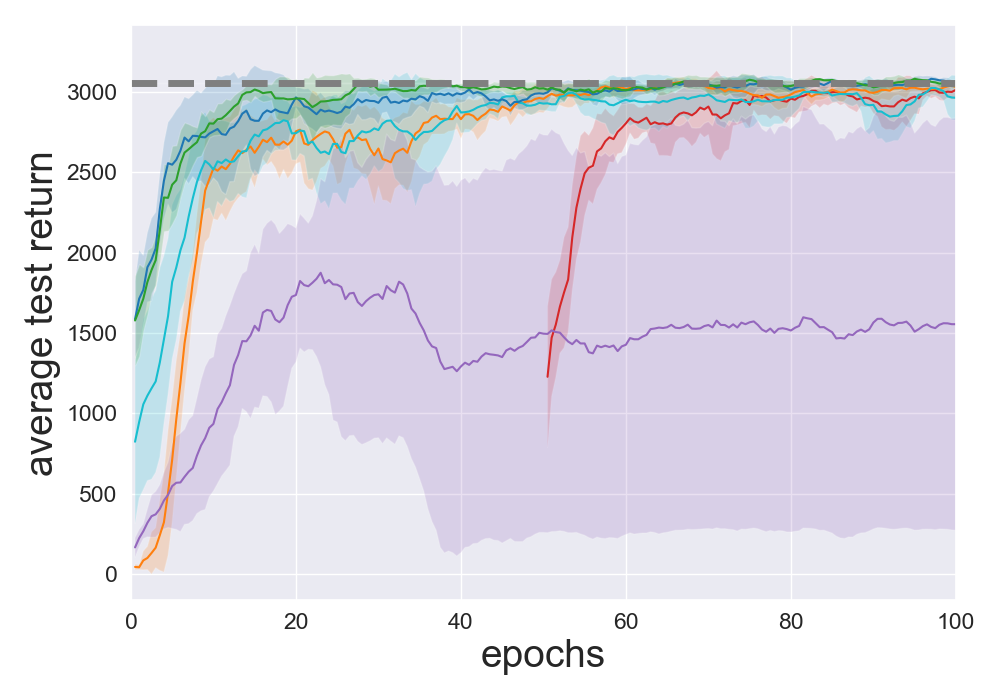}
 \caption{Hopper, batch 2}
\end{subfigure}
\begin{subfigure}{0.24\columnwidth}
 \centering
 \includegraphics[width=0.99\linewidth]{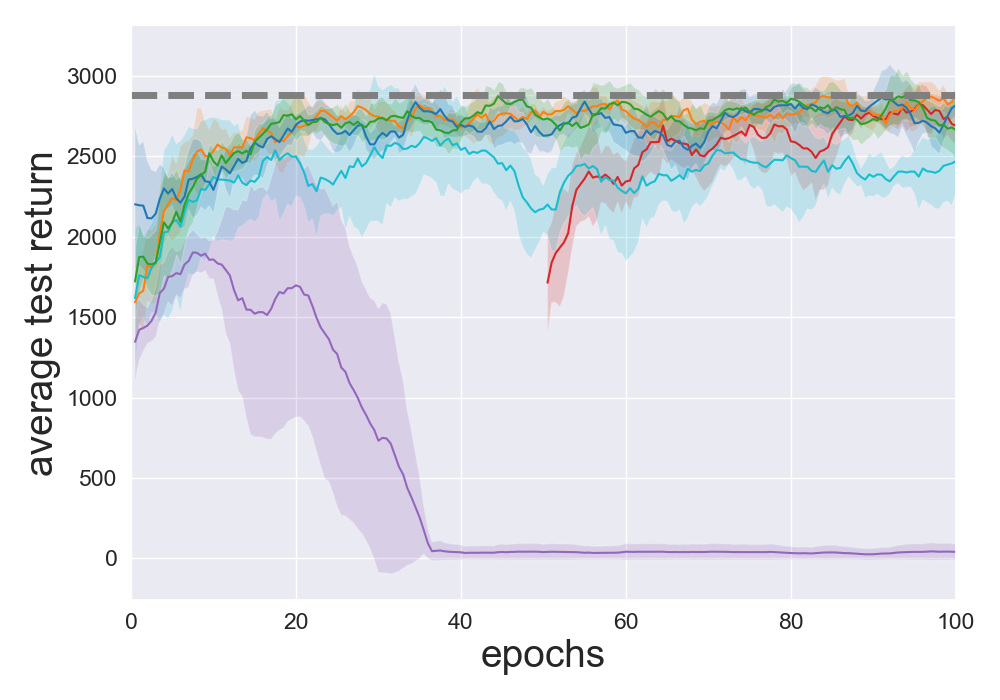}
 \caption{Walker2d, batch 1}
\end{subfigure}
\begin{subfigure}{0.24\columnwidth}
 \centering
 \includegraphics[width=0.99\linewidth]{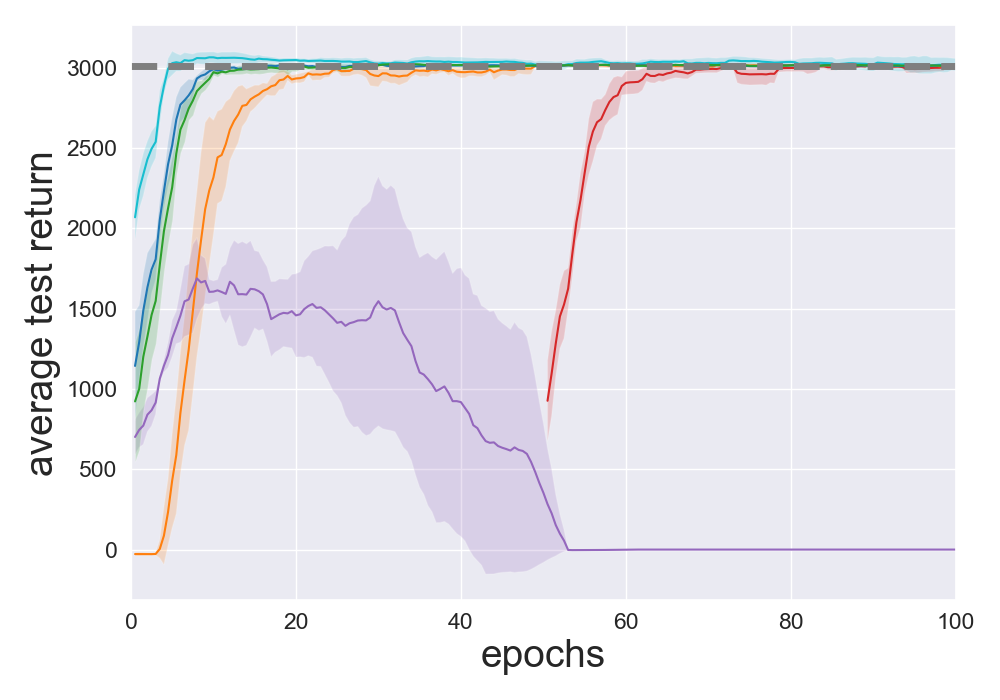}
 \caption{Walker2d, batch 2}
\end{subfigure}
\begin{subfigure}{0.24\columnwidth}
 \centering
 \includegraphics[width=0.99\linewidth]{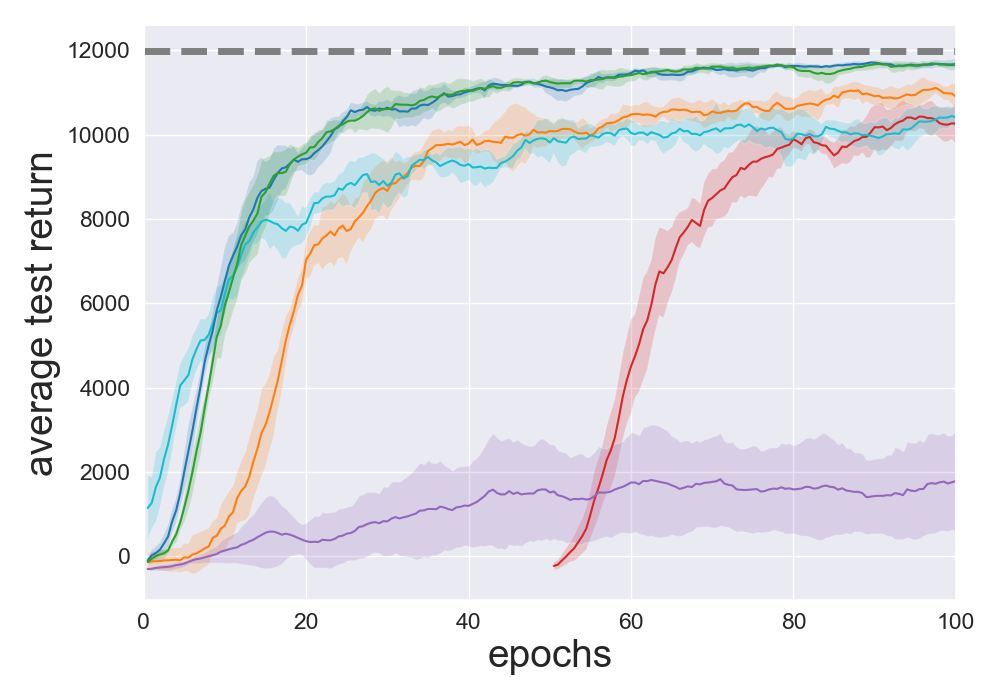}
 \caption{HalfCheetah, batch 1}
\end{subfigure}
\begin{subfigure}{0.24\columnwidth}
 \centering
 \includegraphics[width=0.99\linewidth]{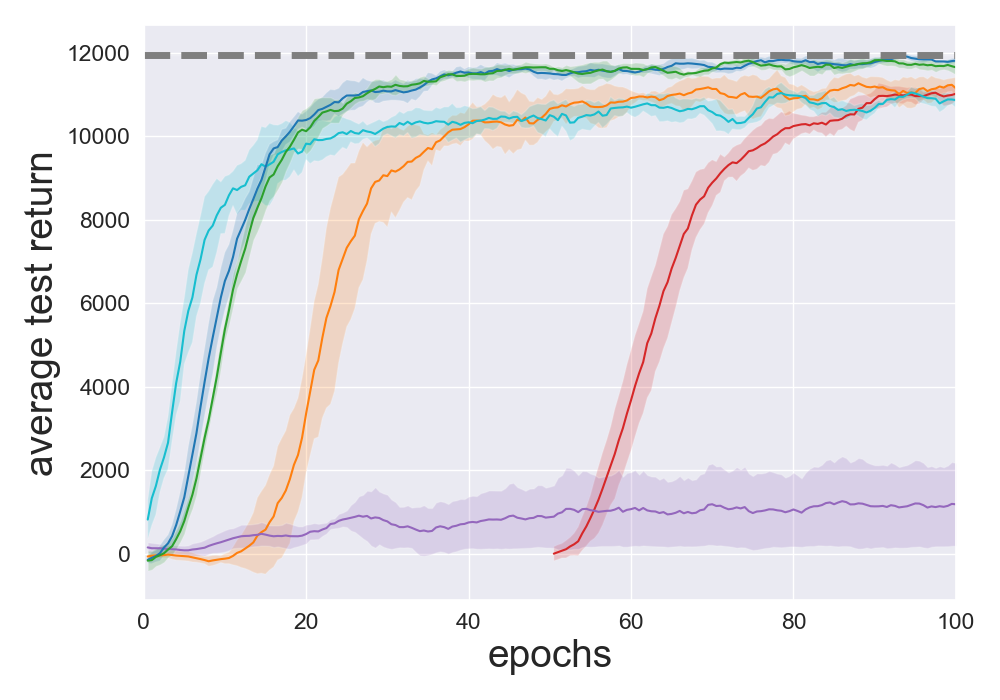}
 \caption{HalfCheetah, batch 2}
\end{subfigure}
\begin{subfigure}{0.24\columnwidth}
 \centering
 \includegraphics[width=0.99\linewidth]{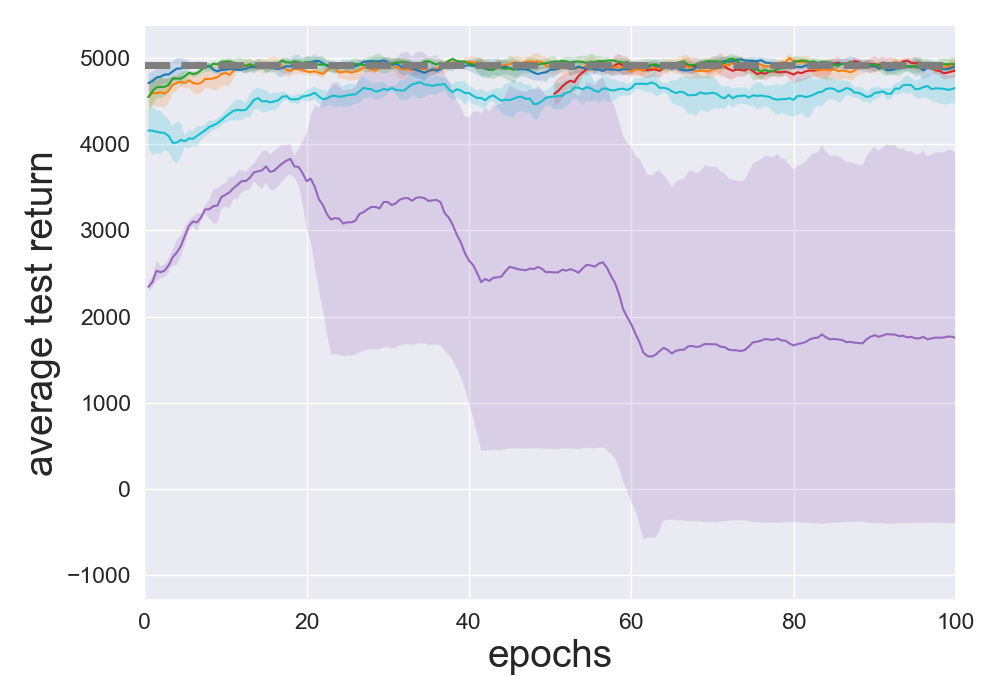}
 \caption{Ant, batch 1}
\end{subfigure}
\begin{subfigure}{0.24\columnwidth}
 \centering
 \includegraphics[width=0.99\linewidth]{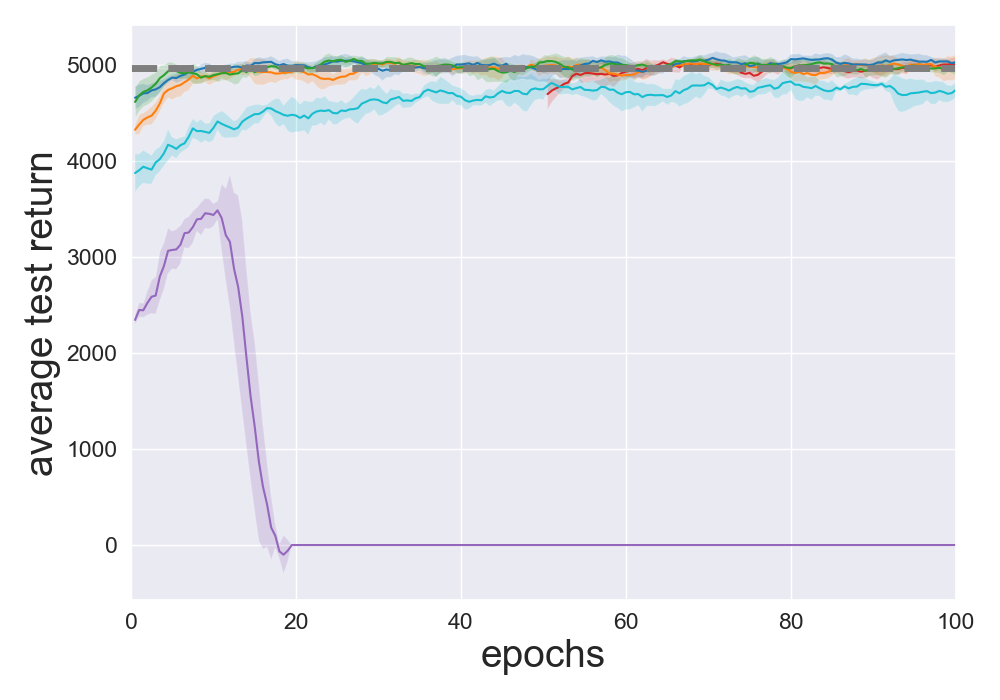}
 \caption{Ant, batch 2}
\end{subfigure}
\caption{Performance of batch DRL algorithms on SAC optimal execution batches with $\sigma = 0$. The policy networks for all algorithms are trained for 100 epochs except BAIL, which is trained for 50 epochs after training the upper envelope for 50 epochs.}
\label{fig:bail_optimal_no_sigma}
\vskip -0.1in
\end{figure}

\begin{figure}[ht]
\vskip 0.1in
\centering
\begin{subfigure}{0.24\columnwidth}
 \centering
 \includegraphics[width=0.99\linewidth]{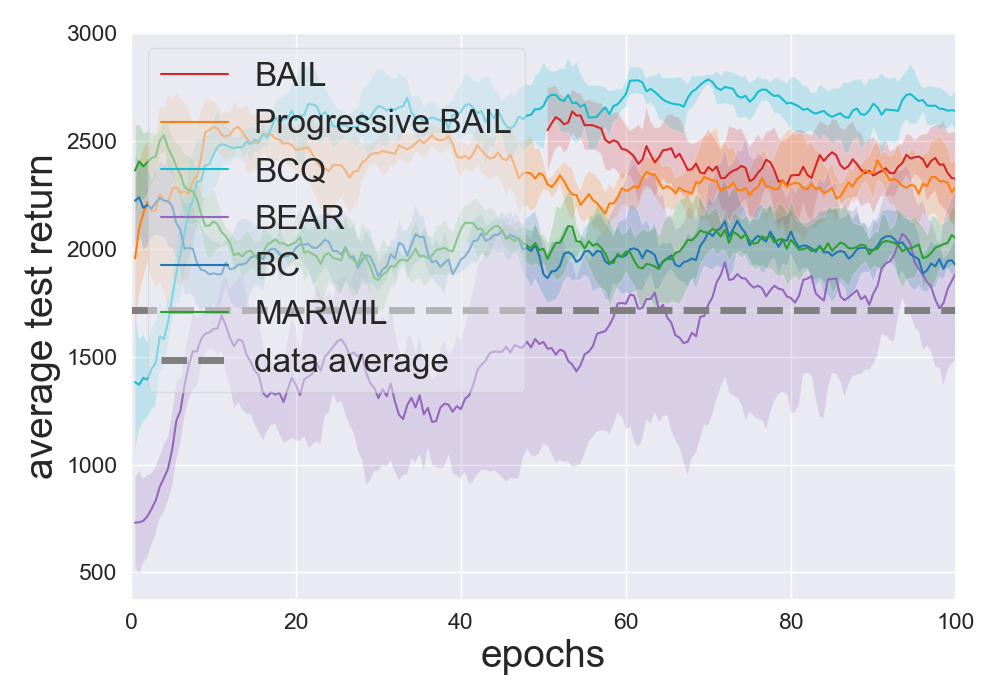}
 \caption{Hopper, batch 1}
\end{subfigure}
\begin{subfigure}{0.24\columnwidth}
 \centering
 \includegraphics[width=0.99\linewidth]{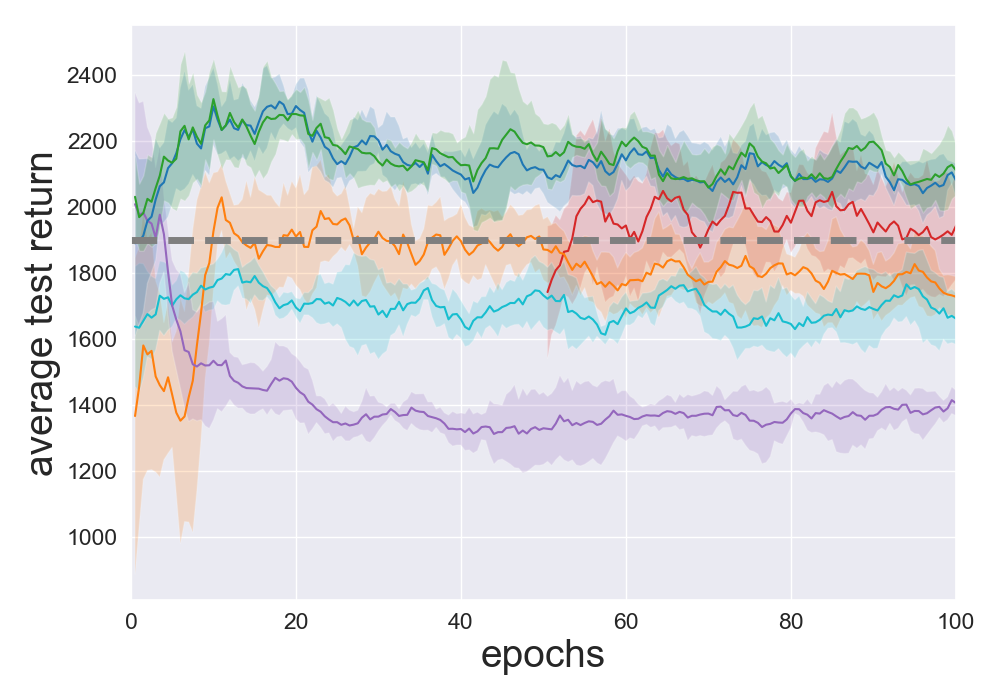}
 \caption{Hopper, batch 2}
\end{subfigure}
\begin{subfigure}{0.24\columnwidth}
 \centering
 \includegraphics[width=0.99\linewidth]{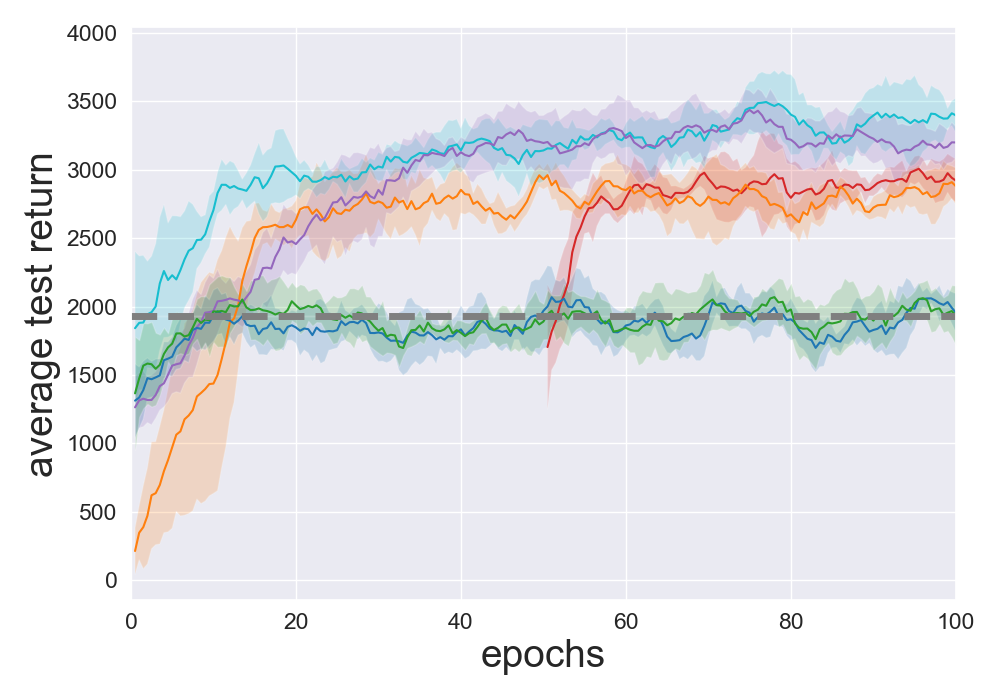}
 \caption{Walker2d, batch 1}
\end{subfigure}
\begin{subfigure}{0.24\columnwidth}
 \centering
 \includegraphics[width=0.99\linewidth]{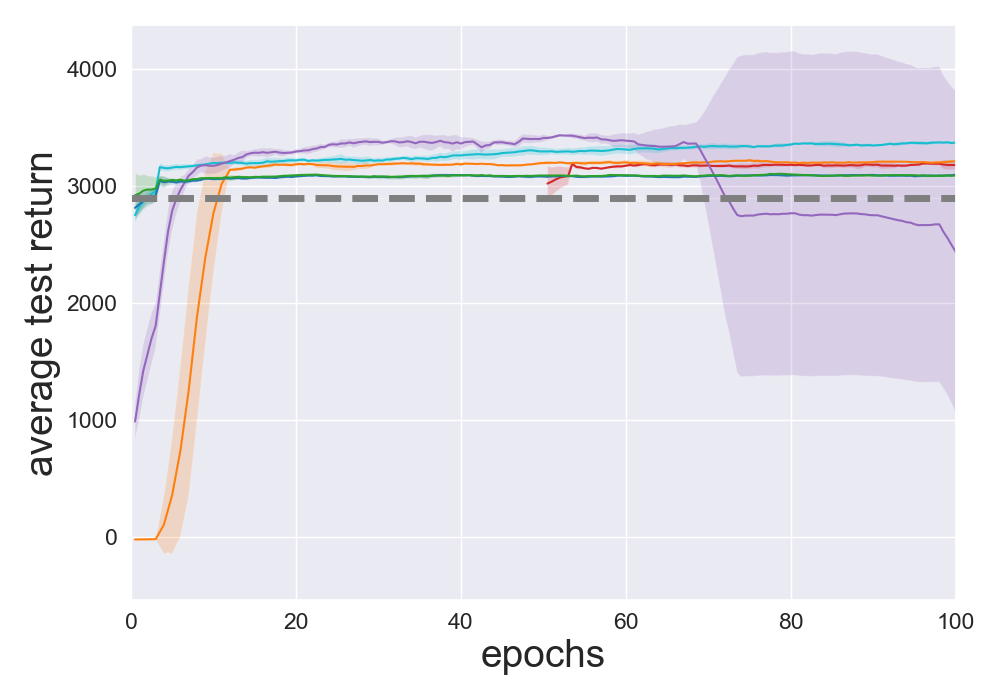}
 \caption{Walker2d, batch 2}
\end{subfigure}
\begin{subfigure}{0.24\columnwidth}
 \centering
 \includegraphics[width=0.99\linewidth]{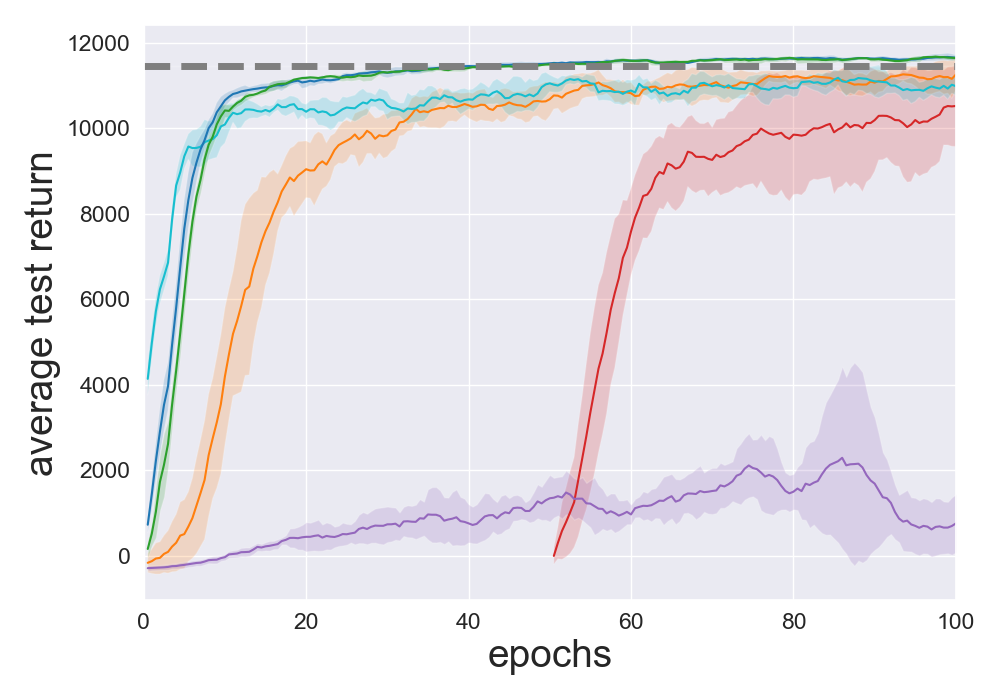}
 \caption{HalfCheetah, batch 1}
\end{subfigure}
\begin{subfigure}{0.24\columnwidth}
 \centering
 \includegraphics[width=0.99\linewidth]{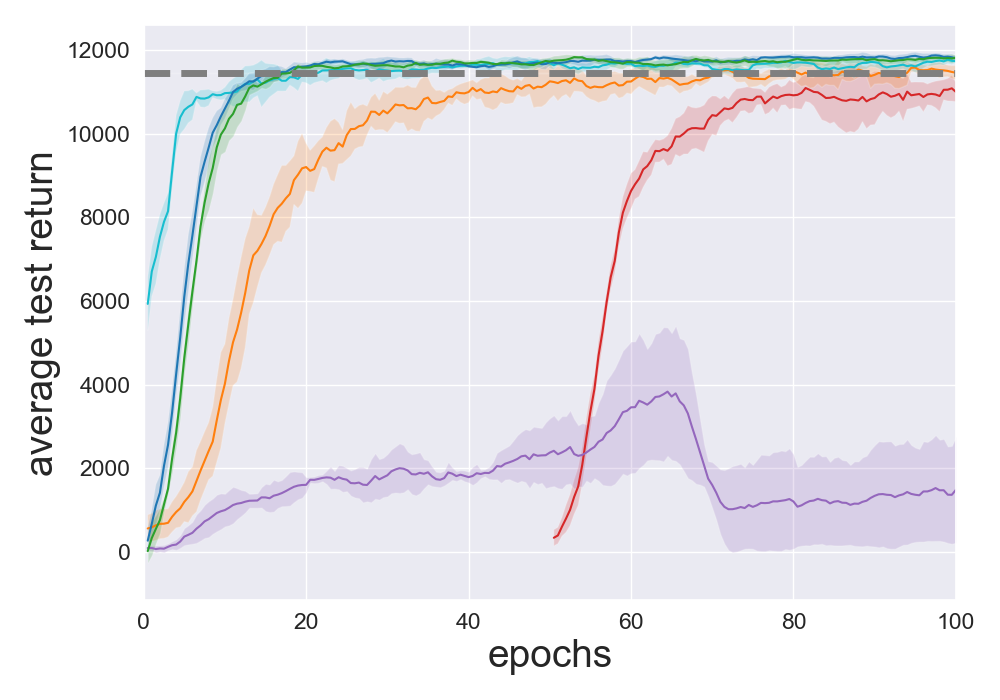}
 \caption{HalfCheetah, batch 2}
\end{subfigure}
\begin{subfigure}{0.24\columnwidth}
 \centering
 \includegraphics[width=0.99\linewidth]{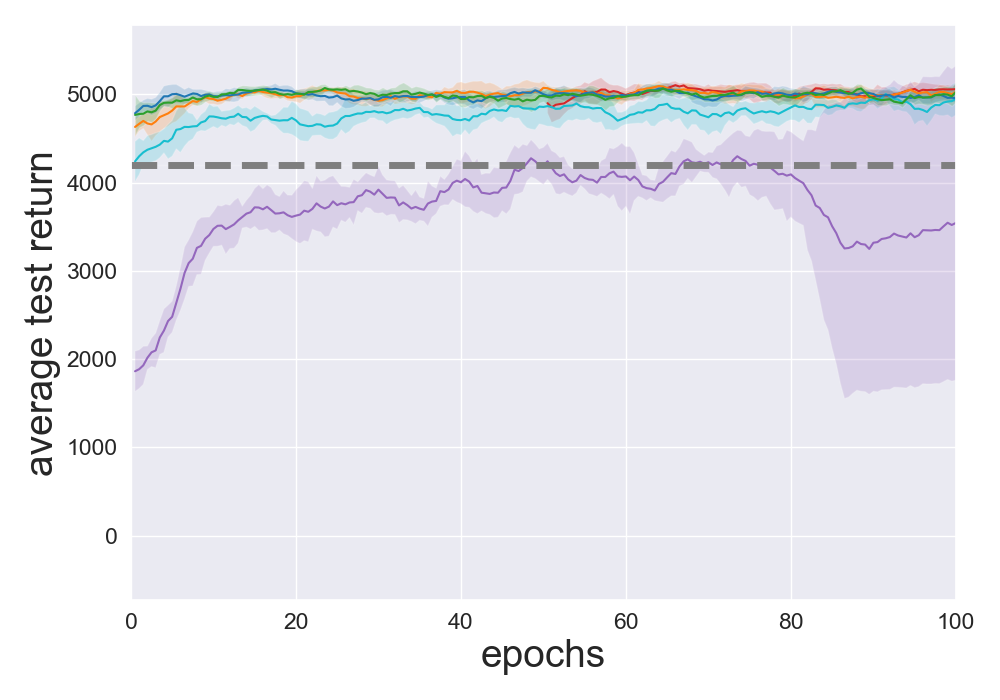}
 \caption{Ant, batch 1}
\end{subfigure}
\begin{subfigure}{0.24\columnwidth}
 \centering
 \includegraphics[width=0.99\linewidth]{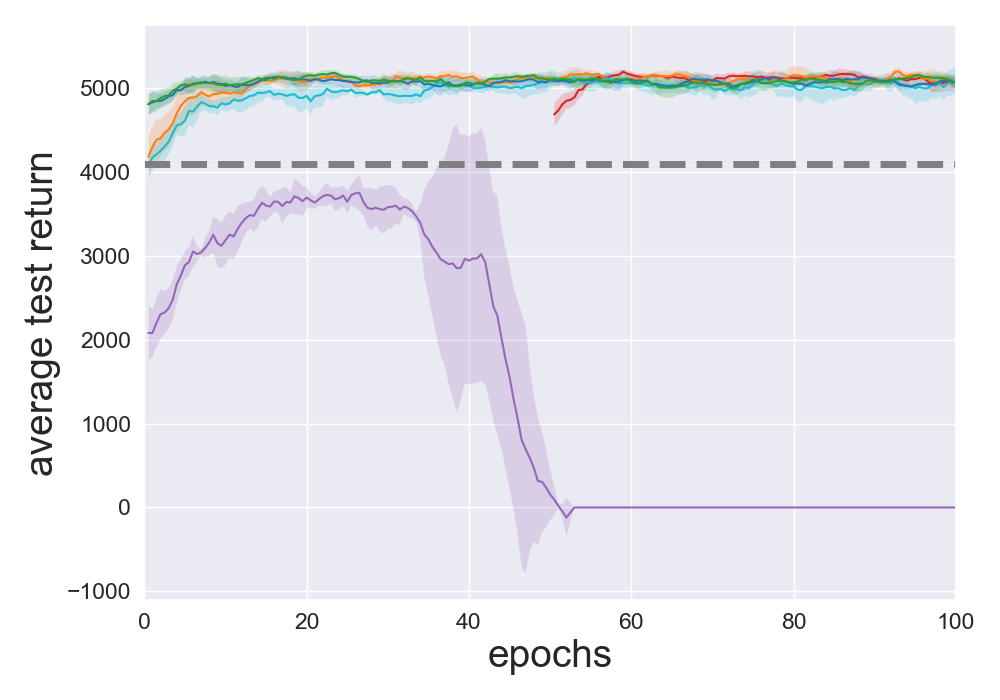}
 \caption{Ant, batch 2}
\end{subfigure}
\caption{Performance of batch DRL algorithms on SAC optimal execution batches with $\sigma = \sigma(s)$. The policy networks for all algorithms are trained for 100 epochs except BAIL, which is trained for 50 epochs after training the upper envelope for 50 epochs.}
\label{fig:bail_optimal_with_sigma}
\vskip -0.1in
\end{figure}

\newpage
\subsection{Learning curves for Humanoid}

\begin{figure}[ht]
\vskip 0.1in
\centering
\begin{subfigure}{0.24\columnwidth}
 \centering
 \includegraphics[width=0.99\linewidth]{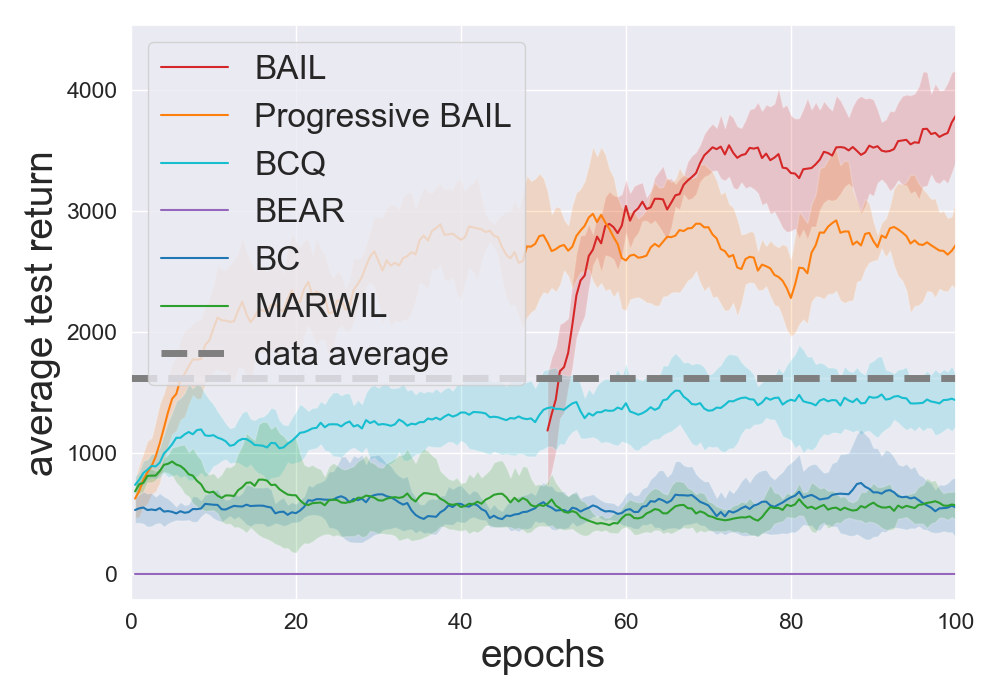}
 \caption{training data, batch 1}
\end{subfigure}
\begin{subfigure}{0.24\columnwidth}
 \centering
 \includegraphics[width=0.99\linewidth]{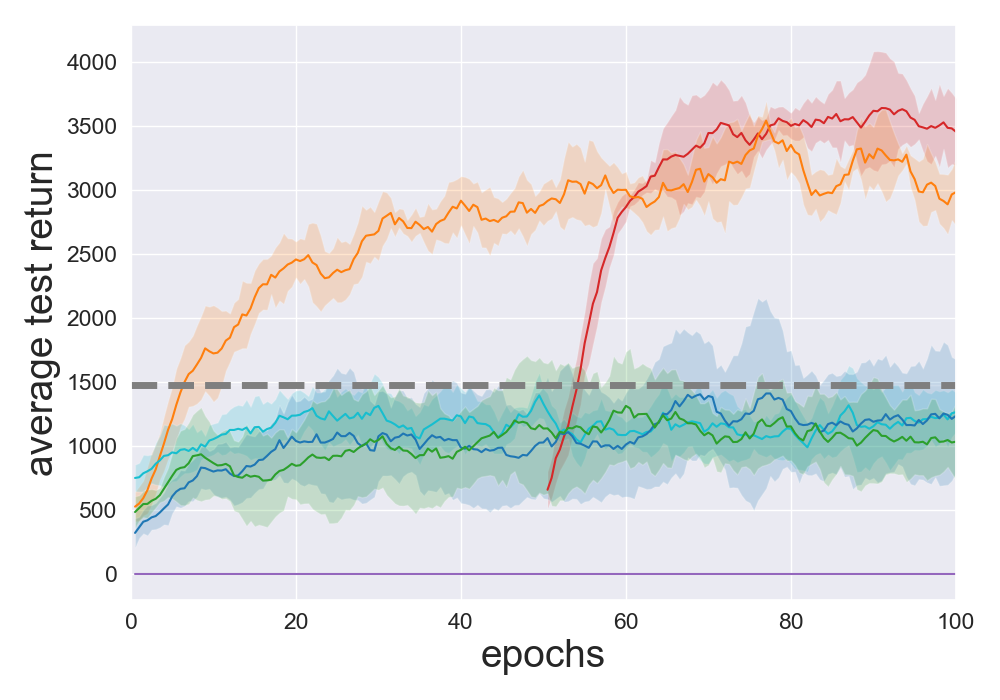}
 \caption{training data, batch 2}
\end{subfigure}

\begin{subfigure}{0.24\columnwidth}
 \centering
 \includegraphics[width=0.99\linewidth]{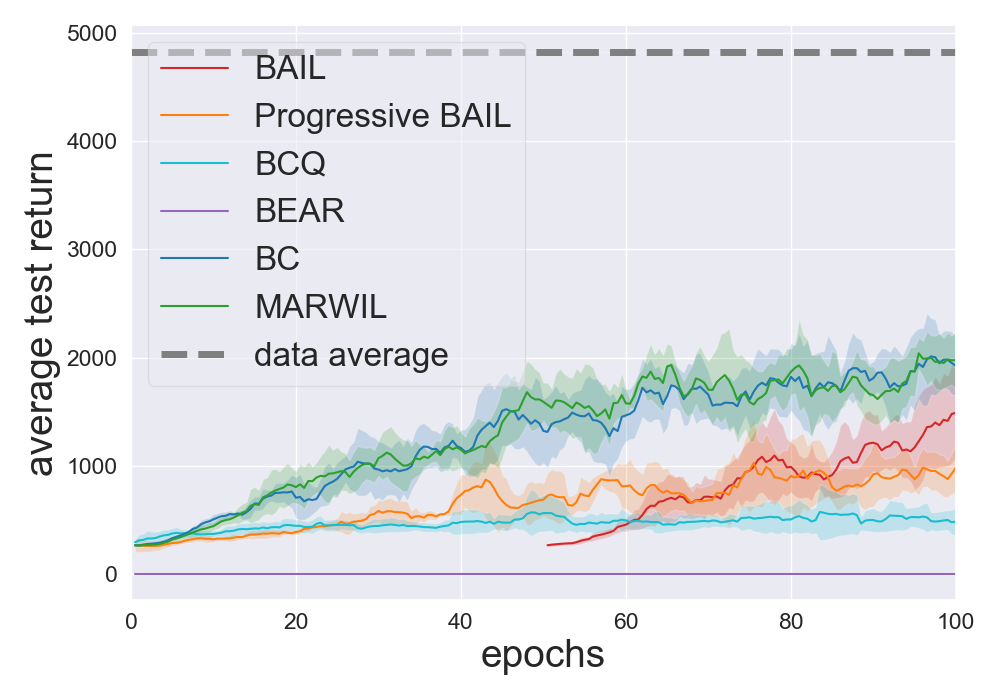}
 \caption{mediocre $\sigma = \sigma(s)$, batch 1}
\end{subfigure}
\begin{subfigure}{0.24\columnwidth}
 \centering
 \includegraphics[width=0.99\linewidth]{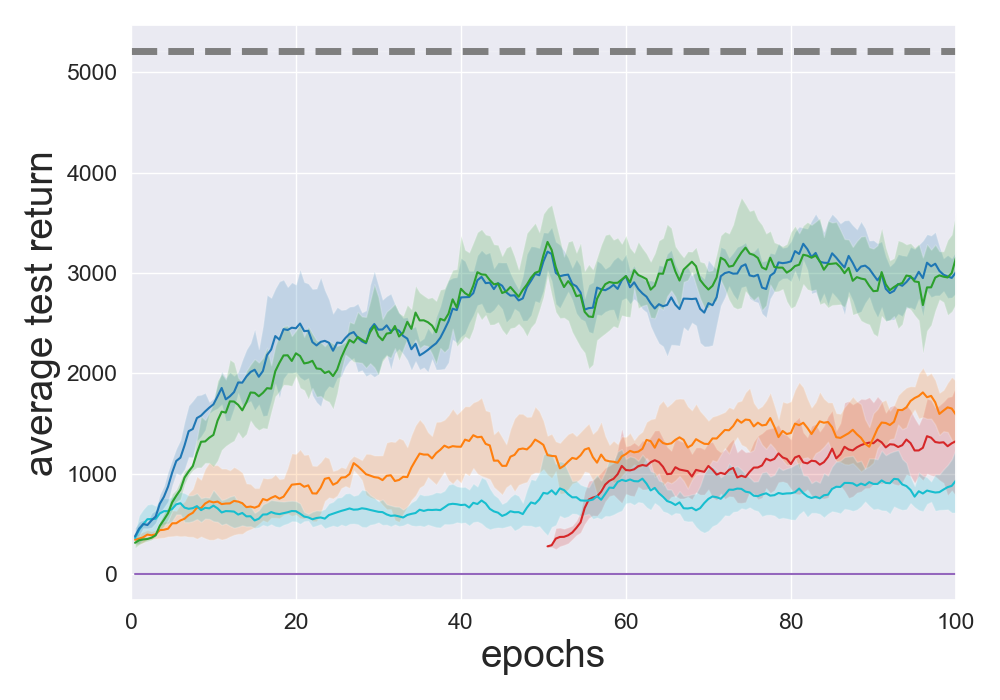}
 \caption{mediocre $\sigma = \sigma(s)$, batch 2}
\end{subfigure}
\begin{subfigure}{0.24\columnwidth}
 \centering
 \includegraphics[width=0.99\linewidth]{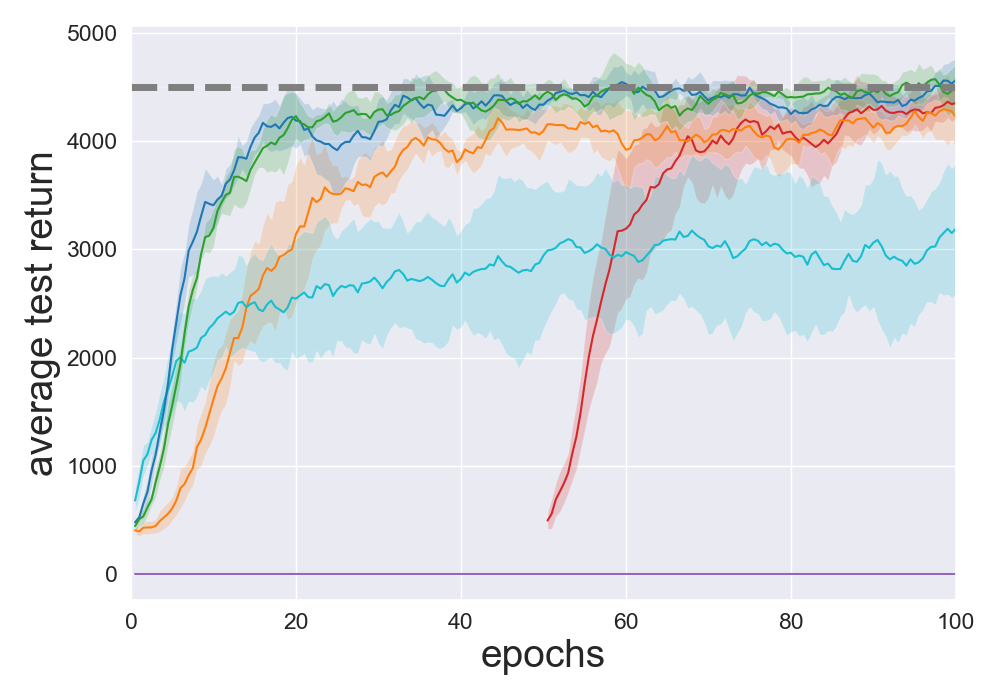}
 \caption{mediocre $\sigma = 0$, batch 1}
\end{subfigure}
\begin{subfigure}{0.24\columnwidth}
 \centering
 \includegraphics[width=0.99\linewidth]{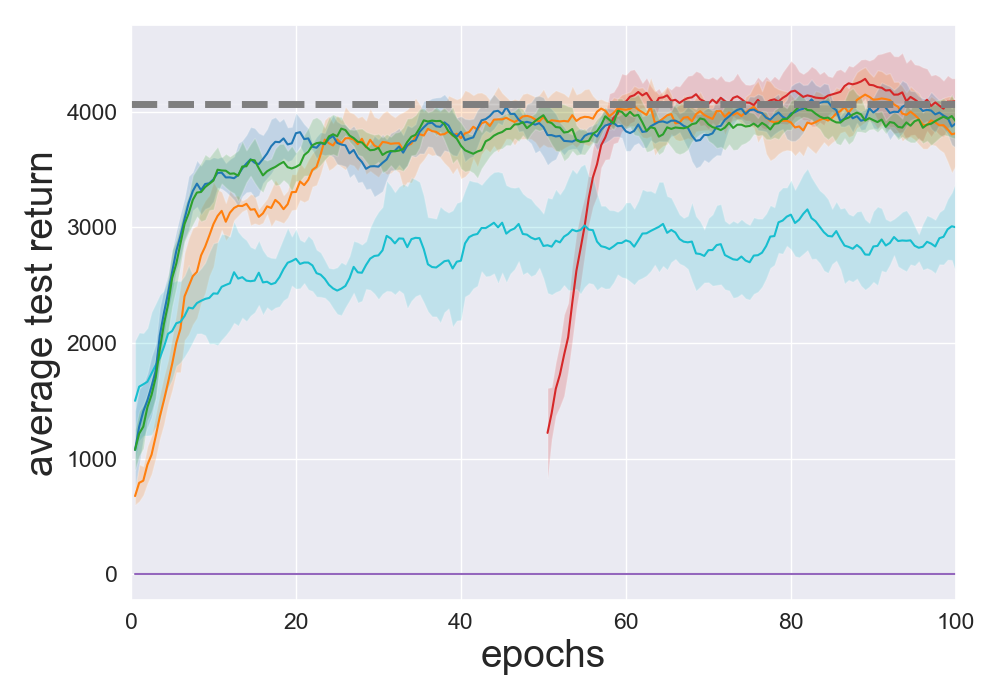}
 \caption{mediocre $\sigma = 0$, batch 2}
\end{subfigure}
\begin{subfigure}{0.24\columnwidth}
 \centering
 \includegraphics[width=0.99\linewidth]{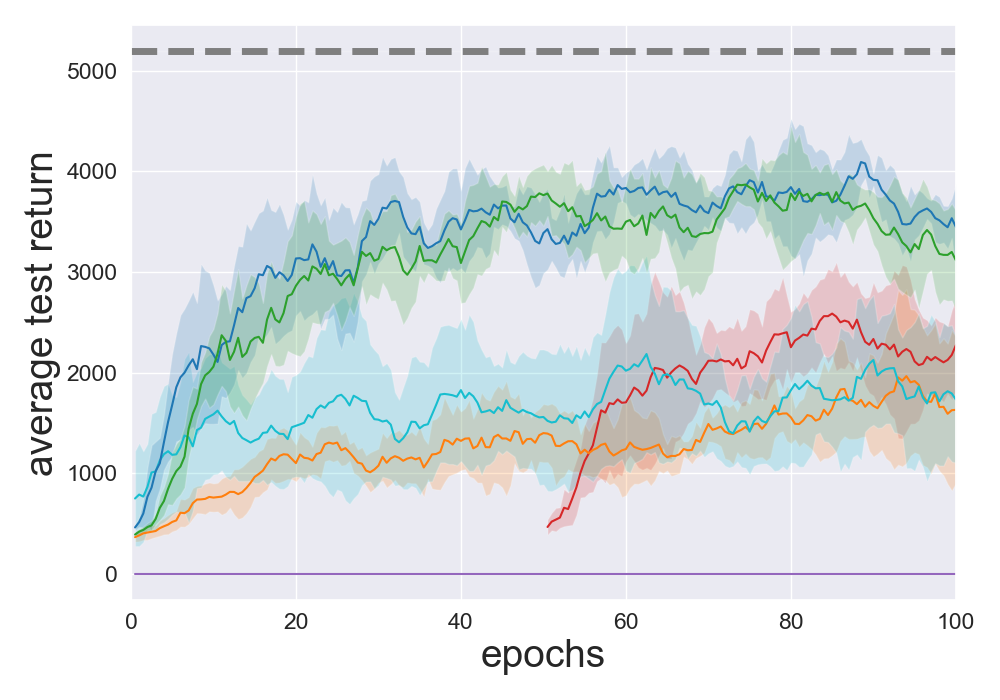}
 \caption{optimal $\sigma = \sigma(s)$, batch 1}
\end{subfigure}
\begin{subfigure}{0.24\columnwidth}
 \centering
 \includegraphics[width=0.99\linewidth]{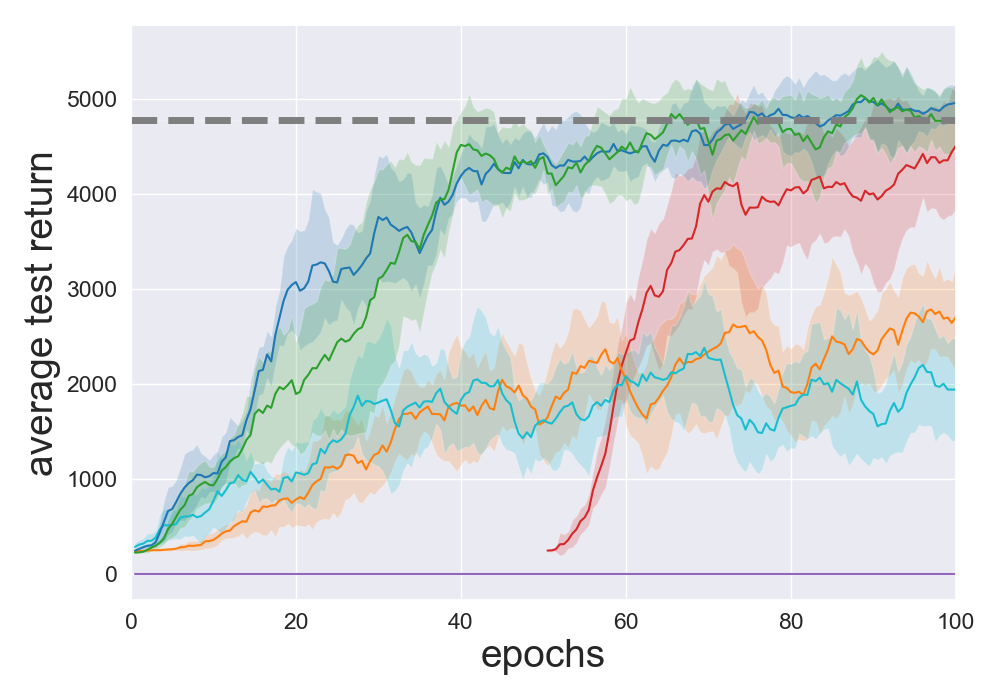}
 \caption{optimal $\sigma = \sigma(s)$, batch 2}
\end{subfigure}
\begin{subfigure}{0.24\columnwidth}
 \centering
 \includegraphics[width=0.99\linewidth]{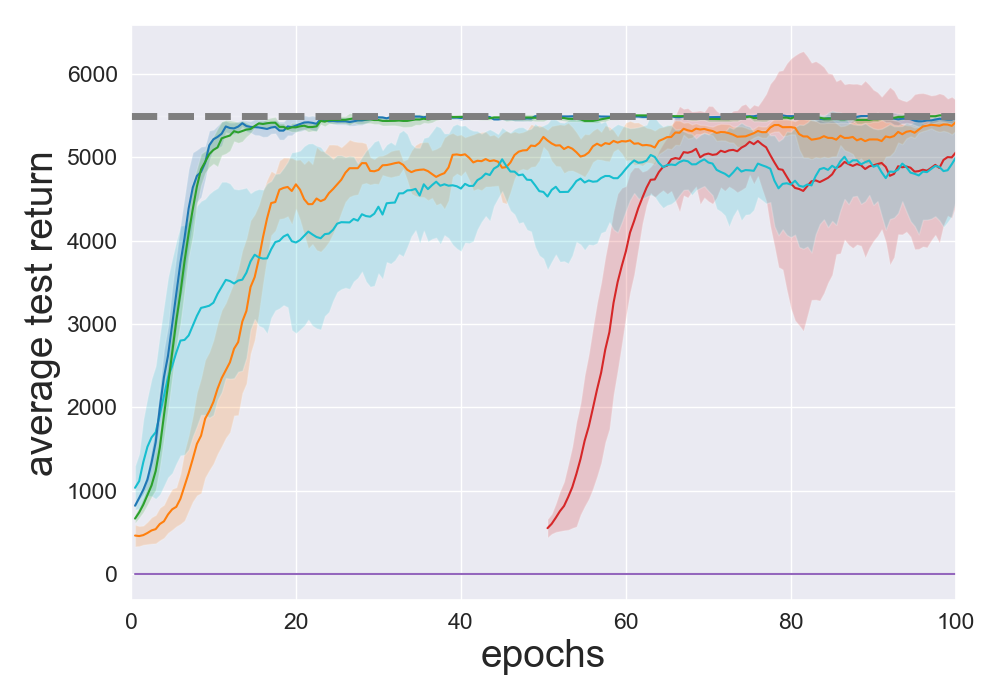}
 \caption{optimal $\sigma = 0$, batch 1}
\end{subfigure}
\begin{subfigure}{0.24\columnwidth}
 \centering
 \includegraphics[width=0.99\linewidth]{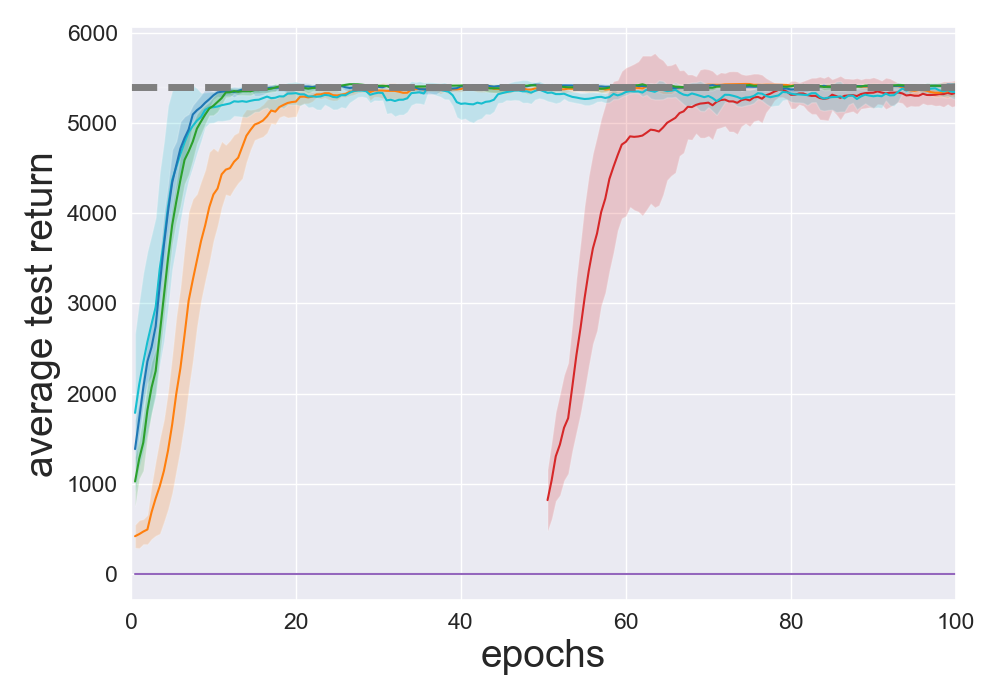}
 \caption{optimal $\sigma = 0$, batch 2}
\end{subfigure}
\caption{Performance of batch DRL algorithms with the  Humanoid-v2 environment. All batches are obtained with SAC.}
\label{fig:bail_humanoid_complete}
\vskip -0.1in
\end{figure}

\newpage
\section{Visualization of the Upper Envelopes}

\begin{figure}[ht]
\vskip 0.1in
\centering
\begin{subfigure}{0.24\columnwidth}
 \centering
 \includegraphics[width=0.99\linewidth]{figures/ue_visual_Stat_FinalSigma0-5_Hopper-v2_3_1000K_r1000_g0-99_Gain_lossk1000_s3.png}
 \caption{Hopper $\sigma=0.5$ 1st}
\end{subfigure}
\begin{subfigure}{0.24\columnwidth}
 \centering
 \includegraphics[width=0.99\linewidth]{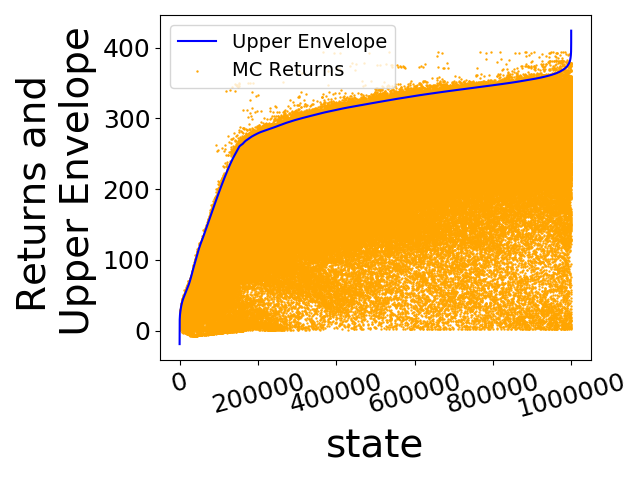}
 \caption{Hopper $\sigma=0.5$ 2nd}
\end{subfigure}
\begin{subfigure}{0.24\columnwidth}
 \centering
 \includegraphics[width=0.99\linewidth]{figures/ue_visual_Stat_FinalSigma0-5_Walker2d-v2_3_1000K_r1000_g0-99_Gain_lossk1000_s5.png}
 \caption{Walker2d $\sigma=0.5$ 1st}
\end{subfigure}
\begin{subfigure}{0.24\columnwidth}
 \centering
 \includegraphics[width=0.99\linewidth]{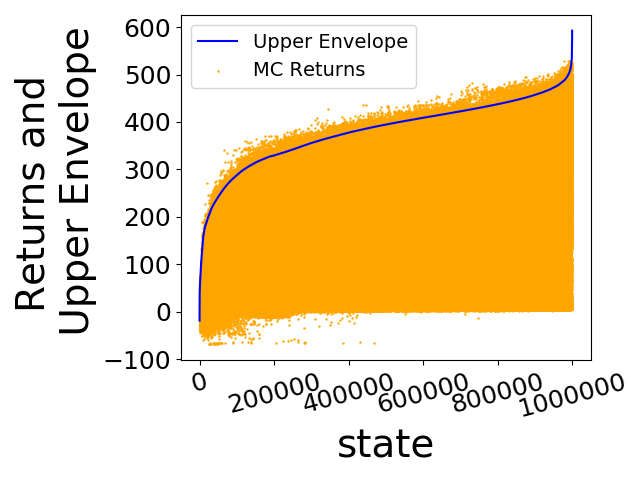}
 \caption{Walker2d $\sigma=0.5$ 2nd}
\end{subfigure}
\begin{subfigure}{0.24\columnwidth}
 \centering
 \includegraphics[width=0.99\linewidth]{figures/ue_visual_Stat_FinalSigma0-5_HalfCheetah-v2_3_1000K_r1000_g0-99_Gain_lossk1000_s3.png}
 \caption{HalfCheetah $\sigma=0.5$ 1st}
\end{subfigure}
\begin{subfigure}{0.24\columnwidth}
 \centering
 \includegraphics[width=0.99\linewidth]{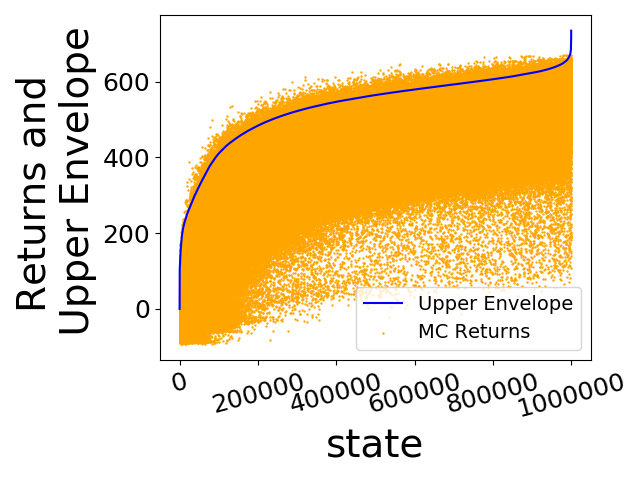}
 \caption{HalfCheetah $\sigma=0.5$ 2nd}
\end{subfigure}
\begin{subfigure}{0.24\columnwidth}
 \centering
 \includegraphics[width=0.99\linewidth]{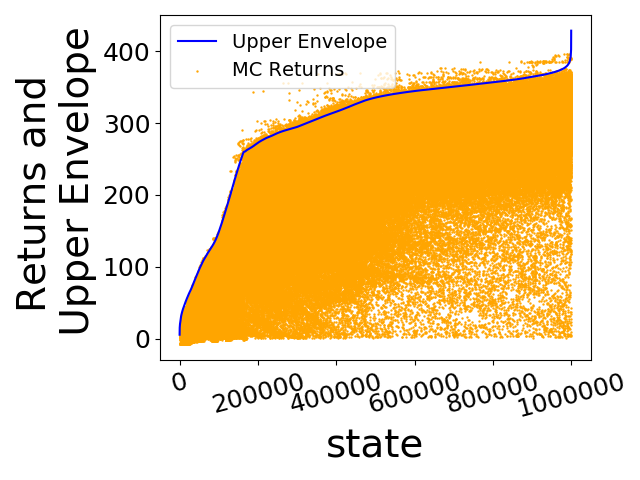}
 \caption{Hopper $\sigma=0.1$ 1st}
\end{subfigure}
\begin{subfigure}{0.24\columnwidth}
 \centering
 \includegraphics[width=0.99\linewidth]{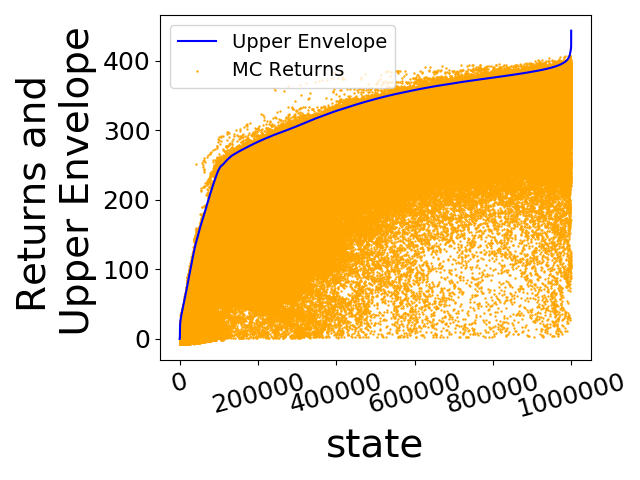}
 \caption{Hopper $\sigma=0.1$ 2nd}
\end{subfigure}
\begin{subfigure}{0.24\columnwidth}
 \centering
 \includegraphics[width=0.99\linewidth]{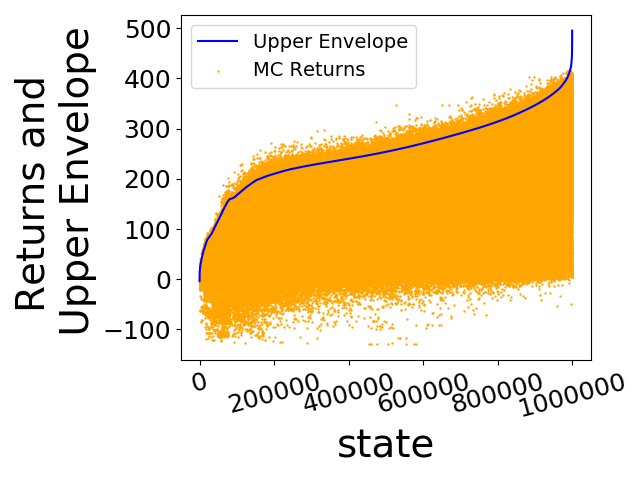}
 \caption{Walker2d $\sigma=0.1$ 1st}
\end{subfigure}
\begin{subfigure}{0.24\columnwidth}
 \centering
 \includegraphics[width=0.99\linewidth]{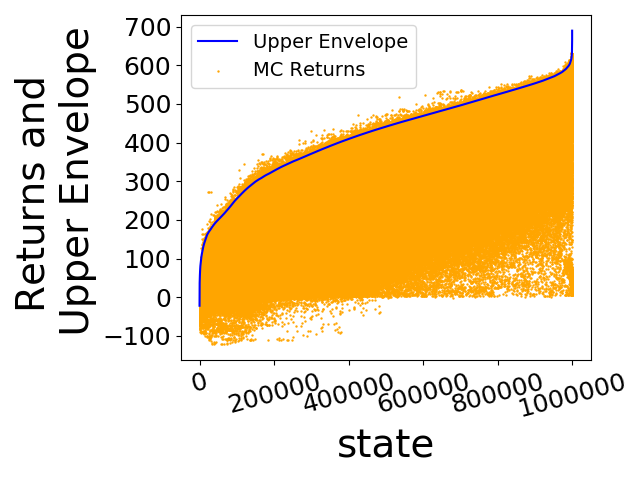}
 \caption{Walker2d $\sigma=0.1$ 2nd}
\end{subfigure}
\begin{subfigure}{0.24\columnwidth}
 \centering
 \includegraphics[width=0.99\linewidth]{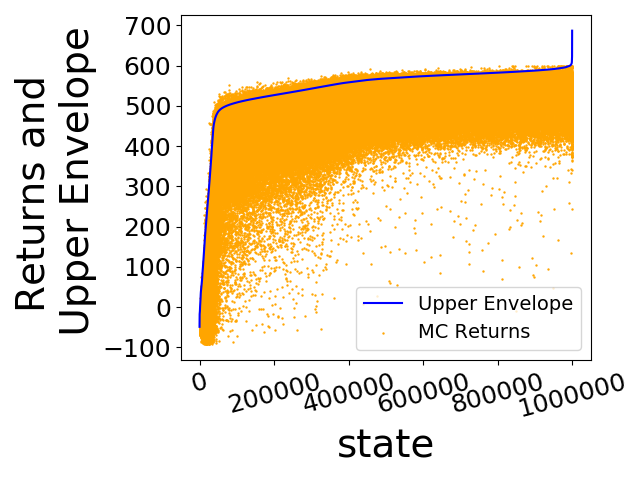}
 \caption{HalfCheetah $\sigma=0.1$ 1st}
\end{subfigure}
\begin{subfigure}{0.24\columnwidth}
 \centering
 \includegraphics[width=0.99\linewidth]{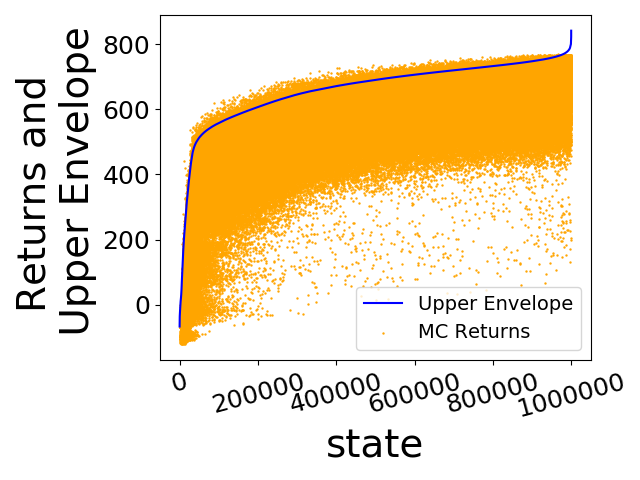}
 \caption{HalfCheetah $\sigma=0.1$ 2nd}
\end{subfigure}
\caption{Typical Upper Envelopes for BAIL. For each figure, states are ordered from lowest $V(s_i)$ upper envelope value to highest. Thus the upper envelope curve is monotonically increasing. Each curve is trained with one million returns, shown with the orange dots. Note that the upper envelope lies above most data points but not all data points.}
\label{fig:ue_visual_complete}
\vskip -0.1in
\end{figure}

\section{Computing Infrastructure}
Experiments are run on Intel Xeon Gold 6248 CPU nodes, each job runs on a single CPU with base frequency of 2.50GHZ.

\end{document}